\documentclass[conference]{IEEEtran}
\usepackage{cite}
\usepackage{amsmath,amssymb,amsfonts}
\usepackage{graphicx}
\usepackage{textcomp}
\usepackage{multirow}
\usepackage{comment}
\usepackage{enumitem, array} 
\usepackage{booktabs} 
\usepackage{adjustbox}
\usepackage{caption}
\usepackage{subcaption}

\usepackage{placeins}
\usepackage{tabularx}
\usepackage{mathtools}

\usepackage[english]{babel}
\usepackage[utf8]{inputenc}
\usepackage{algorithm}
\usepackage[noend]{algpseudocode}

\usepackage{amsthm} 

\theoremstyle{definition}

\algnewcommand{\algorithmicand}{\textbf{ and }} 
\algnewcommand{\algorithmicor}{\textbf{ or }} 
\algnewcommand{\OR}{\algorithmicor} 
\algnewcommand{\AND}{\algorithmicand} 

\usepackage{lipsum} 

\def\BibTeX{{\rm B\kern-.05em{\sc i\kern-.025em b}\kern-.08em
    T\kern-.1667em\lower.7ex\hbox{E}\kern-.125emX}}
\begin{document}

\title{Information Fusion for Assistance Systems in Production Assessment}

\author{\IEEEauthorblockN{Fernando Ar\'{e}valo\IEEEauthorrefmark{1},
		 Christian Alison M. Piolo,
		 M. Tahasanul Ibrahim\IEEEauthorrefmark{4},
		Andreas Schwung\IEEEauthorrefmark{4}
	} 
	\IEEEauthorblockA{\IEEEauthorrefmark{1}
		Ruhr-Universit\"{a}t Bochum, Bochum 44801, Germany\\
		Email: Fernando.ArevaloNavas@ruhr-uni-bochum.de}
	\IEEEauthorblockA{\IEEEauthorrefmark{4}Department of Automation Technology\\
		South Westphalia University of Applied Sciences, Campus Soest, Germany\\
  }
  
}




\maketitle

\begin{abstract}
Assistance systems are becoming a frequent company for machine operators because they often summarize vital information for the production and machine condition. This information supports the operator during decision-making while having an (unknown) fault. Moreover, the assistance systems often provide a procedure to handle the (faulty) condition. Typical components of an assistance system are a detection system, a knowledge base, and an (interactive) user interface. 
Data-based models are a common option for a detection system. However, systems that rely purely on data-based models are normally trained with a specific set of data, which cannot necessarily prevent data drift. Thus, an anomaly or unknown condition detection mechanism is required to handle data with new fault cases. 
Besides, the model's capability to adapt to the unknown condition is equally important to anomaly detection—in other words, its capability to update itself automatically.
Alternatively, expert-centered models are powered by the knowledge of operators, which provides the models with production context and expert domain knowledge.
The challenge lies in combining both systems and which framework can be used to achieve this fusion.
We propose a novel methodology to define assistance systems that rely on information fusion to combine different sources of information while providing an assessment. The main contribution of this paper is providing a general framework for the fusion of $n$ number of information sources using the evidence theory. 
The fusion provides a more robust prediction and an associated uncertainty that can be used to assess the prediction likeliness. Moreover, we provide a methodology for the information fusion of two primary sources: an ensemble classifier based on machine data and an expert-centered model. We demonstrate the information fusion approach using data from an industrial setup, which rounds up the application part of this research. Furthermore, we address the problem of data drift by proposing a methodology to update the data-based models using an evidence theory approach. We validate the approach using the Benchmark Tennessee Eastman while doing an ablation study of the model update parameters.

\end{abstract}



\section{Introduction}\label{section__intro}
systems accompany the operators during the machinery operation by providing assessment during decision-making. These systems support the operators with (real-time) information on the process in terms of production, machine condition, and recommendations to handle faults or to improve the machine's performance.
Assistance systems have typical components such as a (real-time) data collection system, a (fault) detection system, a knowledge base, a computing engine, and an (interactive) user interface \cite{GalloRienzo2022} \cite{MazziaKhaliq2020} \cite{RahmGraube2018}.
Due to their high performance, data-based models are a popular choice when selecting a detection system with reported applications in 
medicine \cite{Shamim2022}, industry \cite{GalloRienzo2022} \cite{RahmGraube2018}, road infrastructure \cite{GagliardiStaderini2022}, and agriculture \cite{MazziaKhaliq2020}.
Usually, the data-based models are trained using a specific dataset presenting good results. However, not all data-based models can handle new upcoming faults in the data.
Hence, an anomaly detection system must have a mechanism to recognize an upcoming anomaly and the capability to learn upcoming data that differs from the original training data. Equally important is the system's capability to adapt or retrain the data-based models automatically.
The retraining or automatic update of the models must consider a minimum size of training data that assures that the models capture the essential patterns to be learned.   

Systems composed by the combination or fusion of several individual models often present better results and robustness than individual models (e.g., bagging and boosting).
Though data-based models attain high performance, alternatively, expert-centered knowledge-based models provide versatile features, which are production context and expert domain knowledge. 
The challenge here lies in how to combine a data-based model and a knowledge-based model. Thus, a common framework is required to perform a fusion of both systems. Such a framework must provide not only a way to combine the models' outputs but to quantify the uncertainty. The uncertainty provides information regarding how reliable the combined system output is.

\begin{table}[!htbp]
	\centering
	\caption{List of symbols and abbreviations.}
	\begin{tabular}{l l}
		\toprule
		\textbf{Symbol} & \textbf{Description} \\ 		
		\midrule
            DSET            & Dempster-Shafer evidence theory\\
            ECET            & Ensemble classification using evidence theory\\
            KLAFATE         & Knowledge transfer framework using evidence theory\\
            DSRC            & Dempster-Shafer rule of combination\\
            YRC             & Yager rule of combination\\
            EC            & Ensemble classifier\\
		  $m$             & Mass function  \\
		  $U$             & Uncertainty  \\
  		$w$             & confidence weight\\
    	  $p$             & prediction\\
		  $D^{Tr}$        & Training data\\
		  $D^{Va}$        & Validation data\\
		  $D^{Te}$        & Testing data\\
		  $k$             & sensitivity to zero factor\\
            $F_{D}$        & Fusion using DSET rule of combination\\
            $F_{Y}$        & Fusion using Yager rule of combination\\
            $Ws$           & Window size\\ 
            $Th$           & Threshold size\\ 
            $Pt$           & Detection patience\\ 
		\bottomrule
	\end{tabular}
	\label{table__list_symbols} 
\end{table}

We propose a novel methodology for assistance systems that rely on information fusion in production assessment, in which several information sources can be combined into a more robust system output. The novelty of this paper is presenting a common framework that allows the fusion of several information sources on the decision level using the evidence theory. Besides, we quantify the uncertainty of the system output to provide a better assessment of system output reliability. An essential contribution of this paper is the ability of the data-based model to handle unknown fault cases in the data, which allows the model to update the models automatically.

The individual contributions of this paper are:
\begin{itemize}
\item A methodology for the automatic model update of ECSs, while feeding up data with unknown fault cases. The methodology includes an uncertainty monitoring strategy that improves the anomaly detection of the EC, stores the data of the unknown condition, and retrains the pool of classifiers of the EC. We present the parameters of the automatic update module: threshold size, window size, and detection patience.
The automatic update methodology is rounded up with experiments using the benchmark dataset Tennessee Eastman. The EC is tested using different fault class scenarios, in which we test the impact of a window during anomaly detection. Moreover, we present a detailed analysis of the automatic update parameters regarding retrained EC performance.
\item A general framework to combine $n$ number of information sources on the decision level to generate a robust system prediction. The framework uses the Dempster-Shafer evidence theory. Besides, the framework quantifies the uncertainty of the prediction, which can be used to assess the reliability of the system prediction.
\item A methodology to combine a multiclass EC with an expert-centered knowledge-based model, in which we apply the general framework of the information fusion. The system architecture shows the components of each model, namely, the inference model and model update module. The application of the information fusion system is tested with the data of an industrial setup using a small-scaled bulk good system. The performance of the individual models (EC and knowledge-based) is compared with the combined system.
\end{itemize}


This paper is structured as follows: Section \ref{section__related} presents a literature survey on the main topics of this paper. The theoretical background is described in section \ref{theoretical}. Our proposed approach is detailed in section \ref{section__infusion}. Section \ref{section__information__fusion} and section \ref{section__model__update} present the methodology for information fusion and model update, respectively. Section \ref{section__usecase__retrain} portrays a use case for retraining the EC using the benchmark Tennessee Eastman. Whereas section \ref{section__usecase__assistance} presents a use case for information fusion using the data of a bulk good system laboratory plant. Finally, section \ref{section__conclusions} summarizes the conclusions and future work.

\section{Related Work}\label{section__related}
This section reviews the literature related to information fusion, update of data-based models, and assistance systems.


\subsection{Assistance Systems} 
Assistance systems provide valuable information for the users. They can be whether non-invasive or with direct control of the process.
The assistance can range from recommendation systems \cite{QinZhang2021} \cite{JinGuo2021}, interactive systems \cite{NagyRuppert2022}, or even systems that prevent actions from the user.
Architectures of assistance systems commonly contemplate the modules:  data collection, a condition detection engine, a knowledge base, and an (interactive) user interface \cite{LohfinkAnton2022}.
The (fault) condition detection engine is vital to identify the current state of the machinery or process. The engine is usually powered either by a knowledge-centered model \cite{LohfinkAnton2022} or a data-based model \cite{ArevaloPiolo2022}.
The knowledge base plays a crucial role in the assistance system because it provides the information that supports the user when a (faulty) condition is active \cite{LohfinkAnton2022}. There are different ways to build a knowledge base, namely using ontologies \cite{LohfinkAnton2022} \cite{BakakeuBrossog2019}, knowledge graphs\cite{NagyRuppert2022} \cite{SahlabKamm2021}, or static databases.
The proposed architectures of assistance systems contain the primary modules to support the users. However, there are factors to be considered, such as the update of the condition detection engine and the knowledge base, and the quantification of the system uncertainty. 
The challenge lies in a holistic architecture that addresses these factors and proposes the interactions of the primary systems.
This research differentiates from the state-of-the-art, in which we propose a holistic methodology using information fusion for assistance systems with a special focus on production assessment. In this sense, the methodology addresses the major components of the assistance system architecture. 
We propose a novel architecture based on the evidence theory that can combine $n$ number of information sources while quantifying the uncertainty of the resulting system prediction. For this purpose, we provide a detailed description of the architecture in terms of components and their relationships, with a special focus on the role of uncertainty.


\subsection{Information Fusion} 
Information fusion is a popular approach to combining several sources of information because the combined system often yields better performance and robustness.
Information fusion on the decision level is a common practice using data-based models (e.g., supervised classifiers in the case of bagging) \cite{UseyaChen2018}. 
The use of information fusion and data-based models is reported in \cite{ZhaoShi2023}\cite{NeagoeGhenea2022}, in which evidence theory combines models at the decision level.
Information fusion using evidence theory provides an additional feature: the uncertainty quantification \cite{ArevaloPiolo2022}. The uncertainty serves to assess the output reliability of the combined system \cite{DeVilliersLaskey2015}. 
Alternatively, knowledge-based models are expert-centered approaches containing valuable expert domain and environment context \cite{GirardiKueng2015}. 
Different knowledge-based approaches can be found in the literature using case-based reasoning (CBR) and natural language processing (NLP) \cite{RahmGraube2018}, ontologies, and assistance systems \cite{LohfinkAnton2022}.  
Though combining the strength of data-based and knowledge-based models might be considered a logical step to follow, finding a common framework to perform the fusion is challenging. Besides, knowledge-based models often have a low number of input features in comparison with data-based models. The last aspect requires special attention while performing an inference of the primary systems before performing an information fusion.
Current research methodologies cover the information fusion of data-based models \cite{ZhaoShi2023}\cite{NeagoeGhenea2022}. 
However, existing literature does not report the fusion of data-based and knowledge-based models, though the heterogeneity of the sources could improve the overall result.
We propose a methodology for the information fusion of a data-based model with an expert-centered model, in which we use the Dempster-Shafer evidence theory as a general framework for the fusion. Besides, we test the feasibility of the methodology using data from an industrial setup.




\subsection{Update of Data-based Models} 
The ability of data-based models to handle data with unknown fault cases has grown interest in the research community \cite{MukaiKumano2022} \cite{GawlikowskiSaha2021}. 
A primary step is identifying the unknown fault case or anomaly from the upcoming data. There are different approaches reported in the literature to detect anomalies, which propose the use of evidence theory \cite{ArevaloIbrahim2023} and unsupervised learning \cite{ChadhaIslam2021} \cite{RatoReis2013}.
After identifying the anomaly from the data, the next step is updating the model. In this sense, some methodologies are focus on concept drift detection \cite{OkawaKobayashi2021} \cite{KidaneTownend2022}, incremental learning \cite{ZhangWang2022} \cite{YangSun2021}, emerging classes or labels \cite{LiHuang2020} \cite{KongsorotHorata2020} \cite{WeiYe2021}, and incremental class \cite{KongsorotHorata2020}.
Thus, detecting an anomaly is followed by an update or retraining of the data-based model.
However, there are challenges associated with the retraining or updating of models: the size of the training data sufficient to capture the essence of the upcoming fault.
An essential factor to consider is the performance evaluation of the retrained models. A careful study of the parameters is required because only some upcoming faults might be handled with the same set of retraining parameters.
Existing literature addresses the anomaly detection \cite{ArevaloIbrahim2023} \cite{ChadhaIslam2021} \cite{RatoReis2013}, and even the identification of emerging classes (or unknown conditions) \cite{LiHuang2020} \cite{KongsorotHorata2020} \cite{WeiYe2021}.
However, the model update using uncertainty remains unexplored. To this end, we propose a methodology for updating data-based models using DSET, in which we monitor the uncertainty of the fusion to trigger a model update.
We focus on the model update of data-based models, specifically for ensemble classification using evidence theory.    
Besides, we perform an ablation study of the retraining parameters while showing their impact on the model performance. We demonstrate the robustness of the model update using the benchmark Tennessee Eastman.

\section{Theoretical Background}\label{theoretical}
This section presents the basic theory for performing information fusion and the transformation of model predictions using an evidential treatment. The equations of this section are applied during the development of the sections \ref{section__information__fusion} and \ref{section__model__update}. 

\subsection{Evidence Theory}\label{evidence_theory}
Dempster-Shafer \cite{Shafer1976} defined a frame of discernment $\Theta=\{A, B\}$ for the focal elements A and B. The power set $2^{\Theta}$ is defined by $2^{\Theta} = \{ \phi, \{A\},\{B\}, \Theta \} \}$.
The definition of a basic probability assignment (BPA) is given by: m: $2^{\Theta} \rightarrow [0,1]$, in which the BPA must comply with $m(\phi) = 0$, and $\sum_{A \subseteq \Theta} m(A) = 1$. The last equation represents the sum of BPAs.
The focal elements of $\Theta$ are mutually exclusive: $A \cap B = \phi$.

The \textit{Dempster-Shafer rule of combination} (DSRC) defines how to perform the fusion of two mass functions (e.g., sources of information) using the equation: 
\begin{equation} \label{dempster__1}
\begin{split}
	m_{DS}(A) & = ( m_{1} \oplus m_{2} ) (A) \\
	& = \frac{\sum_{B \cap C = A \neq \phi} m_{1}(B) m_{2}(C)}{1 - \sum_{B \cap C =\phi} m_{1}(B) m_{2}(C)} 
\end{split}
\end{equation}
where $m_{DS}(A)$ is the fusion of the mass functions $m_{1}$ and $m_{2}$. The conflicting evidence $b_{k}$ is defined by: 
\begin{equation} \label{dempster__2}
\begin{split}
b_{k} = \sum_{B \cap C=\phi} m_{1}(B)  m_{2}(C)
\end{split}
\end{equation}
It is important to remark that, while using DSRC, the conflicting evidence is distributed by each focal element.

Yager \cite{Yager1987} defined an alternative rule of combination, which in contrast to DSRC, assigns the conflicting evidence to the focal element $\Theta$. The \textit{Yager rule of combination} (YRC) is defined by the equation: 
\begin{equation} \label{yager__1}
\begin{split}
m_{Y}(A) = \sum_{B \cap C \neq \phi}^{}m_{1}(B) m_{2}(C)
\end{split}
\end{equation}
where $m_{Y}(A)$ is the fusion of the mass functions $m_{1}(B)$ and $m_{1}(C)$. The focal element $\Theta$ of the mass function $m_{Y}(A)$ is defined by: $m_{Y}(\theta) = q(\theta) + q(\phi)$, where $q(\phi)$ represents the conflicting evidence.
Likewise DSRC, the conflicting evidence $q(\phi)$ is represented by: 
\begin{equation} \label{yager__2}
\begin{split}
q(\phi) =  \sum_{B\bigcap C=\phi}^{}m_{1}(B) m_{2}(C)\\
\end{split}
\end{equation}

In the case of multiple fusion operations, the mass functions are combined using the following equation:
\begin{equation} \label{dempster__3}
\begin{split}
m(A) = \bigg( \Big( m_{1} \oplus m_{2} \Big) ... \oplus m_{N} \bigg)(A)
\end{split}
\end{equation}
where $m(A)$ is the fusion of the $n$ mass functions, and $N \in \mathbb{N}$.

\subsection{Evidential Treatment of Model Predictions}
We consider models with a common frame of discernment $\Theta = \{L_{1}, L_{2},..., L_{N} \}$, where $N$ represents the number of labels or classes, , and $N \in \mathbb{N}$. The power set is represented by $2^{\Theta} = \{\phi, \{L_{1}\}, \{L_{2}\}, \{L_{1},L_{2}\}$. The last term represents the overall uncertainty $U$.
Each model (e.g., classifier or a rule-based system) provides a prediction in the form of a unique label $p=L_{1}$ or as an array, $p = [L_{1}, L_{2},..., L_{n} ]$.
In section \ref{evidence_theory}, the sum of BPAs is defined as $\sum_{A \subseteq \Theta} m(A) = 1$.  
In \cite{ArevaloPiolo2022}, we proposed a strategy to transform a prediction into a mass function. This operation plays an essential role in the fusion of different information sources.
We presented a sum of BPA that considers the weights of each focal element $w_{m}$, and the quantification of the overall uncertainty $U$: $S_{wbpa} = \sum_{j=1}^{N} m_{j} \cdot w_{m_{j}} + U = 1$, where $n \in \mathbb{N}$ and $w_{m}$ is the weight of the evidence $m$. The following conditions must be fulfilled: $ \forall m_{j}. \quad  m_{j}> 0$ and $w_{m_{j}} \rightarrow [0,1]$.
The overall uncertainty is defined as $U = 1- \sum_{j=1}^{N} m_{j} \cdot w_{m_{j}}$, in which a high value of $U$ represents a high uncertainty on the body of evidence (e.g., lack of evidence).
We consider that the focal elements are mutually exclusive, which means that only one label is active at the time, which transforms $S_{wbpa}$ into $S_{wbpa} = m_{R_{j}} \cdot w_{m_{R_{j}}} + U = 1$. However, we adapted the \textit{sensitivity to zero} approach of Cheng et al.\cite{Cheng1988}, using the equation \cite{ArevaloNguyen2017}: $k = 1-10^{-F}$, where $k \in \mathbb{R}$, $F \in \mathbb{N}$, and $F \gg 1$.
Thus, we transform $S_{wbpa}$ into:
\begin{equation} \label{equation__DSET__1}
\begin{split}
S_{awbpa} & = \sum_{j=1}^{N} m'_{p_{j}} \cdot w_{m_{p_{j}}} + U = 1\\
\end{split}
\end{equation}
where $m'_{p_{j}}$ represents the $j_{th}$ focal element, and is defined using:
\begin{equation} \label{equation__DSET__2}
     m'_{p_{j}}=\begin{dcases}
         k& \text{if}\ p_{j} = True \\
         \frac{1-k}{N-1}& \text{otherwise}  \\
         \end{dcases}
\end{equation}
where $k$ is the approximation factor, $N$ is the number of focal elements of $\Theta$, and $N \in \mathbb{N}$.
The active prediction $p$ can be transformed into a mass function $m$ using: $\mathbf{m} = \mathbf{m'_{p}}.\mathbf{w_{p}}$.
The mass function can be represented as a row vector using the following equation:
      \begin{equation} \label{equation__DSET__3}
      \begin{split}
    	m = [m'_{p_{1}} \cdot w_{p_{1}} \quad    
  		... \quad 
  		m'_{p_{N}} \cdot w_{p_{N}}
  		\quad U]      
      \end{split}
   \end{equation}
and the uncertainty $U$ is defined as: 
      \begin{equation} \label{equation__DSET__4}
      \begin{split}
    	U = 1 - \sum_{j=1}^{N} m'_{p_{j}} \cdot w_{m_{p_{j}}}
  		\quad U]      
      \end{split}
   \end{equation}

\section{INFUSION: Information Fusion for Assistance Systems in Production Assessment}\label{section__infusion}

This research proposes an INformation FUsion approach for asSIstance systems in productiON assessment (INFUSION). 
This section covers the topics: theoretical background, prediction systems, information fusion, model update of the prediction system, and the assistance system.

\begin{figure*}[!ht]
	\centering
	\includegraphics[width=0.98\textwidth,keepaspectratio]{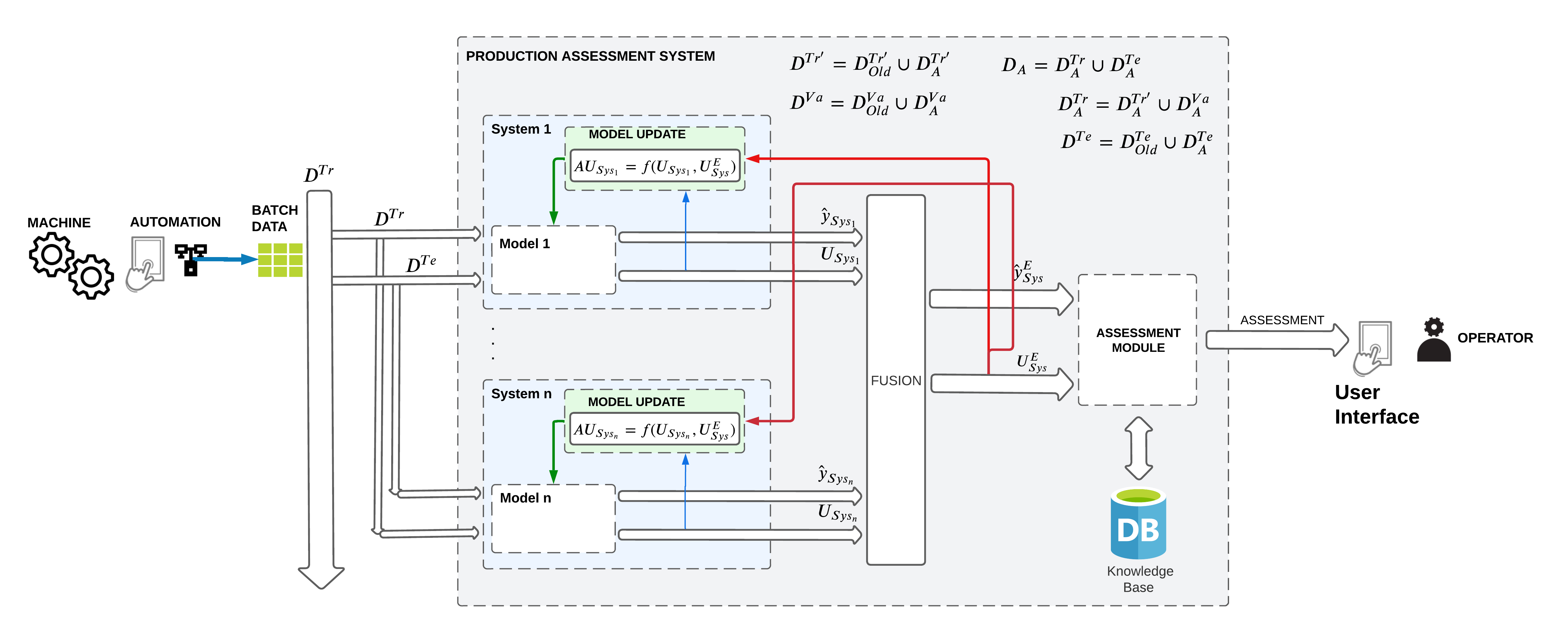}
	\caption{General system overview. }\label{figure__F1__general__overview}
\end{figure*}

As a first insight into this theme, we present a general system overview as seen in Fig. \ref{figure__F1__general__overview}. 

The general system is conformed by $n$ systems used as information sources. The motivation behind this is the creation of a more robust system. 
The general system overview is composed of the blocks:

\begin{itemize}
    \item The batch data is the numerical representation of the physical behavior of a machine. The data is split in three categories: training data $D^{Tr}$, validation data $D^{Va}$, and testing data $D^{Te}$. The data is used during the training and inference processes of the models.
    \item The modules form the production assessment system:
    \begin{itemize}
        \item $n$ Systems, in which each system has a model and a model update module. For instance, model 1 has two outputs: the model prediction $\hat{y}_{Sys_{1}}$ and its associated uncertainty $U_{Sys_{1}}$.
        \item The fusion module, which combines the predictions $\hat{y}_{Sys_{1}}$..$\hat{y}_{Sys_{n}}$ of the information sources \textit{model 1}..\textit{model $n$} into the ensemble prediction $\hat{y}^{E}_{Sys}$.
        \item The model update module is triggered either by each system uncertainty (e.g., $U_{Sys_{1}}$) or by the ensemble uncertainty $U^{E}_{Sys}$.
        \item The assessment module matches each ensemble prediction with its corresponding assessment.
        \item The knowledge base has the assessment for each ensemble prediction.
    \end{itemize}    
    \item The assessment is presented to the user (operator) through a user interface.
\end{itemize}

A primary motivation of this paper is the integration of data-based and knowledge-based models because the combined outcome profits from the strengths of both models.
Therefore, the $n$ systems of Fig. \ref{figure__F1__general__overview} are transformed into two major systems: an ensemble classifier (EC) that groups different data-based models and a knowledge-based model. Section \ref{prediction_system} details both systems. 


\subsection{Prediction Systems}\label{prediction_system}
As presented in Fig. \ref{figure__F1__general__overview}, a (prediction) system is conformed by an inference model and an update module. The trained model represents the physical system and is used to predict the system's answer while feeding data to it. The inference model can be data-based (e.g., a supervised classifier), an ensemble classifier (EC) formed by several models, a model built on equations representing the physical system, an ontology, or a knowledge-based model. The model update module adapts the system when the initial conditions have changed (or unknown events occur). The update is performed automatically or manually, depending on the module strategy. 

A model $M_{i}$ is trained using a training dataset $D^{Tr}$ (in the case of data-based models), or is modeled using the relationships between the process variables and thresholds (in the case of a knowledge-based model). A training dataset $D^{Tr}$ contains $N_{o_{Tr}}$ number of observations, $N_{f_{Tr}}$ number of features, and $N_{c_{Tr}}$ number of classes. A frame of discernment $\Theta$ is formed by all the labels (or classes) that the model can predict: $\Theta=\{C_{1},..., C_{N} \}$, where $N \in \mathbb{N}$.

Thus, a model $M_{i}$ outputs the prediction $\hat{y_{i}}$ while feeding the testing data $D^{Te}$:
\begin{equation} \label{prediction__1}
\begin{split}
    \hat{y_{i}} = M_{i}(D^{Te})\\
\end{split}
\end{equation}
where $\hat{y_{i}} \in \Theta$.
The prediction $\hat{y_{i}}$ is transformed into the mass function $m_{i}$ using equations (\ref{equation__DSET__1})-(\ref{equation__DSET__4}):
\begin{equation} \label{prediction__2}
\begin{split}
    m_{i} = f_{m}(\hat{y_{i}},w_{M_{i}})\\
\end{split}
\end{equation}
where $w_{M_{i}}$ represents the (confidence) weights for each class predicted by the model $M_{i}$. 

\begin{figure*}[!ht]
	\centering
	\includegraphics[width=0.98\textwidth,keepaspectratio]{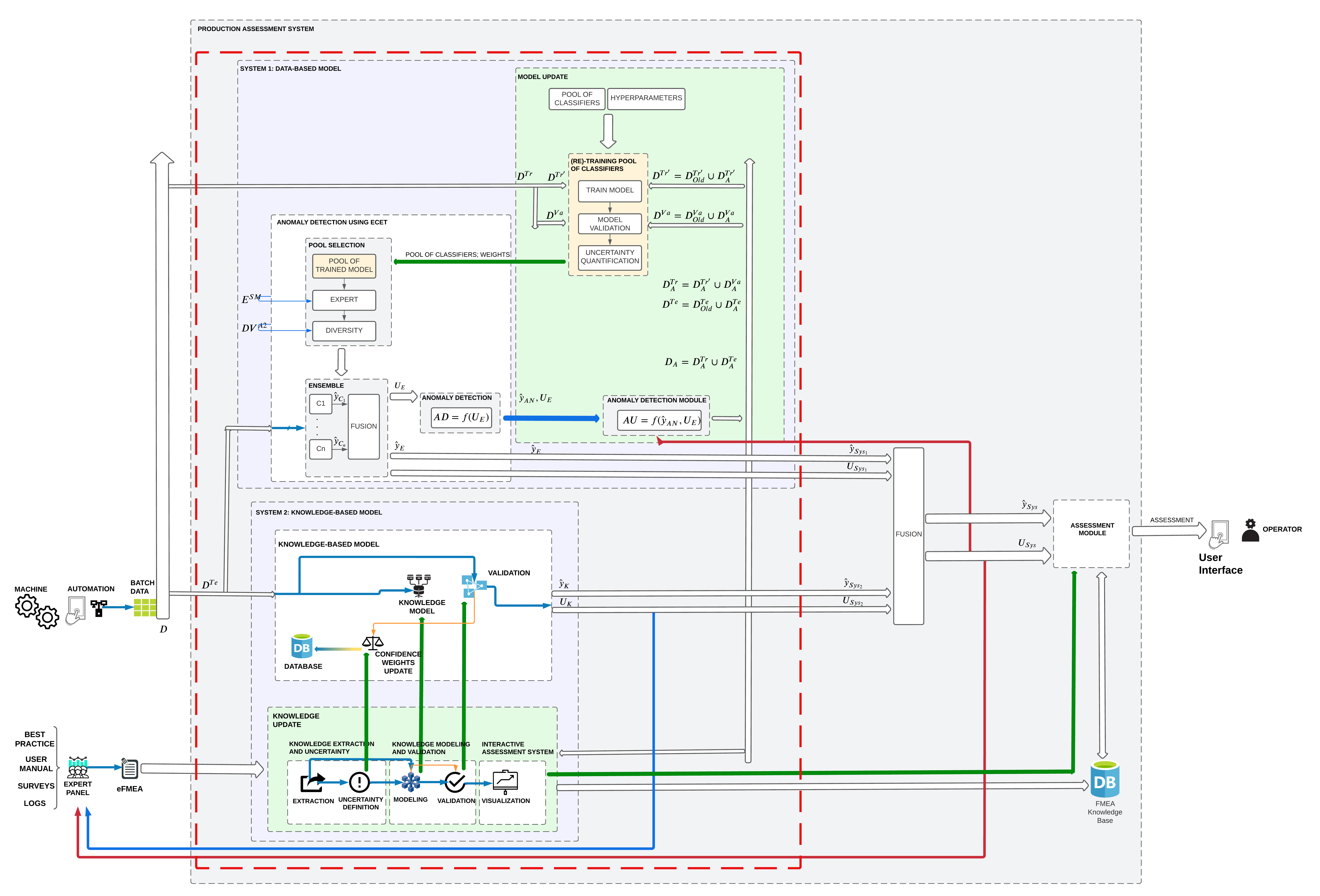}
	\caption{INFUSION overview. }\label{figure__F2__infusion__overview}
\end{figure*}

We focus this research on a prediction system using EC and rule-based knowledge models. Previous research deepened in these two topics separately \cite{ArevaloIbrahim2023} \cite{ArevaloPiolo2022}. 
Fig. \ref{figure__F2__infusion__overview} details the INFUSION system, where the prediction systems are adjusted to a data-based and knowledge-based model. Thus, the data-based model is represented by the EC using the ensemble classification and evidence theory (ECET) approach \cite{ArevaloIbrahim2023}, and the knowledge-based model is built using the knowledge transfer framework and evidence theory (KLAFATE) methodology \cite{ArevaloPiolo2022}. It is important to remark that each system has an inference model and a model update module.  
It is important to note that ECET is an EC formed by $n$ systems, specifically the $n$ supervised classifiers. ECET presents a similar structure from Fig. \ref{figure__F1__general__overview} for the system's prediction, except for the model update module.

The model update module of KLAFATE is manual because it relies on the expertise of the team expert. The methodology is explained in detail in \cite{ArevaloPiolo2022}. 
The automatic model update module of ECET is introduced in this research and is explored in detail in section \ref{section__model__update}. The main blocks of this module are:

\begin{itemize}
    \item The pool of classifiers and the list of hyperparameters reported in \cite{ArevaloIbrahim2023}.
    \item The (re)-training pool of classifiers module, which is formed by the blocks:
    \begin{itemize}
        \item train model using either the prior training data $D^{Tr}$, or using the re-training data $D^{Tr'}$.
        \item model validation either the prior validation data $D^{Va}$, or the new validation data $D^{Va'}$.
        \item uncertainty quantification
    \end{itemize}
    \item The anomaly detection module which monitors the ensemble uncertainty $U_{E}$ and the anomaly prediction $\hat{y}_{AN}$ of ECET, and the system uncertainty $U_{Sys}$ and the system prediction $\hat{y}_{Sys}$.    
\end{itemize}

\subsubsection{ECET Prediction System}
In \cite{ArevaloIbrahim2023}, we presented an approach of \textit{ensemble classification using evidence theory} (ECET), in which we propose the use of information fusion to combine the predictions of $N$ number of classifiers. In this paper, we extend the contribution of \cite{ArevaloIbrahim2023} by formalizing the approach theoretically. This theoretical formalization plays a crucial role in section \ref{section__information__fusion} and section \ref{section__model__update}, which correspond to the methodologies of information fusion and model update, respectively.
Thus, given a $n$ number of classifiers, each classifier produces an output $\hat{y_{i}}$ using equation (\ref{prediction__1}), where $\hat{y_{i}} \in \Theta$. The output is subsequently transformed into a mass function $m_{i}$ using equations (\ref{equation__DSET__1})-(\ref{equation__DSET__3}). 
The ensemble classifier (EC) is obtained by combining all the classifiers, specifically using the DSRC on the mass function of each classifier prediction. 
As described in equation (\ref{dempster__3}), the DSRC can be used for multiple fusion operations. However, the fusion is performed in pairs. For instance, in the case of three classifiers, the fusion of $m_{1}$ (corresponding to the output $\hat{y_{1}}$ of model $C_{1}$) and $m_{2}$ is performed first, the result of this fusion $m_{1} \oplus m_{2}$ is then combined with $m_{3}$.
The fusion of the pair of mass functions $m_{i}$ and $m_{D_{i-1}}$ is represented using:
\begin{equation} \label{ecet__1}
     F_{D_{i}}=\begin{dcases}
         m_{i} \oplus m_{D_{i-1}} & \text{if}\ i > 1 \\
         0 & \text{otherwise}  \\
         \end{dcases}
\end{equation}
where $i \in \mathbb{N}$, $m_{i}$ is the mass function of the current classifier, and $m_{D_{i-1}}$ is the fusion of the previous mass functions.
After obtaining the resulting fusion $F_{D_{i}}$, the previous mass function $m_{D_{i-1}}$ is updated:
\begin{equation} \label{ecet__2}
     m_{D_{i-1}}=\begin{dcases}
         F_{D_{i}} & \text{if}\ i > 1 \\
         m_{i} & \text{otherwise}  \\
         \end{dcases}
\end{equation}
where $i \in \mathbb{N}$. The last element of the fusion $F_{D_{i}}$, which is a row vector, corresponds to the uncertainty $U_{D_{i}}$: 
\begin{equation} \label{ecet__3}
     U_{D_{i}}=\begin{dcases}
         F_{D_{i}}[N] & \text{if}\ i > 1 \\
         0 & \text{otherwise}  \\
         \end{dcases}
\end{equation}
, where $N$ is the cardinality of the frame of discernment $\Theta$, and $N \in \mathbb{N}$. 
After performing the last fusion, the system prediction $\hat{y}_{EN}$ is calculated using:
  \begin{equation} \label{ecet__4}
  	\begin{split}
  		\hat{y}_{EC} = \operatorname*{arg\,max}_\Theta F_{D_{i}}
  	\end{split}
  \end{equation}
where $\hat{y}_{EC} \in \Theta$.
The system uncertainty is calculated using: $U_{D} = U_{D_{i}}$. 
A similar procedure is performed when using the YRC to calculate the fusion $F_{Y_{i}}$, the previous mass function $m_{D_{i-1}}$, and the uncertainty $U_{Y_{i}}$. It is important to remark that the current mass function $m_{i}$ is used for DSRC and YRC.

\subsubsection{KLAFATE Prediction System}
In \cite{ArevaloPiolo2022} we presented a knowledge-based model using the \textit{knowledge transfer framework using evidence theory} (KLAFATE)
\cite{ArevaloPiolo2022}. The knowledge was extracted from a failure mode and effects analysis (FMEA) and modeled in rules. Thus, a knowledge rule $R_{i}$ is defined as the function:
$R_{i} = f(V_{1},...,V_{N_{V}}, T_{1},...,T_{N_{T}})$, where 
$V_{1}$ represents a process variable, $T_{1}$ is a threshold or limit value of the process value, $N_{V}$ is the number of process variables, $N_{T}$ is the number of thresholds, $N_{V}$ and $N_{T}$ $\in \mathbb{N}$.
The knowledge rules are mutually exclusive: $R_{i} \cap R_{i+1} = \phi$. The knowledge model is represented as a set of rules \cite{ArevaloPiolo2022}:
\begin{equation} \label{kext__1}
    \begin{split}
        L_{T_{R}}=\begin{dcases}
         L_{T_{R_{1}}}  & \text{if} \quad R_{1}  \\
         ...\\
         L_{T_{R_{m}}}  & \text{if} \quad R_{m}  \\
         L_{T_{R_{m+1}}} & \text{otherwise}  \\
         \end{dcases}
    \end{split}
\end{equation}
where $L_{T_{R_{i}}}$ represents the approximated rule $R_{i}$, $m$ is the number of knowledge rules, $m \in \mathbb{N}$, and $L_{T_{R}}$, $R_{i} \in \Theta$.
The active rule is obtained using equations (\ref{equation__DSET__1})-(\ref{equation__DSET__4}): 
\begin{equation*} \label{kext__2}
     L_{T_{R_{i}}}=\begin{dcases}
         k & \text{if}\ R_{i} = \text{True} \\
         \frac{1-k}{N-1} & \text{otherwise}  \\
         \end{dcases}
\end{equation*}
where $k$ is the approximation factor, $N$ is the cardinality of $\Theta$, $k \in \mathbb{R}$, and $N \in \mathbb{N}$.
Thus, the mass function is defined using equation (\ref{equation__DSET__3}):
      \begin{equation} \label{kext__3}
      \begin{split}
    	m = [L_{T_{R_{1}}} \cdot w_{R_{1}} \quad    
  		... \quad 
  		L_{T_{R_{N}}} \cdot w_{R_{N}}
  		\quad U]      
      \end{split}
   \end{equation}
where $w_{R_{1}}$ is the (confidence) weight of the rule $R_{1}$, and $U$ is the overall uncertainty.
The uncertainty $U$ is calculated using the equation (\ref{equation__DSET__4}):
      \begin{equation} \label{kext__5}
      \begin{split}
    	U = 1 - \sum_{j=1}^{N} L_{T_{R_{j}}} \cdot w_{R_{j}}      
      \end{split}
   \end{equation}
The (confidence) weight $w_{R_{j}}$ is defined using the equation \cite{ArevaloPiolo2022}:
\begin{equation*} \label{kext__4}
\begin{split}
w_{R_{j}} = \frac{1}{N_{R}}\sum_{i=1}^{N_{R}} w_{R_{C_{i}}}(V,T)
\end{split}
\end{equation*}

The mass function $m_{R_{i}}$ is transformed into the prediction $\hat{y}_{KE}$ using:
  \begin{equation} \label{kext__6}
  	\begin{split}
  		\hat{y}_{KE} = \operatorname*{arg\,max}_\Theta m_{R_{j}}
  	\end{split}
  \end{equation}
where $\hat{y}_{KE} \in \Theta$.

\subsection{Assistance System}
The assistance system provides an interactive source of assessment for the user while receiving the process data. It provides the current status of the system (e.g., system prediction and uncertainty), the assessment (e.g., troubleshooting through the FMEA knowledge base) in the case of a fault case, and notifies in case of an unknown condition for the consequent model update. 

The knowledge of the FMEA is stored as a knowledge tuple $TU_{i}$ \cite{ArevaloPiolo2022}: 
 \begin{equation} \label{assessment__1}
\begin{split}
TU_{i} = (P, SP, FM, \textbf{C}, \textbf{E}, \mathbf{RE}, R, w_{R})
\end{split}
\end{equation}
where $FM$ represents a failure mode, $P$ is a process, $SP$ a subprocess, $\textbf{C}$ a set of causes, $\textbf{E}$ a set of effects, $\mathbf{RE}$ a set of recommendations, and $i \in \mathbb{N}$. 
A set of recommendation is also represented as: $\mathbf{RE} = [RE_{1},...,RE_{N_{RE}}]$, where $N_{RE} \in \mathbb{N}$. The latest representation applies to the sets of effects and causes.

In the assessment context, the rule $R$ corresponds to the system prediction $\hat{y}_{Sys}$, and the confidence weight $w_{R}$ to the system weight $w_{\hat{y}_{Sys}}$, where $R, \hat{y}_{Sys} \in \Theta_{Sys}$, and $w_{Sys}=1$.
It is important to remark, that each system prediction $\hat{y}_{Sys}$ is linked to a knowledge tuple $TU{i}$, a failure mode $FM$, and to a weight $w_{Sys}$: $\hat{y}_{Sys} \iff TU_{i}$, $\hat{y}_{Sys} \iff FM$, and $\hat{y}_{Sys} \iff w_{\hat{y}_{Sys}}$. In contrast, a system prediction $\hat{y}_{Sys}$ can be associated to a set of causes $\textbf{C}$, effects $\textbf{E}$, and recommendations $\textbf{RE}$.
The assessment module is modeled through a matching function that associates a system prediction $\hat{y}_{Sys}$ to the rest of the knowledge of the tuple $TU_{i}$:
 \begin{equation} \label{assessment__2}
\begin{split}
P, SP, FM, \textbf{C}, \textbf{E}, \mathbf{RE} = f_{Ma}(\hat{y}_{Sys}, TU)
\end{split}
\end{equation}
where $i \in \mathbb{N}$.
The matching function $f_{Ma}$ provides the assessment while feeding the system prediction $\hat{y}_{Sys}$, specifically returning the troubleshooting information associated with the failure mode: the process $P$, the subprocess $SP$, the set of causes $\textbf{C}$, the set of effects $\textbf{E}$, and the set of recommendations $\mathbf{RE}$.
The assistance system was described in detail in a previous work \cite{ArevaloPiolo2022}.

\subsection{Information fusion}\label{section__information__fusion}
Information fusion has a growing research interest because it improves robustness while combining different models. 
To this end, we propose a novel framework for combining $n$ number of models using DSET. Moreover, this framework is used for the fusion of a data-based model and a knowledge-based model.

Thus, as presented in Fig. \ref{figure__F1__general__overview}, the system is formed by $n$ number of subsystems. The system mass function $m_{Sys}$ is obtained after applying the information fusion to all subsystems:

\begin{equation} \label{information_fusion__1}
\begin{split}
m_{Sys}(A) = \bigg( \Big( m_{Sys_{1}} \oplus m_{Sys_{2}} \Big) ... \oplus m_{Sys_{n}} \bigg)(A)
\end{split}
\end{equation}

where $n \in\mathbb{N}$, and $m_{Sys}(A) \in \Theta_{Sys}$. The system mass function $m_{Sys}$ is also referred as $F_{Sys}$.
It is important to remark that all the systems share the same frame of discernment: $\Theta_{KE} = \Theta_{EC} = \Theta_{Sys}$, and  
\begin{equation} \label{information_fusion__3}
\begin{split}
    \Theta_{Sys}=\{C_{1},...,C_{N_{Sys}}\}
\end{split}
\end{equation}
where $C_{1}$ represents the first class (or fault case), $N_{Sys}$ is the number of classes (or fault cases), and $N_{Sys} \in \mathbb{N}$.  

The equation (\ref{information_fusion__1}) can also be represented as:

\begin{equation} \label{information_fusion__4}
     m_{Sys}(A)=\begin{dcases}
         (\bigoplus_{i}^{N_{Sys}} m_{Sys_{i}})(A) & \text{if}\ i > 1 \\
         0 & \text{otherwise}  \\
         \end{dcases}
\end{equation}

where $i, N_{Sys} \in \mathbb{N}$.

This paper adapts the system to two main subsystems: a data-based model $M_{EC}$ and a knowledge-based model $M_{KE}$. 

As a first step we obtain the outputs $\hat{y_{EC}}$ and $\hat{y_{EC}}$ by feeding data to the models $M_{KE}$ and $M_{EC}$:
\begin{equation} \label{information_fusion__5}
\begin{split}
    \hat{y_{EC}} = M_{EC}(D^{Te})\\
\end{split}
\end{equation}
and
\begin{equation} \label{information_fusion__6}
\begin{split}
    \hat{y_{KE}} = M_{KE}(D^{Te})\\
\end{split}
\end{equation}
where $D^{Te}$ is the testing data.

The predictions $\hat{y_{EC}}$ and $\hat{y_{KE}}$ are transformed into the mass functions $m_{EC}$ and $m_{KE}$ respectively, using equations (\ref{equation__DSET__1})-(\ref{equation__DSET__4}):
\begin{equation} \label{information_fusion__7}
\begin{split}
    m_{EC} = f_{m}(\hat{y_{EC}},w_{M_{i}})\\
\end{split}
\end{equation}
and
\begin{equation} \label{information_fusion__8}
\begin{split}
    m_{KE} = f_{m}(\hat{y_{KE}},w_{M_{i}})\\
\end{split}
\end{equation}
where $w_{M_{i}}= 1$ $\forall i$, and $i \in \mathbb{N}$. 

The next step is to obtain the system fusion $F_{Sys}$ by applying either DSRC or YRC.

Thus, the system fusion $F_{D_{Sys}}$ is calculated using DSRC and applying the equations (\ref{dempster__1}), (\ref{dempster__2}), (\ref{information_fusion__1}), (\ref{information_fusion__4}):
\begin{equation} \label{information_fusion__9}
\begin{split}
F_{D_{Sys}}(A) & = (\bigoplus_{i}^{N_{Sys}} m_{Sys_{i}})(A)\\
            & = (m_{Sys_{1}} \oplus m_{Sys_{2}})(A)\\
            & = (m_{EC} \oplus m_{KE})(A)\\
\end{split}
\end{equation}
Likewise, the system fusion $F_{Y_{Sys}}$ is calculated using YRC and applying the equations (\ref{yager__1}), (\ref{yager__2}), (\ref{information_fusion__1}), (\ref{information_fusion__4}): 
\begin{equation} \label{information_fusion__9a}
\begin{split}
F_{Y_{Sys}} = (m_{EC} \oplus m_{KE})(A)
\end{split}
\end{equation}.
The system uncertainty $U_{D}$ is calculated using the last DSRC fusion $F_{D_{i}}$:
$\hat{y}_{Sys}$ using:
  \begin{equation} \label{information_fusion__9b}
  	\begin{split}
  		U_{D_{i}}=F_{D_{i}}[|\Theta_{Sys}|]
  	\end{split}
  \end{equation}
where $F_{D_{i}}[|\Theta_{Sys}|]$ corresponds to the overall uncertainty of the system fusion $F_{D_{i}}$.
Likewise, the system uncertainty $U_{Y}$ is calculated using the last YRC fusion $F_{Y_{i}}$: 
  \begin{equation} \label{information_fusion__9c}
  	\begin{split}
  		U_{Y_{i}}=F_{Y_{i}}[|\Theta_{Sys}|]
  	\end{split}
  \end{equation}
where $F_{Y_{i}}[|\Theta_{Sys}|]$ corresponds to the overall uncertainty of the system fusion $F_{Y_{i}}$.

The last step is the calculation of the system mass function $m_{Sys}$ and the system uncertainties using DSRC $U_{D}$ and YRC $U_{Y}$.
The system mass function $m_{Sys}$ is obtained from the last DSRC system fusion $F_{D_{Sys}}$: $m_{Sys}=F_{D_{i}}$.
The mass function $m_{Sys}$, then, is transformed into the prediction $\hat{y}_{Sys}$ using:
  \begin{equation} \label{information_fusion__10}
  	\begin{split}
  		\hat{y}_{Sys} = \operatorname*{arg\,max}_\Theta m_{Sys}
  	\end{split}
  \end{equation}
where $\hat{y}_{Sys} \in \Theta_{Sys}$. Algorithm \ref{algorithm__fusion} describes the steps for the information fusion of $N_{Sys}$ number of subsystems while feeding the testing data $D^{Te}$, where $N_{Sys} \in \mathbb{N}$. Algorithm \ref{algorithm__fusion} is an updated version of the algorithm presented in \cite{ArevaloIbrahim2023}.

\begin{algorithm}[!ht]
\caption{Information Fusion of $N_{Sys}$ Systems \cite{ArevaloIbrahim2023} 
}\label{algorithm__fusion}
\begin{algorithmic}[1]
\Procedure{Information Fusion}{}
\For{$j=1$ to $N_{Sys}$} \Comment $N_{Sys}$ Subsystems
    \For{$i=1$ to $N_{D^{Te}}$} \Comment $N_{D^{Te}}$ Samples
        \State $\hat{y_{i}} \gets M_{j}(S_{i})$ \Comment by Eq. (\ref{information_fusion__5})
        \State $m_{i} \gets f_{m}(\hat{y_{i}},w^{M_{j}}_{i})$ \Comment by Eq.(\ref{equation__DSET__1})-(\ref{equation__DSET__4}), (\ref{information_fusion__7})
        \If{i = 1}
            \State $F_{D_{i-1}} = F_{Y_{i-1}} = 0$
            \State $m_{D_{i-1}} = m_{Y_{i-1}} = m_{i}$
            \State $U_{D_{i-1}} = U_{Y_{i-1}} = 0$
        \Else 
            \State $F_{D_{i}} = m_{i} \oplus m_{D_{i-1}}$ \Comment by Eq. (\ref{information_fusion__9}) 
            \State $F_{Y_{i}} = m_{i} \oplus m_{Y_{i-1}}$ \Comment by Eq.(\ref{information_fusion__9a}) 
            \State $m_{D_{i-1}}=F_{D_{i}}$
            \State $m_{Y_{i-1}}=F_{Y_{i}}$
            \State $U_{D_{i}}=F_{D_{i}}[|\Theta_{Sys}|]$ \Comment by Eq.(\ref{information_fusion__9b})
            \State $U_{Y_{i}}=F_{Y_{i}}[|\Theta_{Sys}|]$ \Comment by Eq.(\ref{information_fusion__9c})
        \EndIf
    \EndFor
\EndFor
    \State $m_{Sys}=F_{D_{i}}$
    \State $\hat{y}_{Sys} = \operatorname*{arg\,max}_\Theta m_{Sys}$ \Comment by Eq.(\ref{information_fusion__10})
    \State $U_{D} \gets U_{D_{i}}$ 
    \State $U_{Y} \gets U_{Y_{i}}$ 
\State \textbf{return $\hat{y_{Sys}}$, $U_{D}$, $U_{Y}$} 
\EndProcedure
\end{algorithmic}
\end{algorithm}	







\subsection{Model Update}\label{section__model__update}



The anomaly detection functionality is crucial in the model update because it identifies when an unknown condition is present. We present an (automatic) model update for ECET based on uncertainty monitoring. The (manual) model update of KEXT was proposed in \cite{ArevaloPiolo2022}. The model update is a sequence of five steps: anomaly detection, collection of unknown data, data isolation using a window, retraining, and inference. 

\subsubsection{Model Update for ECET}\label{model_update__ecet}

Performing ECs are usually the result of a suitable dataset that fits the patterns of the existing data. However, the occurrence of new unknown fault cases might undermine the performance of the ECs, leading to a retraining procedure of the models. 
To this end, our methodology provides the theoretical basis for updating the data-based models using DSET, in which we monitor the uncertainty of the fusion to trigger a model update.
The \textit{model update of ECET} is performed automatically using an anomaly detection strategy, in which the uncertainty is monitored. However, The model update can be set as semi-automatic (e.g., the user receives a notification from executing the model update module) in case the unknown condition needs to be analyzed in detail first.
Algorithm \ref{algorithm__retrain} describes the sequence of the model update.

\begin{algorithm}
\caption{Model Update of ECET.}\label{algorithm__retrain}
\begin{algorithmic}[1]
\Procedure{Model Update}{}         
        \State $\hat{y_{EC}} \gets M_{EC}(S_{j})$
        \State $m_{EC} \gets f_{m}(\hat{y_{EC}},w^{EC})$ \Comment by Eq.(\ref{equation__DSET__1})-(\ref{equation__DSET__4})
        \If{$C_{A} = True$} \Comment by Eq. (\ref{model_update__2})
            \State $\hat{y}_{A} = A_{K}$
            \State $D_{Temp_{j}} \gets collect\_data(X_{A}, \hat{y}_{A})$
            \State $i_{A} \gets i_{A} +1 $
            \If{$C_{S} = True$} \Comment by Eq. (\ref{model_update__5})
                \State $D_{A} \gets D_{Temp}$
                \State $D'^{Tr}_{A}, D^{Va}_{A}, D^{Te}_{A} \gets split\_data(D_{A})$
                \State $D'^{Tr} \gets D'^{Tr}_{Old} \cup D'^{Tr}_{A}$ \Comment by Eq. (\ref{model_update__7})
                \State $D^{Va} \gets D'^{Tr}_{Old} \cup D^{Va}_{A}$ \Comment by Eq. (\ref{model_update__8})
                \State $\hat{\textbf{M}}_{Tr} \gets retrain(\textbf{M}, D'^{Tr})$
                \State $\textbf{M}_{Tr} \gets \hat{\textbf{M}}_{Tr}$  \Comment Replace old models          
            \EndIf
        \Else
            \State $\hat{y}_{A} = \operatorname*{arg\,max}_\Theta m_{EC}$ \Comment by Eq.(\ref{information_fusion__10})
            \State $i_{A} \gets 0$
        \EndIf
\State  \textbf{return} $M_{Tr}$ 
\EndProcedure
\end{algorithmic}
\end{algorithm}

We proposed an \textit{anomaly detection} strategy using ECET in \cite{ArevaloIbrahim2023}, in which an unknown condition $A_{K}$ was detected:
\begin{equation} \label{model_update__1}
     \hat{y}_{A}=\begin{dcases}
         A_{K} & \text{if}\ C_{A} = True \\
         \hat{y}_{EC} & \text{otherwise}  \\
         \end{dcases}
\end{equation}
where $\hat{y}_{A}$ is a parallel prediction to the EC prediction $\hat{y}_{EC}$, $A_{K} \in \mathbb{Z}$, and $K \in \mathbb{N}$.
The condition for anomalies $C_{A}$ is defined as: 
\begin{equation} \label{model_update__1a}
\begin{split}
C_{A} = (U_{D} > Tr_{D_{Mx}})\ \text{and}\ (U_{Y} > Tr_{Y_{Mx}})\\
\end{split}
\end{equation}
where $U_{D}=b_{k}$, $U_{Y}=q(\phi)$, $Tr_{D_{Mx}}$ represents the maximum threshold for $U_{D}$, $Tr_{Y_{Mx}}$ is the maximum threshold for $U_{Y}$. The terms $b_{k}$ and $q(\phi)$ are calculated using the equations (\ref{dempster__1})-(\ref{dempster__2}), and (\ref{yager__1})-(\ref{yager__2}), respectively.

In this paper, we propose the monitoring of the EC uncertainties $U_{D_{EC}}$ and $Y_{D_{EC}}$, as well as the system uncertainties $U_{D_{Sys}}$ and $Y_{D_{Sys}}$. The condition for anomalies from equation (\ref{model_update__1a}) is transformed into:
\begin{equation} \label{model_update__2}
\begin{split}
C_{A} = C_{A_{EC}}\ or\ C_{A_{Sys}}\\
\end{split}
\end{equation}
where $C_{A_{EC}}$ and $C_{A_{Sys}}$ represent the condition for anomalies of EC and system, respectively. 
Thus, the anomaly detection of the system is defined as:
\begin{equation} \label{model_update__3}
     \hat{y}_{A_{Sys}}=\begin{dcases}
         A_{K} & \text{if}\ C_{A} = True \\
         \hat{y}_{Sys} & \text{otherwise}  \\
         \end{dcases}
\end{equation}

The \textit{data collection of (unknown) conditions} needs to satisfy the condition $C_{D}$: 
\begin{equation} \label{model_update__4}
\begin{split}
C_{D} = C_{A}\ and\ C_{S}\\
\end{split}
\end{equation}
where $C_{S}$ is the condition that satisfies a minimum number of consecutive data samples.
The condition $C_{S}$ is defined as: 
\begin{equation} \label{model_update__5}
\begin{split}
C_{S} = i_{A} > S_{Mn}\\
\end{split}
\end{equation}
where $i_{A}$ is the number of consecutive data samples, $S_{Mn}$ is the minimum number of consecutive data samples, and $i_{A}, S_{Mn} \in \mathbb{N}$.

The collected data of the unknown condition $D_{A}$ has the same features $f_{Tr}$ of the (old) original data $D$, such as $f_{A}=f_{Tr}$. In contrast, the number of observations $o_{A}$ might differ from that of the original data $o_{Tr}$.  
Thus, the data $D_{A}$ is represented by a number of observations $N_{o_{A}}$, in which each observation is composed by the features $X_{A}=f_{A}$ and the associated label (or class) $\hat{y}_{A}$.

The data $D_{A}$ is represented as: 
\begin{equation} \label{model_update__5a}
\begin{split}
X_{A_{S_{Mn} \times N_{f_{A}}}} \times Y_{A_{S_{Mn} \times 1}}\
\end{split}
\end{equation}
where $S_{Mn}$ is the minimum number of consecutive samples of the unknown condition, $N_{f_{A}}$ is the number of features, $S_{Mn}, N_{f_{A}} \in \mathbb{N}$, $X_{A} \in \mathbb{R}$, and $Y_{A} \in \mathbb{Z}$.



The data $D_{A}$ is split into training $D^{Tr}_{A}$ and testing data $D^{Te}_{A}$:
\begin{equation} \label{model_update__6}
     D_{A}=\begin{dcases}
         D^{Tr}_{A} \cup D^{Te}_{A} & \text{if}\ C_{D} = True\\
         0 & \text{otherwise}  \\
         \end{dcases}
\end{equation}

The training data is split $D^{Tr}_{A}$ into training data $D'^{Tr}_{A}$ and validation data $D^{Va}_{A}$: 
\begin{equation} \label{model_update__6a}
\begin{split}
D^{Tr}_{A} = D'^{Tr}_{A} \cup D^{Va}_{A}
\end{split}
\end{equation}

The next step is to integrate the existing data $D$ with the collected data $D_{A}$ using the following equations:
\begin{equation} \label{model_update__7}
\begin{split}
D'^{Tr}=D'^{Tr}_{Old} \cup D'^{Tr}_{A}
\end{split}
\end{equation}
\begin{equation} \label{model_update__8}
\begin{split}
D^{Va}=D^{Va}_{Old} \cup D^{Va}_{A}
\end{split}
\end{equation}
\begin{equation} \label{model_update__9}
\begin{split}
D^{Te}_{A} = D^{Te}_{Old} \cup D^{Te}_{A}
\end{split}
\end{equation}



The EC prediction $\hat{y_{EC}}$ usually has not a constant steady value because of the diversity of the classifier's predictions. For this reason, we propose a \textit{window} on the EC prediction $\hat{y_{EC}}$ that can ease the data isolation of the unknown condition. The window smoothes the EC output because it considers a window of $N_{w}$ number of the last samples for the calculation of the windowed EC output $\hat{y_{EC}}$:   

\begin{equation} \label{model_update__10}
\begin{split}
\hat{y}_{EC_{i}}^{w} = \frac{1}{N_{w}+1}\sum_{k=i-N_{w}}^{i} \hat{y}_{EC_{k}}\\
\end{split}
\end{equation}

where $\hat{y_{EC_{i}}}^{w} \in \Theta_{Sys}$.

\begin{figure}[!ht]
	\centering
	\includegraphics[width=0.48\textwidth,keepaspectratio]{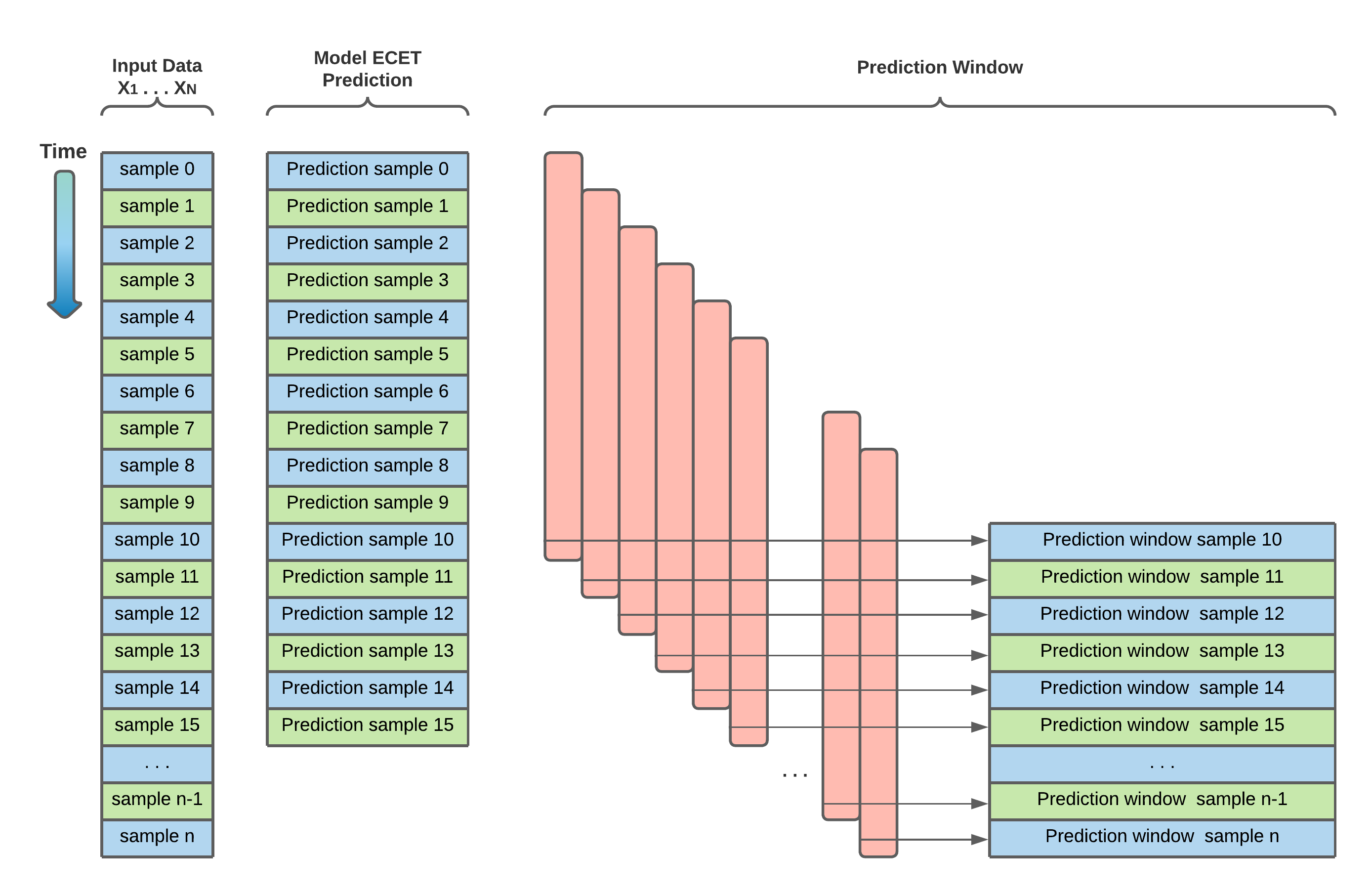}
	\caption{EC using a window.}\label{figure__F3__window_EC}
\end{figure}

A graphical representation of the window procedure is exemplified in Fig. \ref{figure__F3__window_EC}.

Having the data and the frame of discernment updated, we can proceed with the \textit{retraining of the pool of classifiers}. The retraining is performed using the training methodology presented in \cite{ArevaloIbrahim2023}. 

The last step is \textit{to test the EC} using the testing data $D^{Te}$. For this purpose, we first update the frame of discernment $\Theta_{Sys}$: 
\begin{equation} \label{model_update__11}
   \begin{split}
   	\Theta_{Sys} &= \Theta_{Sys_{Old}} \cup {A_{K}} \\
   \end{split}
\end{equation}
where $\Theta_{Sys_{Old}}$ is the old frame of discernment, $A_{K}$ is the new focal element, and $N, K \in \mathbb{N}$.

Thus, the updated $\Theta_{Sys}$ is transformed into: 
\begin{equation} \label{model_update__12}
   \begin{split}
   	\Theta_{Sys} = \{F_{1},...,F_{N}, A_{K}\}
   \end{split}
\end{equation}

\subsubsection{Model Update for KLAFATE}\label{model_update__kext}
Though knowledge-based models contain valuable expert-domain knowledge, the modeling process is time-consuming and requires frequent updates to avoid knowledge obsoleteness. 
To this end, our methodology provides the theoretical framework for uncertainty monitoring using DSET, which can be used to trigger the update of the knowledge model by the team of experts.
The \textit{model update of KLAFATE} is triggered by an uncertainty rise, either on the system or the knowledge model. Thus, the expert team is gathered to analyze the possibility of an unknown condition. Consequently, the expert team recommends adding information sources by including signals, process variables, or hardware to capture new physical signals. The latest purpose is to ease the identification of unknown conditions to create new knowledge rules in the FMEA.  
Once the expert team analyzes the acquired knowledge, the knowledge rules are validated using key performance indicators (KPI) in the short and long term.  
The process to create a rule-based system is described in \cite{ArevaloPiolo2022}.

\section{Use Case: Model Update for Ensemble Classification using Tennessee Eastman Dataset}\label{section__usecase__retrain}

As described in section \ref{section__model__update}, the approach's novelty is a methodology for updating data-based models while injecting unknown fault cases in the data. The methodology uses primarily an uncertainty monitoring approach based on DSET.
This section presents the results of the improved anomaly detection approach and the model update methodology. 
The robustness of the approaches is tested using the benchmark Tennessee Eastman. We present a description of the dataset. We describe the experiment design explaining the defined scenarios and the performance metrics. The subsection results provide the performance of the experiments. A discussion subsection closes this section by presenting the findings and limitations of the approach. The model update for the data-based model (ECET) and knowledge-based model (KLAFATE) are green highlighted in Fig. \ref{figure__F2__infusion__overview}.  



\subsection{Description of the Tennessee Eastman Dataset}
The benchmark Tennesse Eastman (TE) was created by Down and Vogel with the motivation to provide an industrial-like dataset based on the Tennesse Eastman chemical plant \cite{Downs1993}. The TE chemical plant have 
five principal process components: condenser, reactor, compressor, separator, and stripper.
The dataset is amply used in literature to compare the performance of data-based models.
The dataset models a chemical process considering 21 fault cases and a normal operation case. The dataset is divided into training sets and testing sets. The training set consists of 480 rows of data containing 52 features for each fault. In contrast, the training set of the normal condition contains 500 rows of data. The testing set consists of 960 rows of data, in which the first 160 rows belong to the normal condition and the rest 800 rows belong to the fault case. Given the prediction difficulty, the fault cases are usually grouped into three categories: easy cases (1, 2, 4, 5, 6, 7, 12, 14, 18), medium cases (8, 10, 11, 13, 16, 17, 19, 20) and hard cases (3, 9, 15 and 21) \cite{ChadhaPanambilly2020}. A detailed dataset description can be found in \cite{Downs1993} \cite{ArevaloIbrahim2023}.


\subsection{Experiment Design}
We followed the procedure proposed in \cite{ArevaloIbrahim2023}, in which we used the benchmark TE to test the performance of the proposed approaches. Besides, we considered a pool of ten classifiers (e.g., five NN-based models and five non-NN-based models) as the basis of the ECs. We considered only experiments using ML-based ECs, and Hybrid ECs (a combination of non-NN-based classifiers and NN-based classifiers). 
The procedure is documented in detail in \cite{ArevaloIbrahim2023}.
We trained the classifiers of the ECs using the fault cases (0,1,2,6,12) as the basis of the experiments. 
We defined two experiment scenarios: data isolation using a window and an update of ECs.
We develop the approach using the IDE Anaconda and the libraries Scikit-learn and PyTorch \cite{PedregosaVaroquaux2011} \cite{PaszkeGross2019} \cite{Anaconda2016}. We perform the experiments on a Ubuntu 20.04.3 LTS environment using a CPU i7-7700 @3.60GHz x 8, 32GB RAM, and a GPU NVIDIA GeForce GTX 1660 SUPER.

\subsubsection{Data isolation using a window}
We selected the MC ECs M3 and H5-2 from the previous work \cite{ArevaloIbrahim2023} with the best performance criteria. The EC M3 consists of non-NN classifiers, whereas the EC H5-2 is hybrid.
We compared the results obtained by performing a variation on the window size.
The base classifiers' and ECs' hyperparameters are detailed in \cite{ArevaloIbrahim2023}. 

\subsubsection{Update of ECs}
We selected the ML-based ECs M3, M4, and M5 to perform the experiments and comparisons. Given the constraint of limited retraining data, we discard NN-based and Hybrid ECs.
The procedure consists of two data batches for each experiment. The first batch contains the known fault cases (0,1,2,6,12) and one anomaly case (e.g., fault case 7). The EC identifies the anomaly through uncertainty monitoring, collects the anomalous data, and retrains the EC if the data is sufficient. We assign the anomaly data with the arbitrary label 30.
The second batch contains testing data of the fault cases (0,1,2,6,12) and the anomaly (e.g., fault case 7). For comparison purposes, the original label 7 is changed by the new label 30.
We defined three main experiments, namely, the retraining of the ECs using all the fault cases (1,...,21), the study of the retraining parameters (e.g., threshold size, window size, and detection patience) using the fault cases (7,8,15), and the fine-tuned retrained ECs using all the faults (1,...,21). We selected the fault cases (7,8,15) as anomalies to have a case for each primary data group (easy, medium, and hard). 

\subsubsection{Performance Metrics}
We use the performance metrics F1-score (F1) and fault detection rate (FDR, also known as recall). F1 and FDR are detailed in \cite{Panda2021}.

\subsection{Results}
This subsection presents the experiment results of the model update approach. For this purpose, the experiments are divided into two parts: data isolation using a window and a model update of EC.

\subsubsection{Data Isolation using a Window}
We perform experiments using different window sizes to study their impact on the EC performance. We compare the effects of using no-window ($w=0$) and a window ($w=20$, $w=50$).

Table \ref{table__results__anomaly__classification__MC__2} presents the F1-scores of the BIN EC M5 and MC EC H5-2. The hyperparameters of the base classifiers and ECs were reported in detail in \cite{ArevaloIbrahim2023}. 
The BIN EC M5 presents comparable results while varying the window size with average F1-scores of 0.6\%, 0.64\%, and 0.65\% for the window sizes (0, 20, 50), respectively. In contrast, the MC EC H5-2 presented higher results using a window (20,50) compared to no-window $w=0$. The MC EC H5-2 presented average F1-scores of 0.63\%, 0.81\%, and 0.88\% for the window sizes (0,20,50), respectively.

\begin{table}[!ht]
\centering
\caption{Anomaly detection results of selected ensemble multiclass classifiers using all the fault cases, and F1-score.}
\begin{tabular}{c|ccc|ccc}
\hline
\multirow{2}{*}{\textbf{Fault}} & \multicolumn{3}{c|}{\textbf{BIN   EC M5}} & \multicolumn{3}{c}{\textbf{MC EC H5-2}} \\ \cline{2-7} 
 & \textbf{w=0} & \textbf{w=20} & \textbf{w=50} & \textbf{w=0} & \textbf{w=20} & \textbf{w=50} \\ \hline
1 & 0.61 & 0.70 & 0.74 & 0.61 & 0.79 & 0.88 \\
2 & 0.48 & 0.53 & 0.62 & 0.55 & 0.72 & 0.72 \\
3 & 0.50 & 0.41 & 0.36 & 0.63 & 0.88 & 0.93 \\
4 & 0.50 & 0.44 & 0.35 & 0.60 & 0.87 & 0.95 \\
5 & 0.65 & 0.77 & 0.80 & 0.60 & 0.81 & 0.89 \\
6 & 0.91 & 0.94 & 0.92 & 0.42 & 0.35 & 0.32 \\
7 & 0.67 & 0.74 & 0.74 & 0.59 & 0.67 & 0.75 \\
8 & 0.44 & 0.38 & 0.25 & 0.72 & 0.82 & 0.93 \\
9 & 0.58 & 0.57 & 0.56 & 0.64 & 0.88 & 0.95 \\
10 & 0.55 & 0.61 & 0.58 & 0.64 & 0.87 & 0.91 \\
11 & 0.65 & 0.70 & 0.76 & 0.61 & 0.84 & 0.94 \\
12 & 0.57 & 0.58 & 0.65 & 0.62 & 0.77 & 0.86 \\
13 & 0.52 & 0.55 & 0.61 & 0.70 & 0.81 & 0.88 \\
14 & 0.63 & 0.74 & 0.76 & 0.62 & 0.87 & 0.94 \\
15 & 0.59 & 0.66 & 0.71 & 0.64 & 0.89 & 0.94 \\
16 & 0.57 & 0.62 & 0.60 & 0.63 & 0.83 & 0.93 \\
17 & 0.61 & 0.69 & 0.69 & 0.63 & 0.87 & 0.95 \\
18 & 0.80 & 0.80 & 0.78 & 0.85 & 0.93 & 0.96 \\
19 & 0.62 & 0.68 & 0.68 & 0.62 & 0.88 & 0.93 \\
20 & 0.61 & 0.64 & 0.61 & 0.60 & 0.84 & 0.89 \\
21 & 0.63 & 0.72 & 0.81 & 0.62 & 0.88 & 0.92 \\ \hline
\multicolumn{1}{r|}{Avg F1-score} & 0.60 & 0.64 & 0.65 & 0.63 & 0.81 & 0.88 \\ \hline
\end{tabular}
\label{table__results__anomaly__classification__MC__2}
\end{table}

\begin{figure*}[!ht]
	\centering
	\begin{subfigure}[b]{0.3\textwidth}
	\includegraphics[width=\textwidth,keepaspectratio]{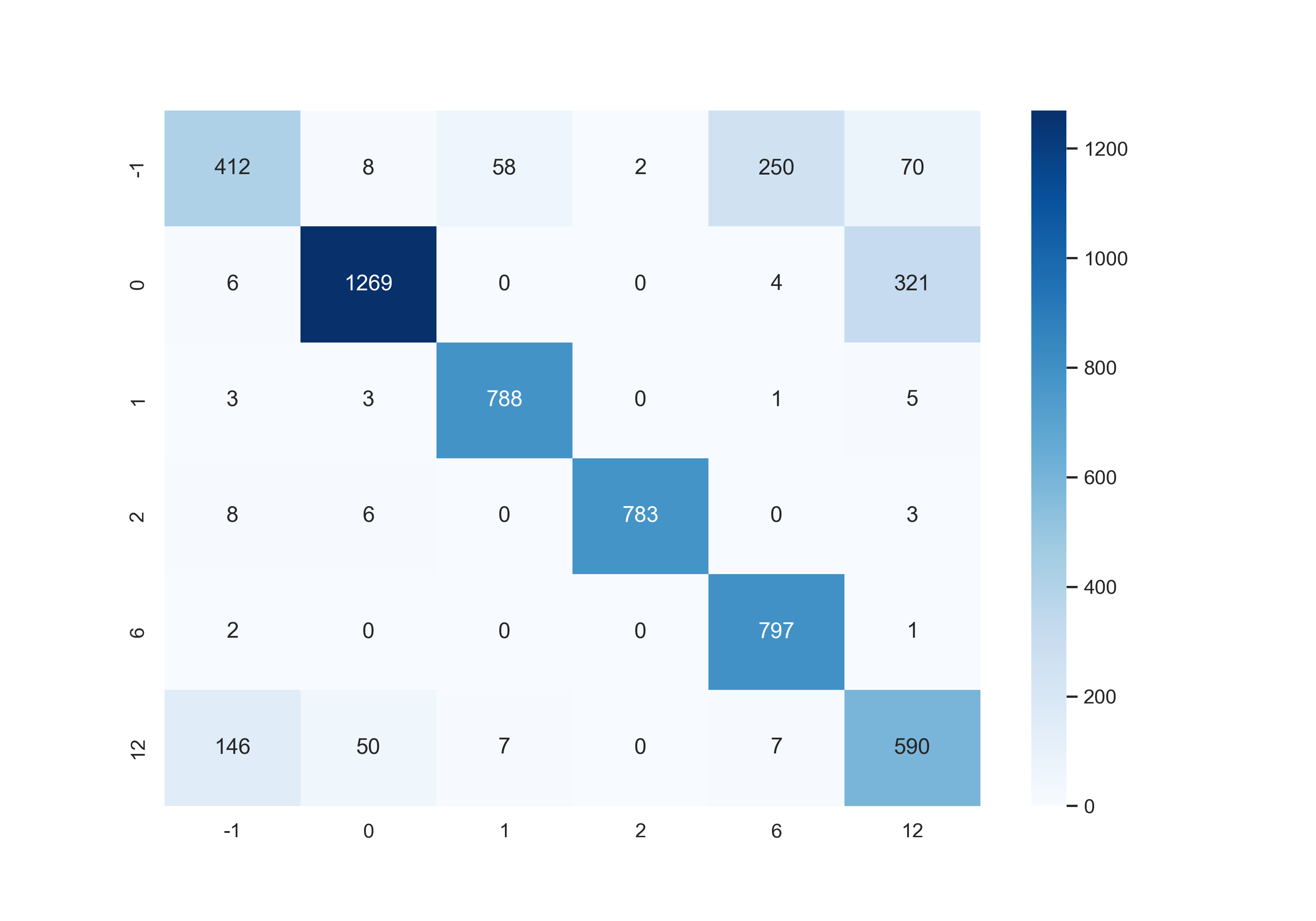}
	\caption{Confusion matrix of EC H5-2 w=0}\label{fig__window__cm__w0}
	\end{subfigure}
    \begin{subfigure}[b]{0.3\textwidth}
	\includegraphics[width=\textwidth,keepaspectratio]{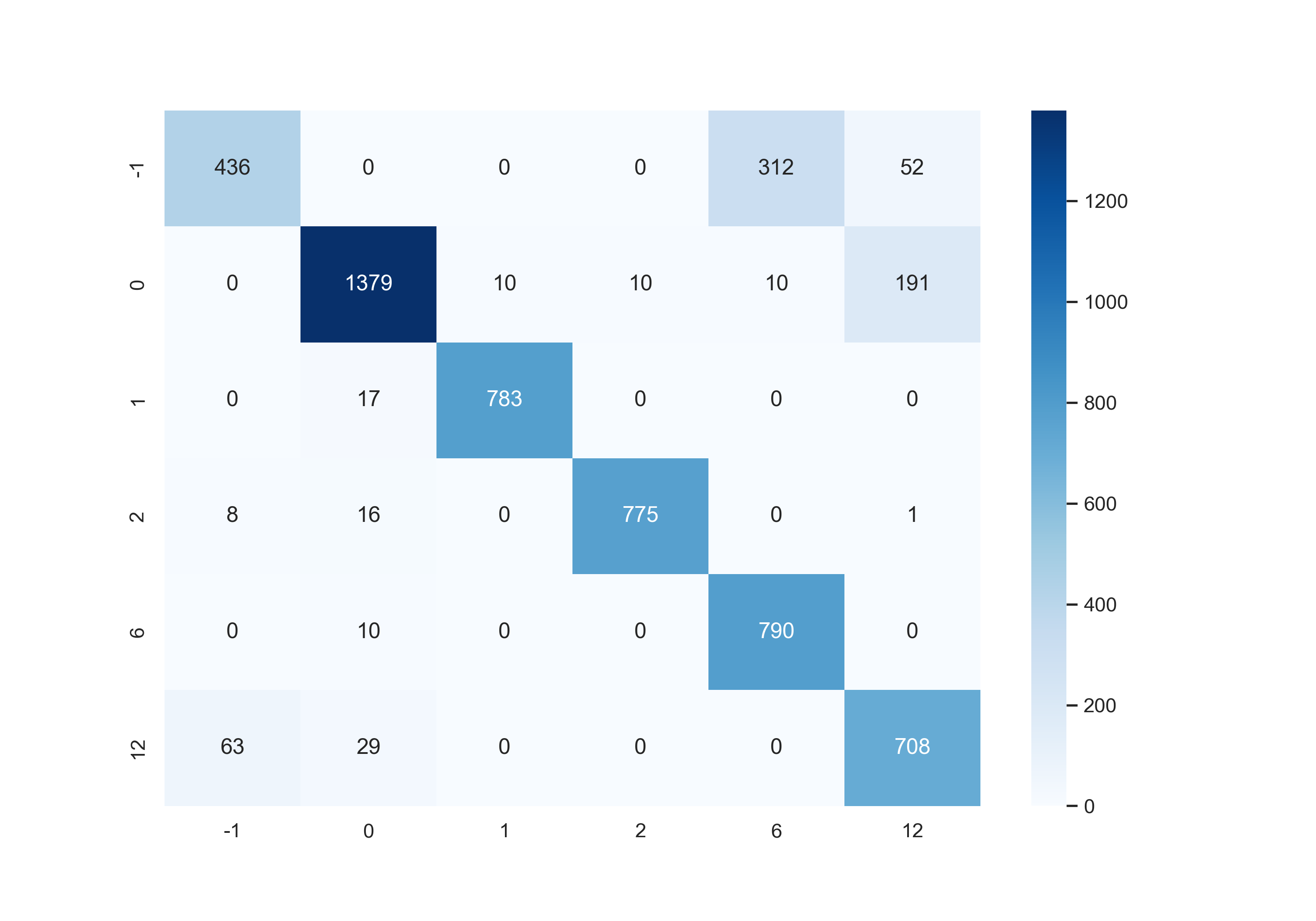}
	\caption{Confusion matrix of EC H5-2 w=20}\label{fig__window__cm__w20}
	\end{subfigure}
    \begin{subfigure}[b]{0.3\textwidth}
	\includegraphics[width=\textwidth,keepaspectratio]{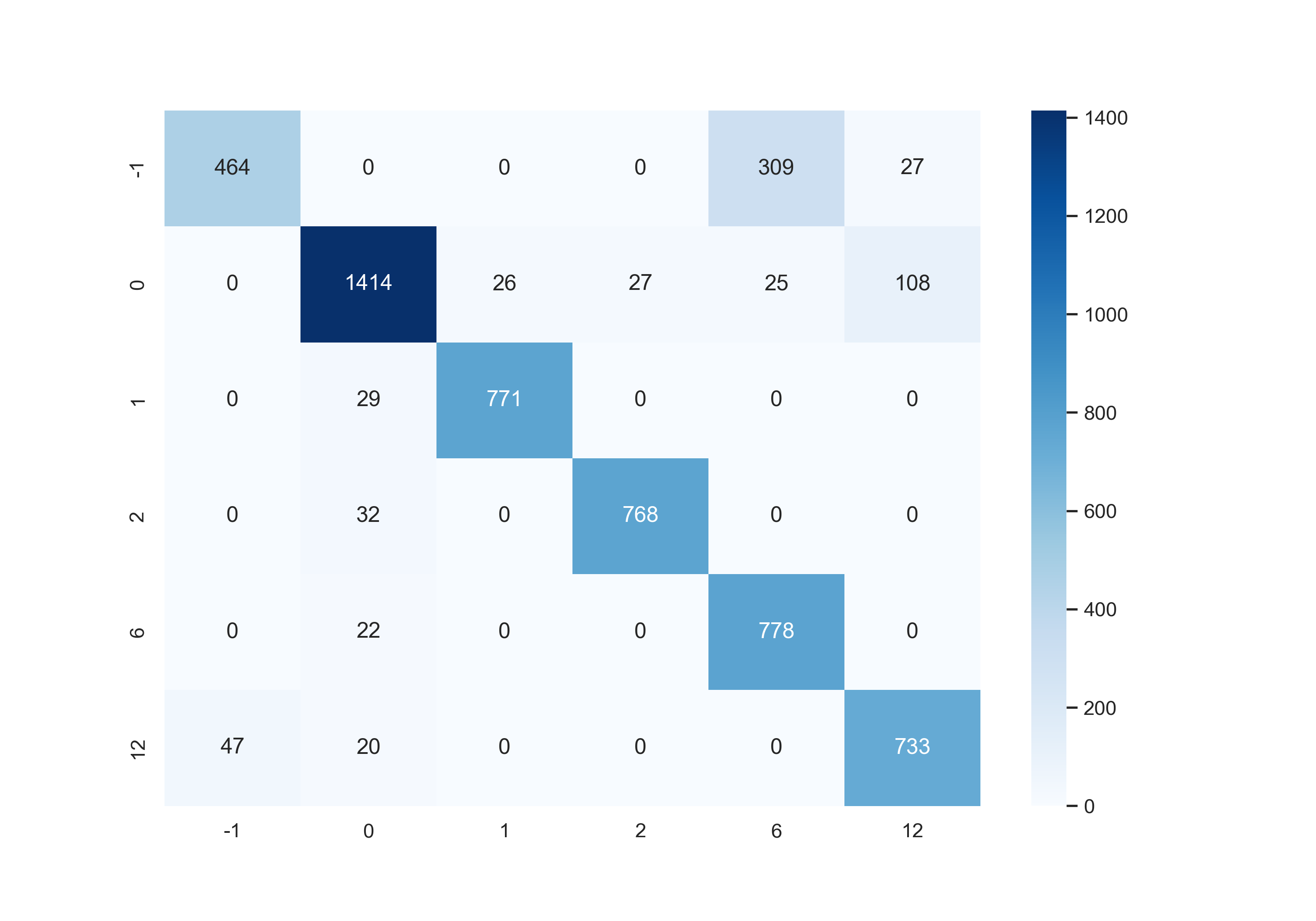}
	\caption{Confusion matrix of EC H5-2 w=50}\label{fig__window__cm__w50}
	\end{subfigure}    
    ~
    	\begin{subfigure}[b]{0.3\textwidth}
	\includegraphics[width=\textwidth,keepaspectratio]{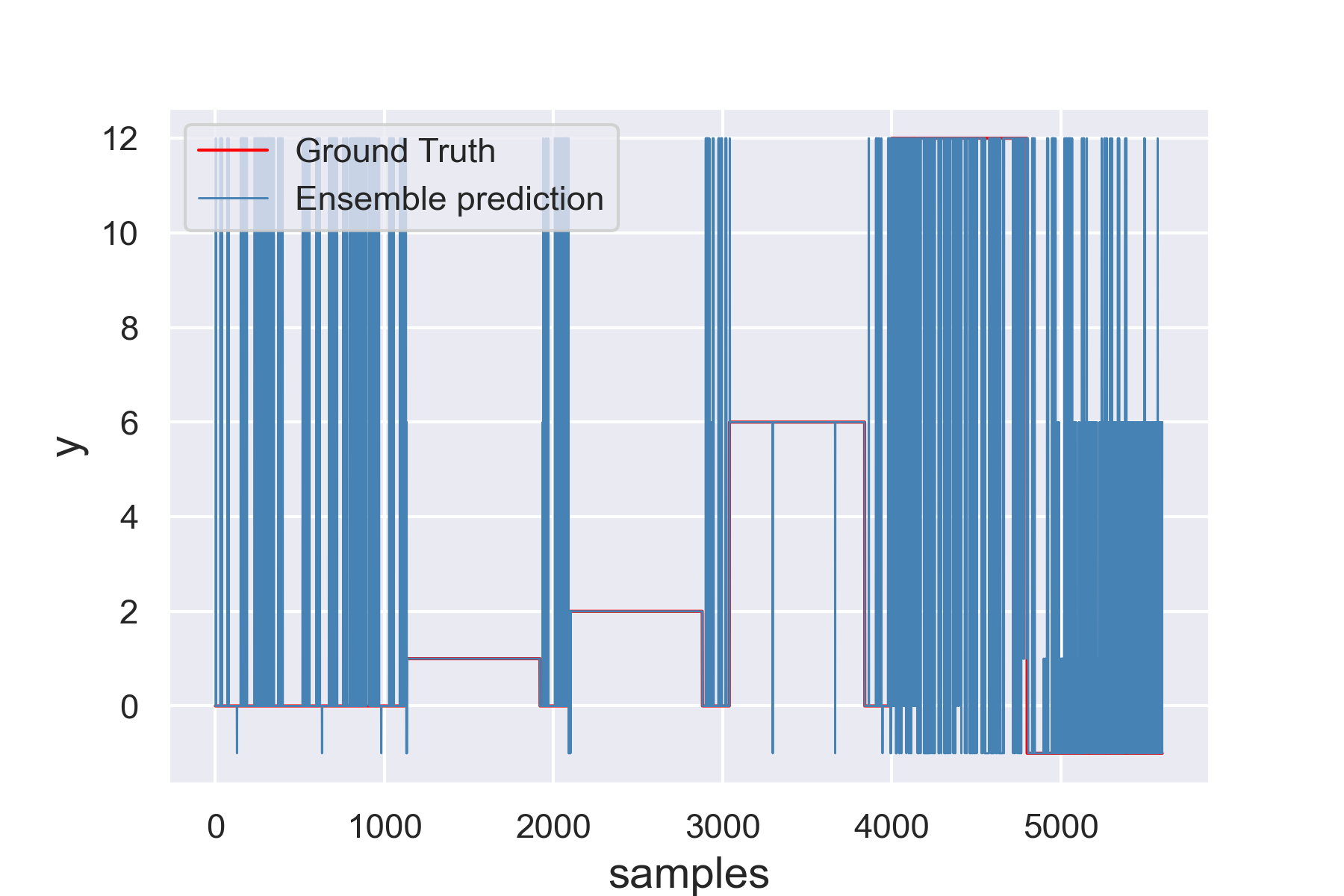}
	\caption{Predictions of EC H5-2 w=0}\label{fig__window__prediction__w0}
	\end{subfigure}
    \begin{subfigure}[b]{0.3\textwidth}
	\includegraphics[width=\textwidth,keepaspectratio]{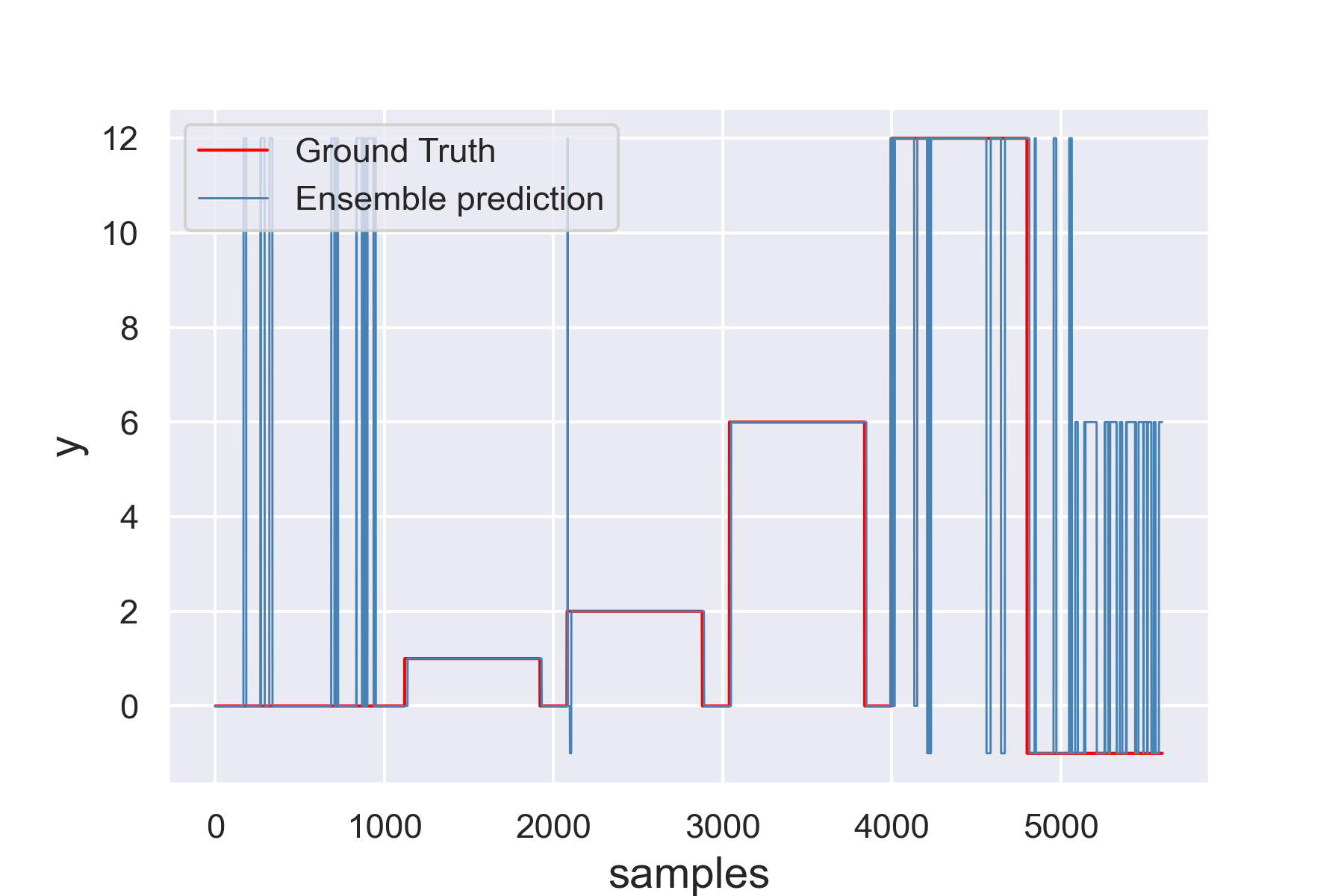}
	\caption{Predictions of EC H5-2 w=20}\label{fig__window__prediction__w20}
	\end{subfigure}
    \begin{subfigure}[b]{0.3\textwidth}
	\includegraphics[width=\textwidth,keepaspectratio]{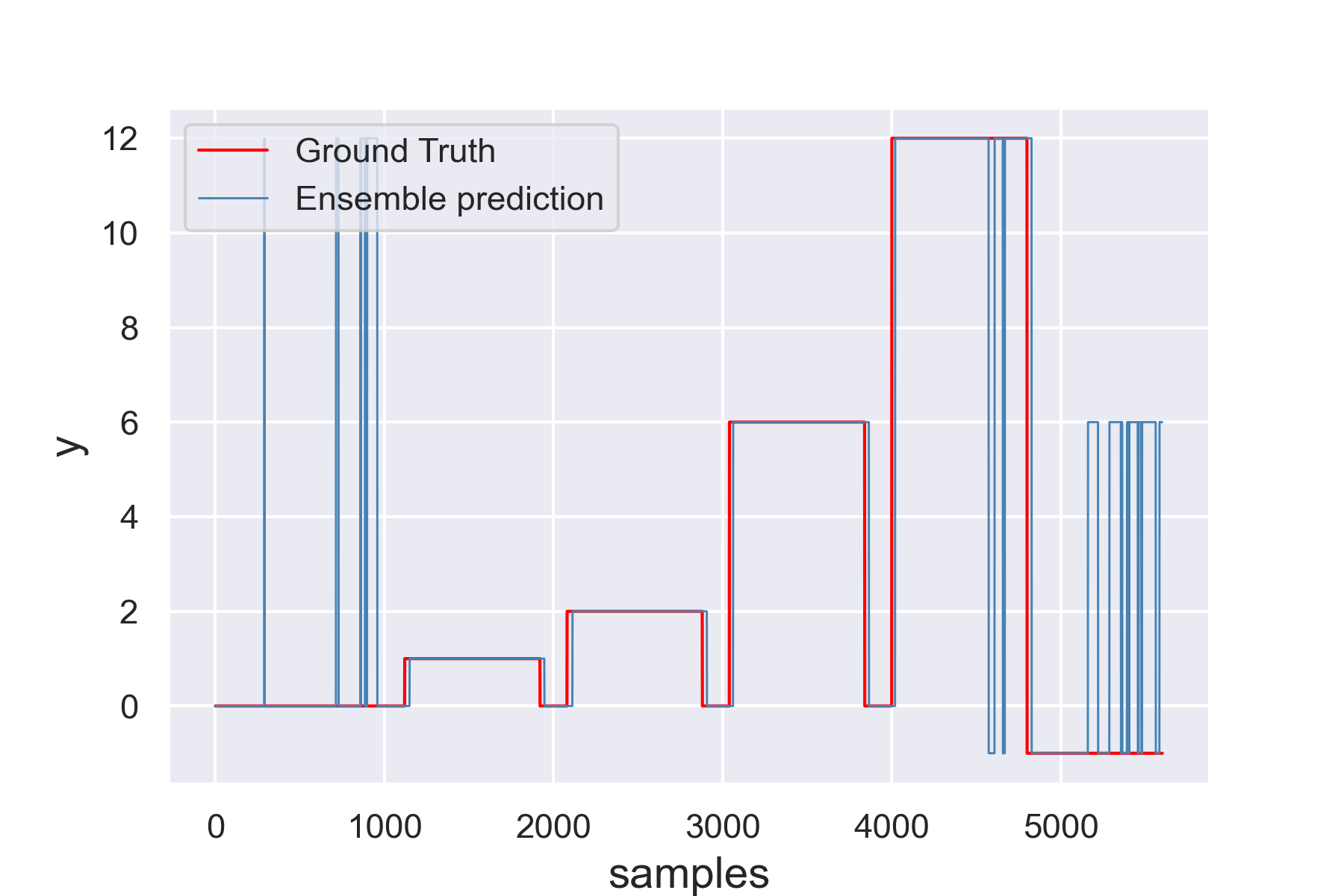}
	\caption{Predictions of EC H5-2 w=50}\label{fig__window__prediction__w50}
	\end{subfigure}
	\caption{Anomaly detection using different window sizes for 
    the MC EC H5-2 trained with the known cases 0,1,2,6,12, and using the fault case (7) as an anomaly. The confusion matrices of H5-2 are displayed in (a)-(c), and the predictions in (d)-(f). 
    }\label{figure__EC__window_size}
\end{figure*}

Fig. \ref{figure__EC__window_size} presents the plots of the MC EC H5-2 trained with fault cases (0,1,2,6,12) and using the anomaly fault case (7) while doing a variation on the window size (0, 20, 50). Figures \ref{fig__window__cm__w0}, \ref{fig__window__cm__w20} and \ref{fig__window__cm__w50} show the confusion matrices for the window sizes $w=0$, $w=20$, and $w=50$, respectively. The confusion matrices for the window sizes $w=20$ and $w=50$ present better results than the confusion matrix with window size $w=0$. The predictions plots of figures \ref{fig__window__prediction__w0}, and \ref{fig__window__prediction__w0} confirm the results of the confusion matrices, in which the predictions (blue) are closer to the ground truth (red) for EC using the window sizes $w=20$ and $w=50$. The anomaly case (7) is represented as the label (-1) in the predictions plot. It is important to remark that the approach using a window smooths the EC predictions.


\subsubsection{Model update of EC}

We perform three different experiments in this subsubsection: the model update of the EC (retraining), the study of the variation of the retraining parameters, and finally, selecting a fine-tuned retrained EC.

We test the \textit{model update of the EC} using all the fault cases of the TE dataset. For this purpose, we selected the MC ECs M3, M4, and M5. The hyperparameters of the base classifiers and ECs were reported in detail in \cite{ArevaloIbrahim2023}.
Table \ref{table__results__RT__raw} presents the F1-scores of the MC ECs M3, M4, and M5 trained with the fault cases (0,1,2,6,12). The MC ECs M3, M4, and M5 present comparable results with an average F1-score of 0.39, 0.36, and 0.37, respectively. The EC MC M3 detected the anomalies (7,17) with F1-scores higher or equal to 0.43 and the anomalies (13,14) with F1-scores higher or equal to 0.33 and less than 0.43. 
The EC MC M4 detected the anomalies (8,14,17) with F1-scores higher or equal to 0.67 and the anomalies (7,10,11,15) with F1-scores higher or equal to 0.38 and less than 0.54. Alternatively, the EC M5 detected the anomalies (14,18,20) with F1-scores higher or equal to 0.54 and the anomalies (8,17) with F1-scores higher or equal to 0.43 and less than 0.54.  

\begin{table}[!ht]
\centering
\caption{Classification results of the ECs after retraining using all the fault cases, and F1-score. The retraining parameters are threshold size $th=100$, window size $ws=20$, and detection patience $pt=15$. }
\begin{tabular}{c|ccc}
\hline
\multirow{2}{*}{\textbf{Fault}} & \multicolumn{3}{c}{\textbf{RT MC EC (0,1,2,6,12)}} \\ \cline{2-4} 
 & \textbf{M3} & \textbf{M4} & \textbf{M5} \\ \hline
1 & 0.98 & 0.99 & 0.99 \\
2 & 0.99 & 0.99 & 0.98 \\
3 & 0.03 & 0.01 & 0.00 \\
4 & 0.30 & 0.15 & 0.26 \\
5 & 0.22 & 0.07 & 0.15 \\
6 & 1.00 & 1.00 & 1.00 \\
7 & 0.71 & 0.41 & 0.28 \\
8 & 0.23 & 0.46 & 0.43 \\
9 & 0.07 & 0.03 & 0.10 \\
10 & 0.25 & 0.13 & 0.12 \\
11 & 0.29 & 0.15 & 0.09 \\
12 & 0.95 & 0.95 & 0.95 \\
13 & 0.36 & 0.21 & 0.08 \\
14 & 0.33 & 0.66 & 0.60 \\
15 & 0.22 & 0.07 & 0.00 \\
16 & 0.03 & 0.27 & 0.02 \\
17 & 0.43 & 0.41 & 0.47 \\
18 & 0.06 & 0.02 & 0.78 \\
19 & 0.30 & 0.08 & 0.00 \\
20 & 0.28 & 0.55 & 0.54 \\
21 & 0.05 & 0.05 & 0.00 \\ \hline
\textbf{Avg   F1-score} & \textbf{0.39} & \textbf{0.36} & \textbf{0.37} \\ \hline
\end{tabular}
\label{table__results__RT__raw}
\end{table}

\begin{figure*}[!ht]
	\centering
	\begin{subfigure}[b]{0.3\textwidth}
	\includegraphics[width=\textwidth,keepaspectratio]{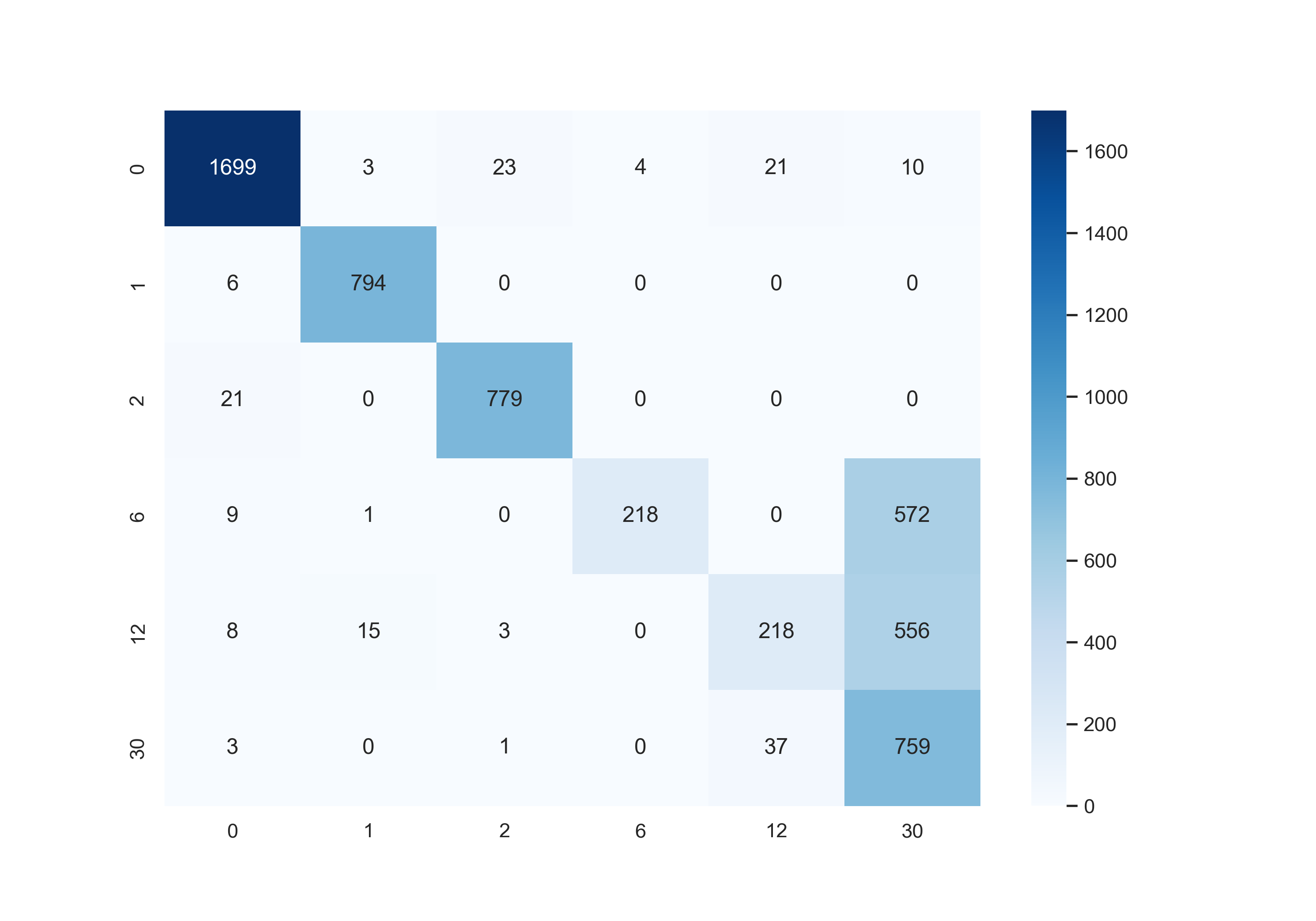} 
	\caption{CM for MC EC M3}\label{figure__use_cm__AD__M3__RT}
	\end{subfigure}
	\begin{subfigure}[b]{0.3\textwidth}
	\includegraphics[width=\textwidth,keepaspectratio]{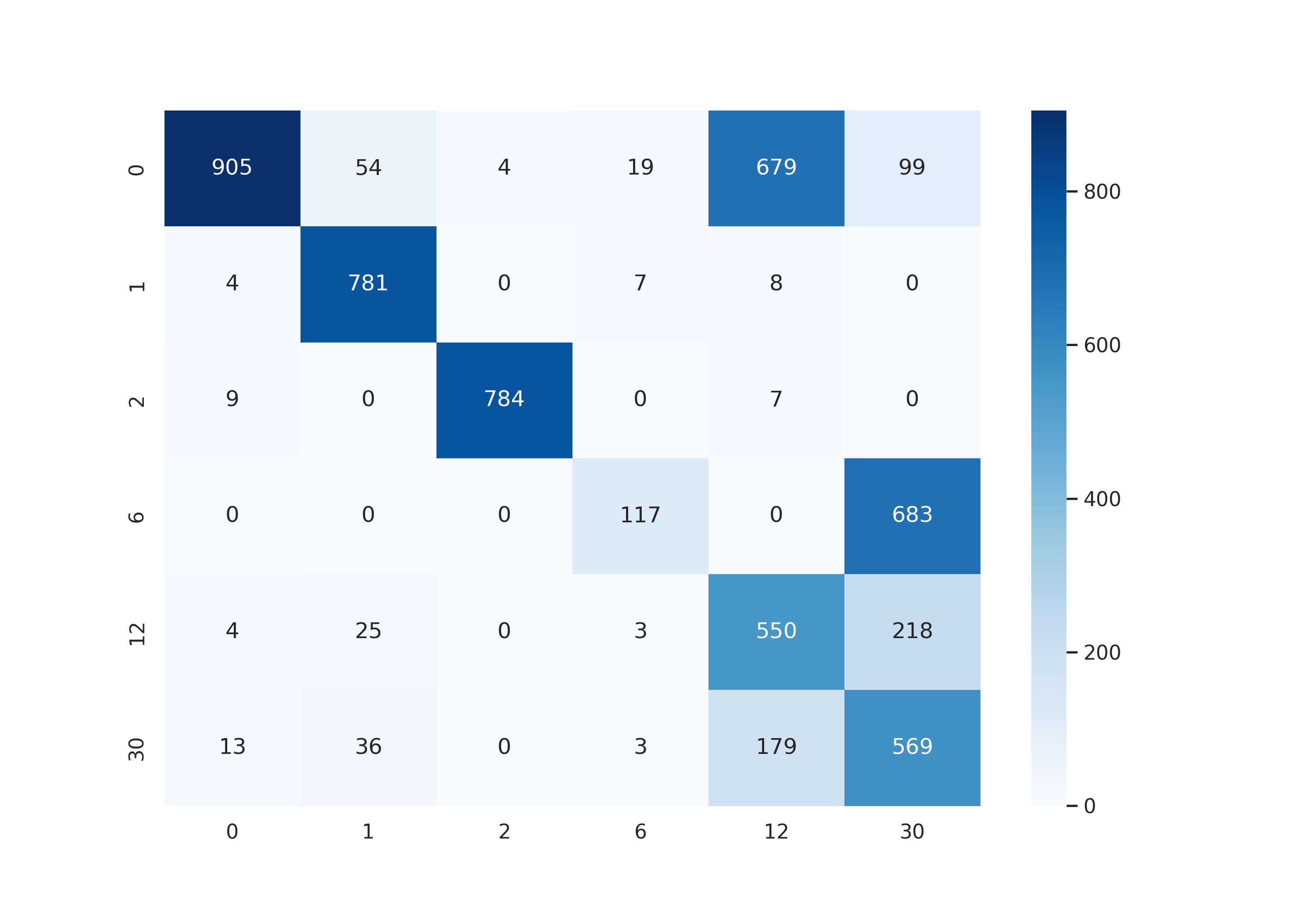} 
	\caption{CM for MC EC M4}\label{figure__use_cm__AD__M4__RT}
	\end{subfigure}
	\begin{subfigure}[b]{0.3\textwidth}
	\includegraphics[width=\textwidth,keepaspectratio]{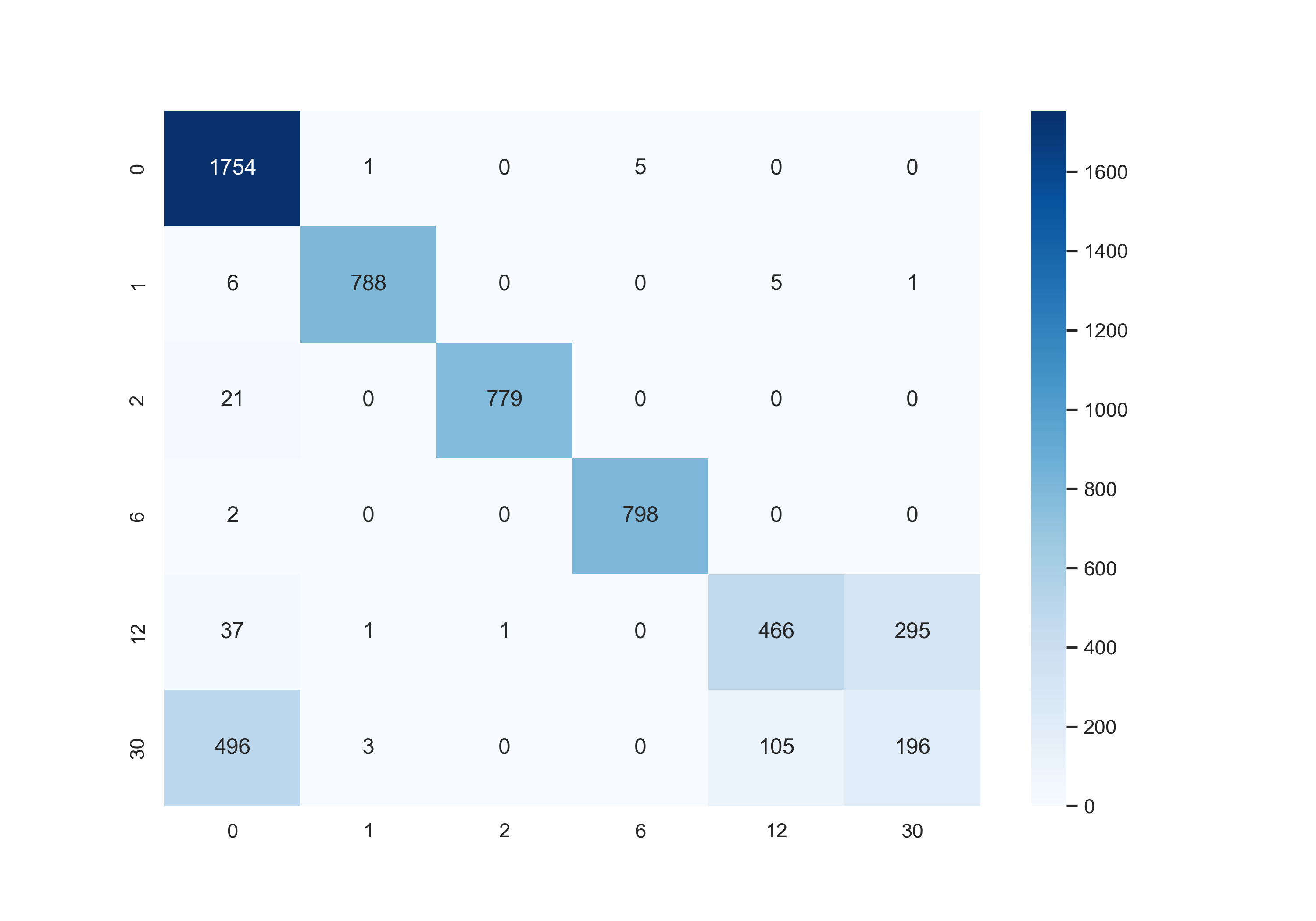} 
	\caption{CM for MC EC M5}\label{figure__use_cm__AD__M5__RT}
	\end{subfigure}
	~
	\begin{subfigure}[b]{0.3\textwidth}
	\includegraphics[width=\textwidth,keepaspectratio]{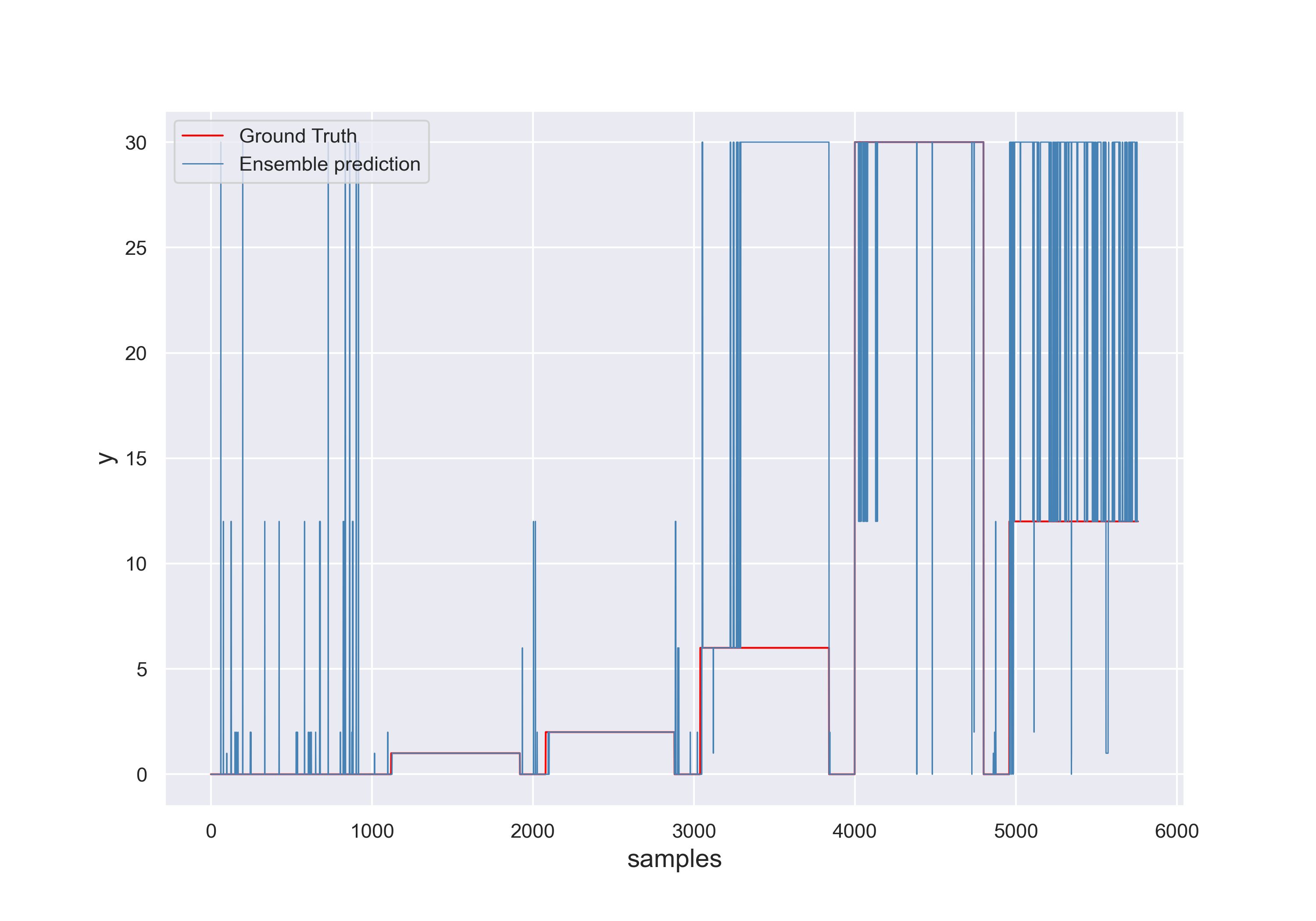} 
	\caption{Predictions for MC EC M3}\label{figure__use_classification__AD__M3__RT}
	\end{subfigure}
	\begin{subfigure}[b]{0.3\textwidth}
	\includegraphics[width=\textwidth,keepaspectratio]{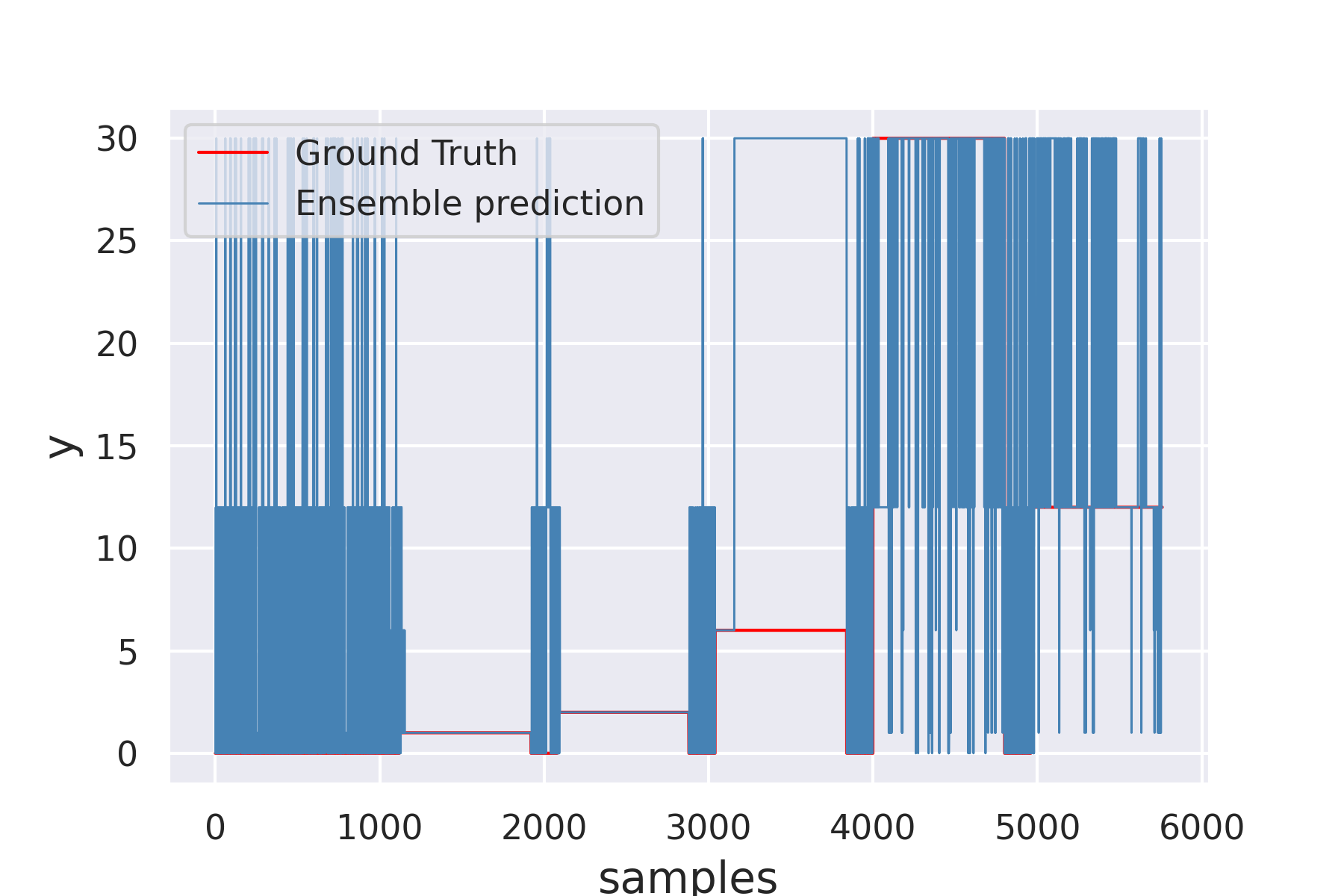}  
	\caption{Predictions for MC EC M4}\label{figure__use_classification__AD__M4__RT}
	\end{subfigure}
	\begin{subfigure}[b]{0.3\textwidth}
	\includegraphics[width=\textwidth,keepaspectratio]{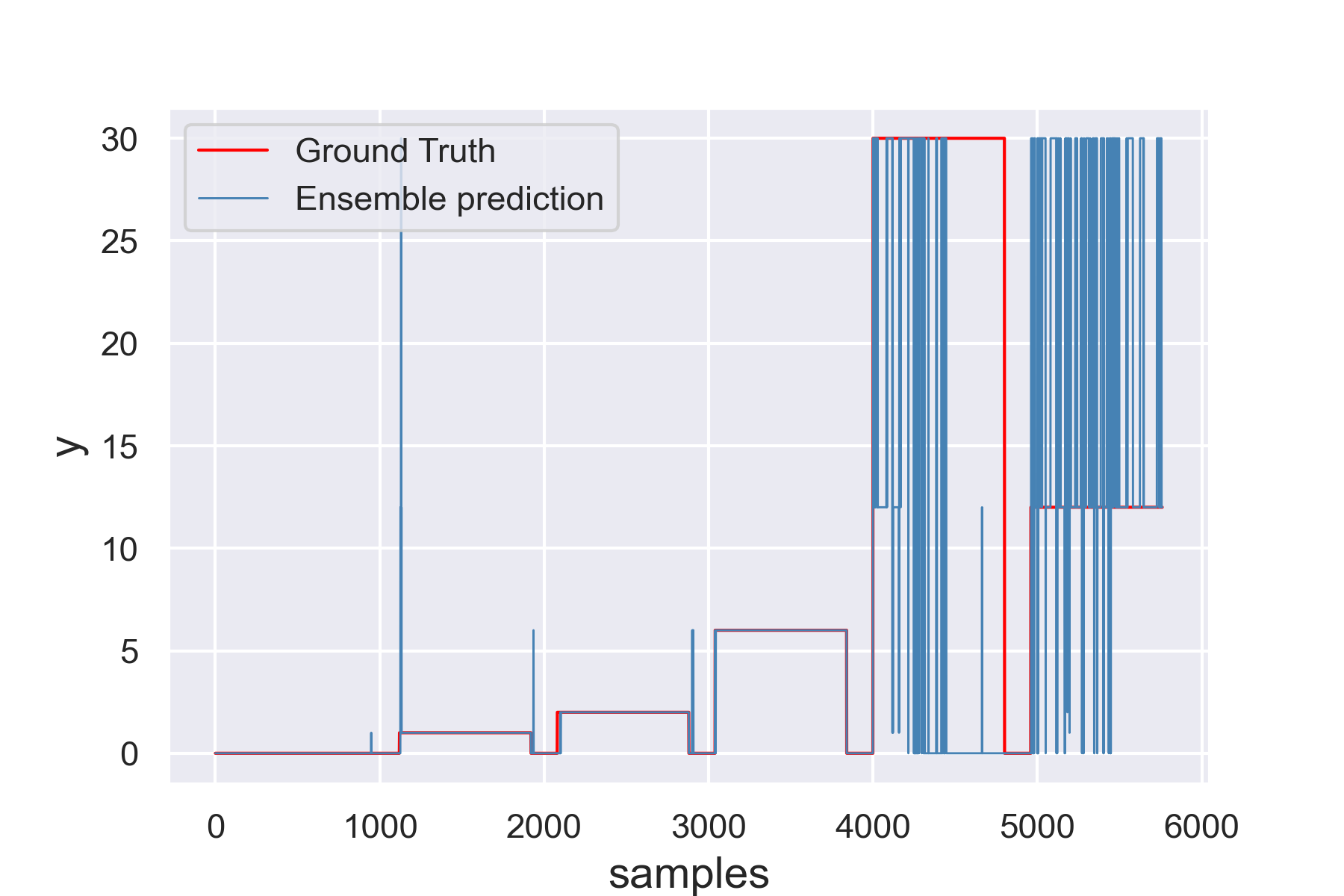} 
	\caption{Predictions for MC EC M5}\label{figure__use_classification__AD__M5__RT}
	\end{subfigure}
	~
	\begin{subfigure}[b]{0.3\textwidth}
	\includegraphics[width=\textwidth,keepaspectratio]{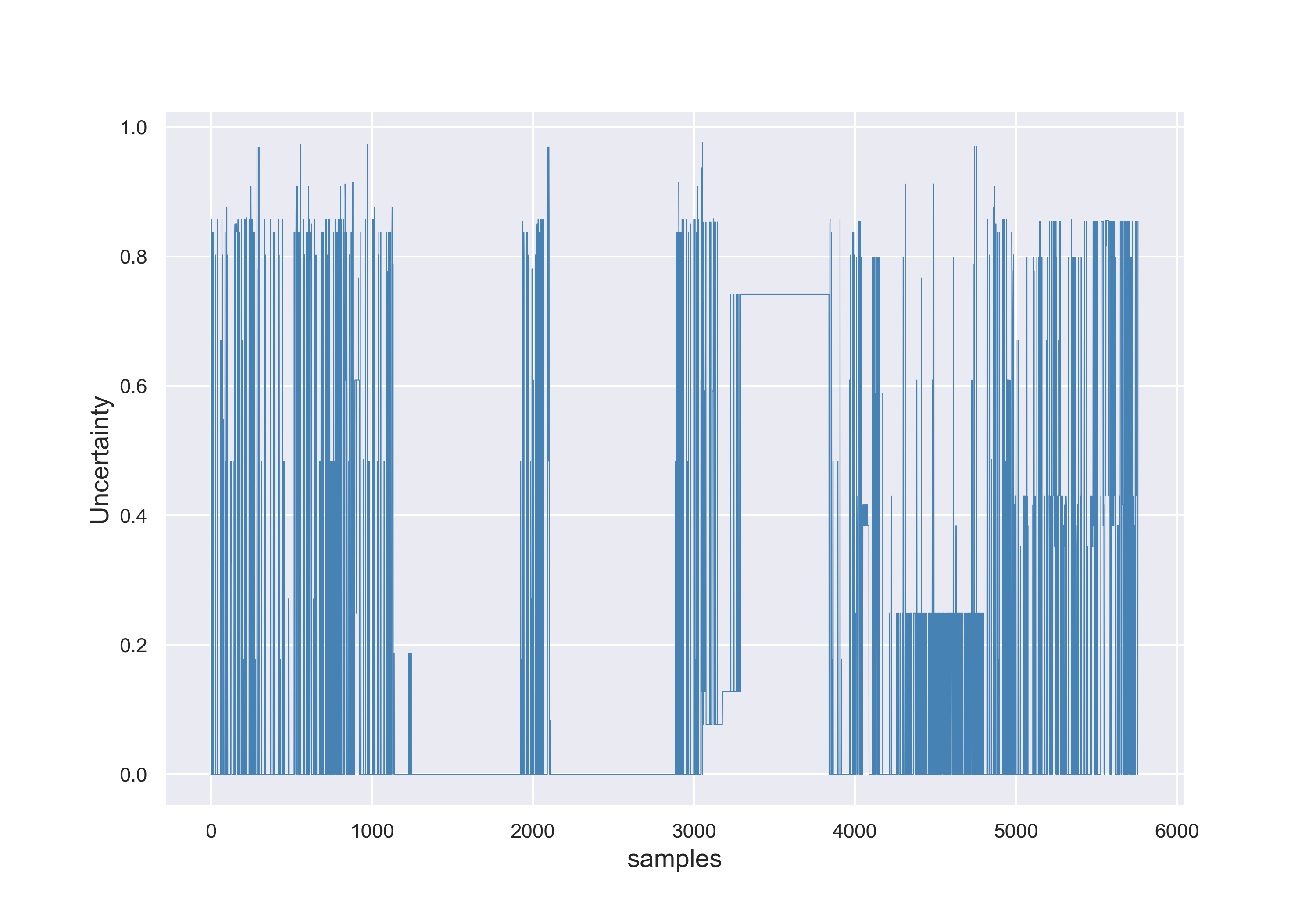}
	\caption{DSET UQ for MC EC M3}\label{figure__use_UQDS__AD__M3__RT}
	\end{subfigure}
		\begin{subfigure}[b]{0.3\textwidth}
	\includegraphics[width=\textwidth,keepaspectratio]{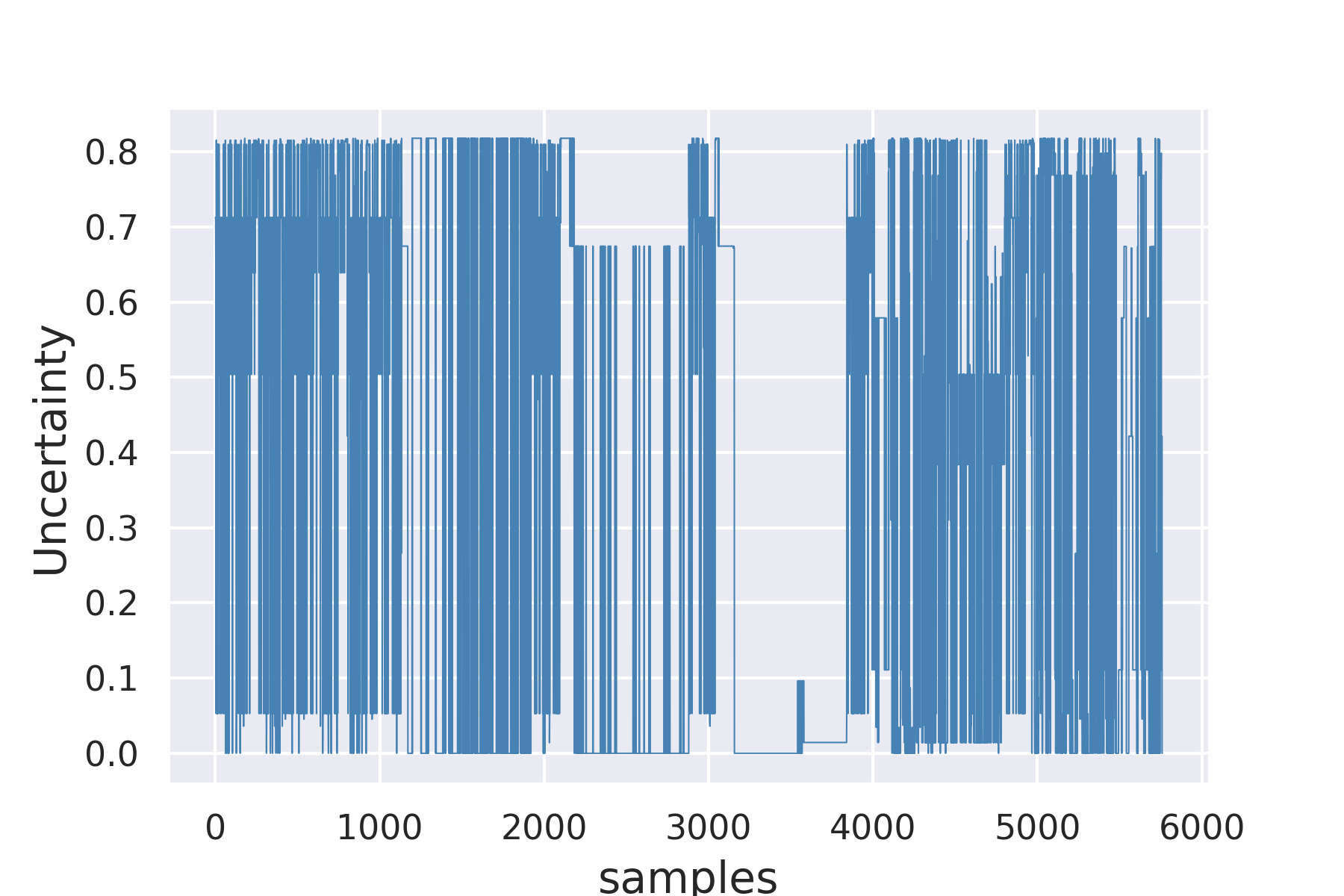}
	\caption{DSET UQ for MC EC M4}\label{figure__use_UQDS__AD__M4__RT}
	\end{subfigure}
	\begin{subfigure}[b]{0.3\textwidth}
	\includegraphics[width=\textwidth,keepaspectratio]{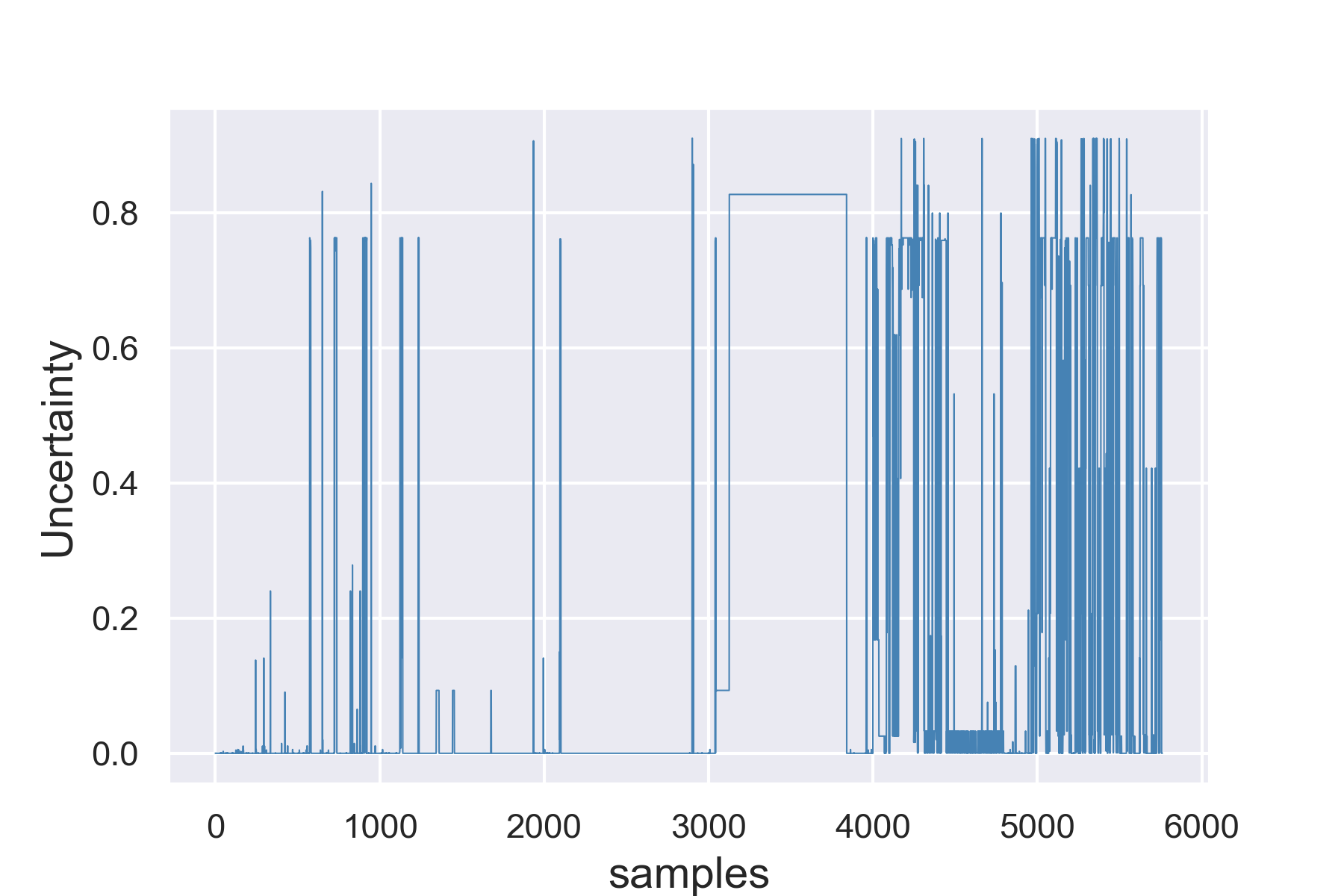}
	\caption{DSET UQ for MC EC M5}\label{figure__use_UQDS__AD__M5__RT}
	\end{subfigure}
	\caption{Anomaly Detection and UQ results for MC ECs M3, M4 and M5 trained with the fault cases (0,1,2,6,12): Confusion matrices (a)-(c), classification results (d)-(f), and DSET UQ (g)-(i) while injecting anomaly 7.
  }\label{figure__results__anomaly}
\end{figure*}

Fig. \ref{figure__results__anomaly} presents the plots of the MC ECs M3, M4, and M5 trained with fault cases (0,1,2,6,12) and using the anomaly fault 7. Figures \ref{figure__use_cm__AD__M3__RT}, \ref{figure__use_cm__AD__M4__RT} and \ref{figure__use_cm__AD__M5__RT} show the confusion matrices for the ECs M3, M4, and M5, respectively. The confusion matrix of the MC EC M5 presents better results than the confusion matrices of the other ECs. Alternatively, the prediction plots of figures \ref{figure__use_classification__AD__M3__RT}, \ref{figure__use_classification__AD__M4__RT} and \ref{figure__use_classification__AD__M5__RT} present mixed results, in which M3 identifies the anomaly better, but the case (12) is confused with the anomaly. In addition, M5 presents a better prediction of the known fault cases but has a lower anomaly detection. The uncertainty quantification (UQ) using DSET is presented in figures \ref{figure__use_UQDS__AD__M3__RT}, \ref{figure__use_UQDS__AD__M4__RT} and \ref{figure__use_UQDS__AD__M5__RT} for the MC ECs M3, M4, and M5, respectively. The MC EC M5 presents steadier values than the MC ECs M3 and M4, which confirms the prediction pattern. The latest can be enunciated as the lower the uncertainty, the better the classification performance (likeliness). 

The next step is the \textit{study of the retraining parameters}. For this purpose, we test the effects of the threshold size, window size, and detection patience. We chose the MC EC M3 to perform the experiments and selected the threshold sizes (150,250,350) and anomalies (7,8,15).

\paragraph{Effects of the threshold size}
Table \ref{table__results__RT__threshold} presents the F1-scores of the MC ECs M3, M4, and M5 trained with the fault cases (0,1,2,6,12). The retraining parameters window size and detection patience are fixed with values of $ws=20$ and $pt=15$, respectively. 
The MC EC M3 presented higher results using a threshold size $th=150$ with an average F1-score of 0.81 for the anomaly (7), compared with the values of 0.57 and 0.50, corresponding to the threshold sizes (250, 250).
The MC EC M3 presents comparable results for the anomaly (8) with average F1-scores of 0.81, 0.82, and 0.82 for the threshold sizes (150, 250, 350), respectively. In contrast, the MC EC M3 presented higher results using a threshold size $th=350$ with an average F1-score of 0.74 for the anomaly 15, in comparison with the values of 0.54 and 0.55, which correspond to the threshold sizes (150, 250), respectively. 

\begin{table*}[!ht]
\centering
\caption{Anomaly detection results of MC EC M3 using the fault cases (0,1,2,6,12), the anomalies (7,8,15), thresholds variations (150,250,350), window size (20), patience (15), and F1-score.}
\begin{tabular}{c|ccccccccc}
\hline
\multirow{2}{*}{\textbf{Fault}} & \multicolumn{9}{c}{\textbf{MC EC M3}} \\ \cline{2-10} 
 & \multicolumn{3}{c|}{\textbf{A7}} & \multicolumn{3}{c|}{\textbf{A8}} & \multicolumn{3}{c}{\textbf{A15}} \\ \hline
 & \textbf{Th=150} & \textbf{Th=250} & \multicolumn{1}{c|}{\textbf{Th=350}} & \textbf{Th=150} & \textbf{Th=250} & \multicolumn{1}{c|}{\textbf{Th=350}} & \textbf{Th=150} & \textbf{Th=250} & \textbf{Th=350} \\ \cline{2-10} 
0 & 0.92 & 0.38 & \multicolumn{1}{c|}{0.11} & 0.9 & 0.91 & \multicolumn{1}{c|}{0.91} & 0.35 & 0.51 & 0.76 \\
1 & 0.99 & 0.95 & \multicolumn{1}{c|}{0.95} & 1 & 0.95 & \multicolumn{1}{c|}{0.94} & 0.91 & 0.91 & 0.97 \\
2 & 0.96 & 0.98 & \multicolumn{1}{c|}{0.96} & 0.9 & 0.91 & \multicolumn{1}{c|}{0.91} & 0.98 & 0.98 & 0.97 \\
6 & 0.99 & 0.22 & \multicolumn{1}{c|}{0.21} & 1 & 0.99 & \multicolumn{1}{c|}{0.99} & 0.22 & 0.28 & 1.00 \\
12 & 0.47 & 0.44 & \multicolumn{1}{c|}{0.35} & 0.7 & 0.73 & \multicolumn{1}{c|}{0.74} & 0.5 & 0.42 & 0.71 \\
30 & 0.53 & 0.42 & \multicolumn{1}{c|}{0.43} & 0.4 & 0.42 & \multicolumn{1}{c|}{0.42} & 0.26 & 0.19 & 0.00 \\ \hline
\textbf{Avg F1-score} & \textbf{0.81} & \textbf{0.57} & \multicolumn{1}{c|}{\textbf{0.50}} & \textbf{0.81} & \textbf{0.82} & \multicolumn{1}{c|}{\textbf{0.82}} & \textbf{0.54} & \textbf{0.55} & \textbf{0.74} \\ \hline
\end{tabular}
\label{table__results__RT__threshold}
\end{table*}

\begin{figure*}[!ht]
	\centering
	\begin{subfigure}[b]{0.3\textwidth}
	\includegraphics[width=\textwidth,keepaspectratio]{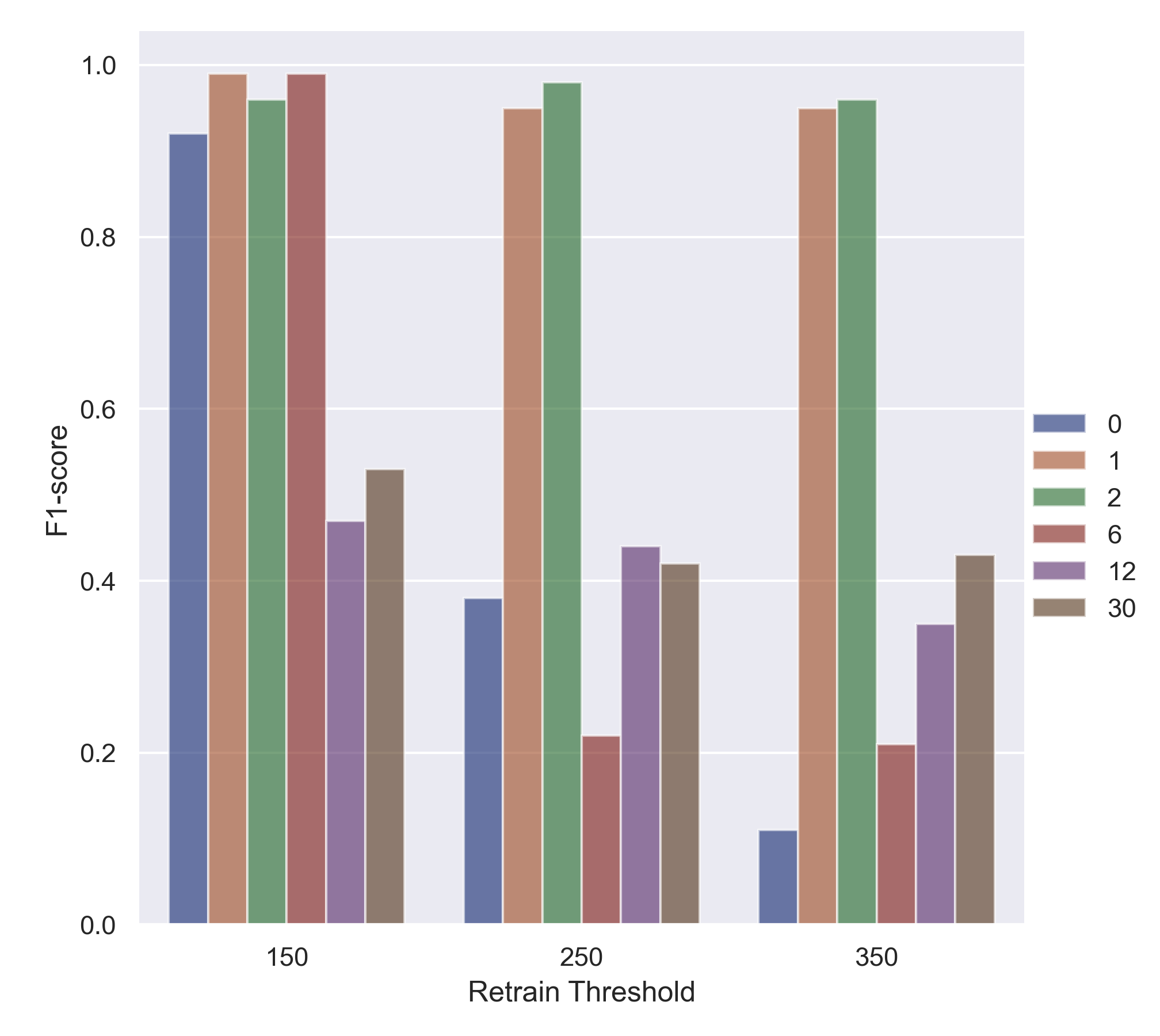} 
	\caption{Bar chart for M3 using A7}\label{figure__use_FUS__M3__A7__Threshold}
	\end{subfigure}
	\begin{subfigure}[b]{0.3\textwidth}
	\includegraphics[width=\textwidth,keepaspectratio]{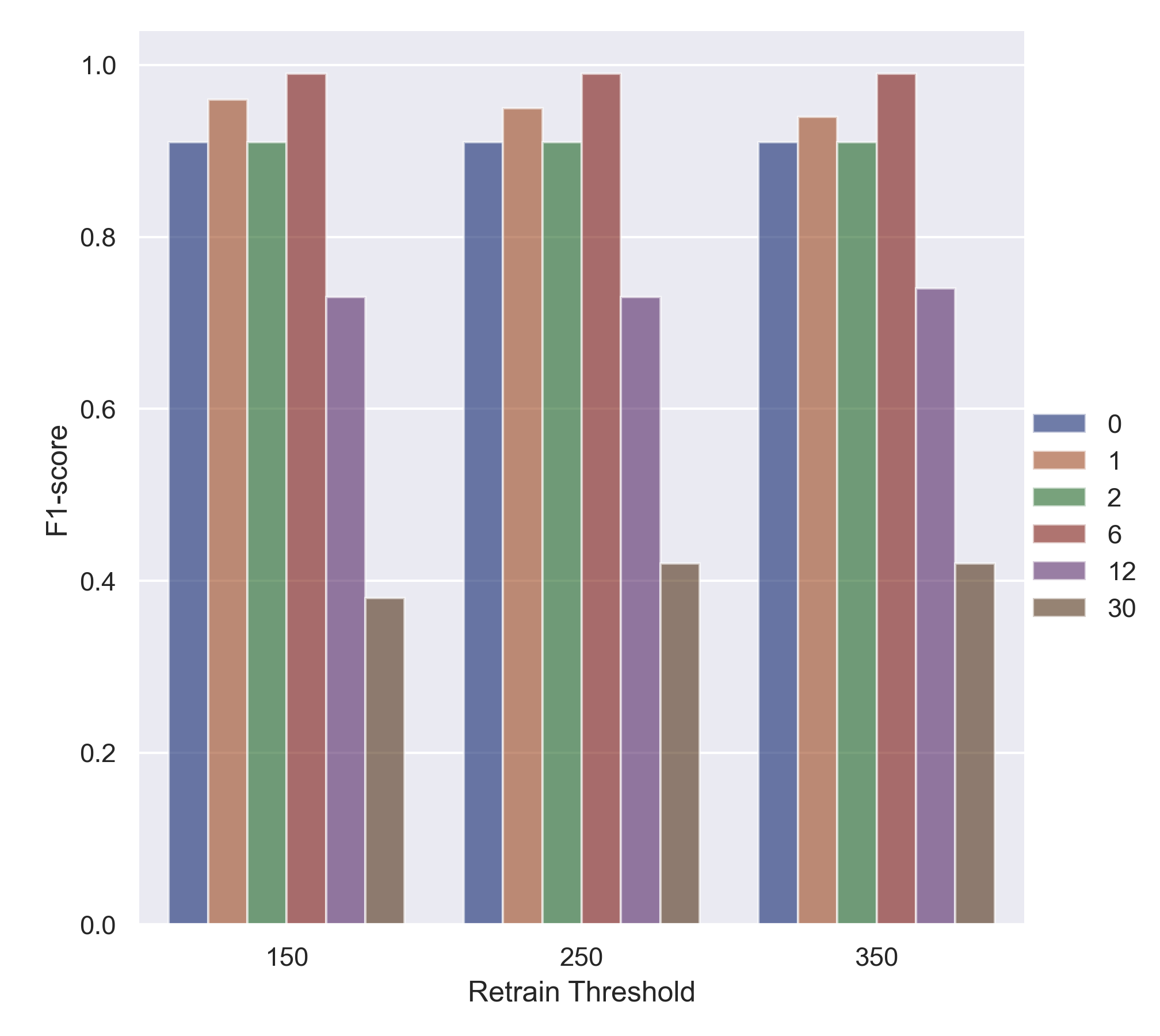} 
	\caption{Bar chart for M3 using A8}\label{figure__FUS__M3__A8__Threshold}
	\end{subfigure}
	\begin{subfigure}[b]{0.3\textwidth}
	\includegraphics[width=\textwidth,keepaspectratio]{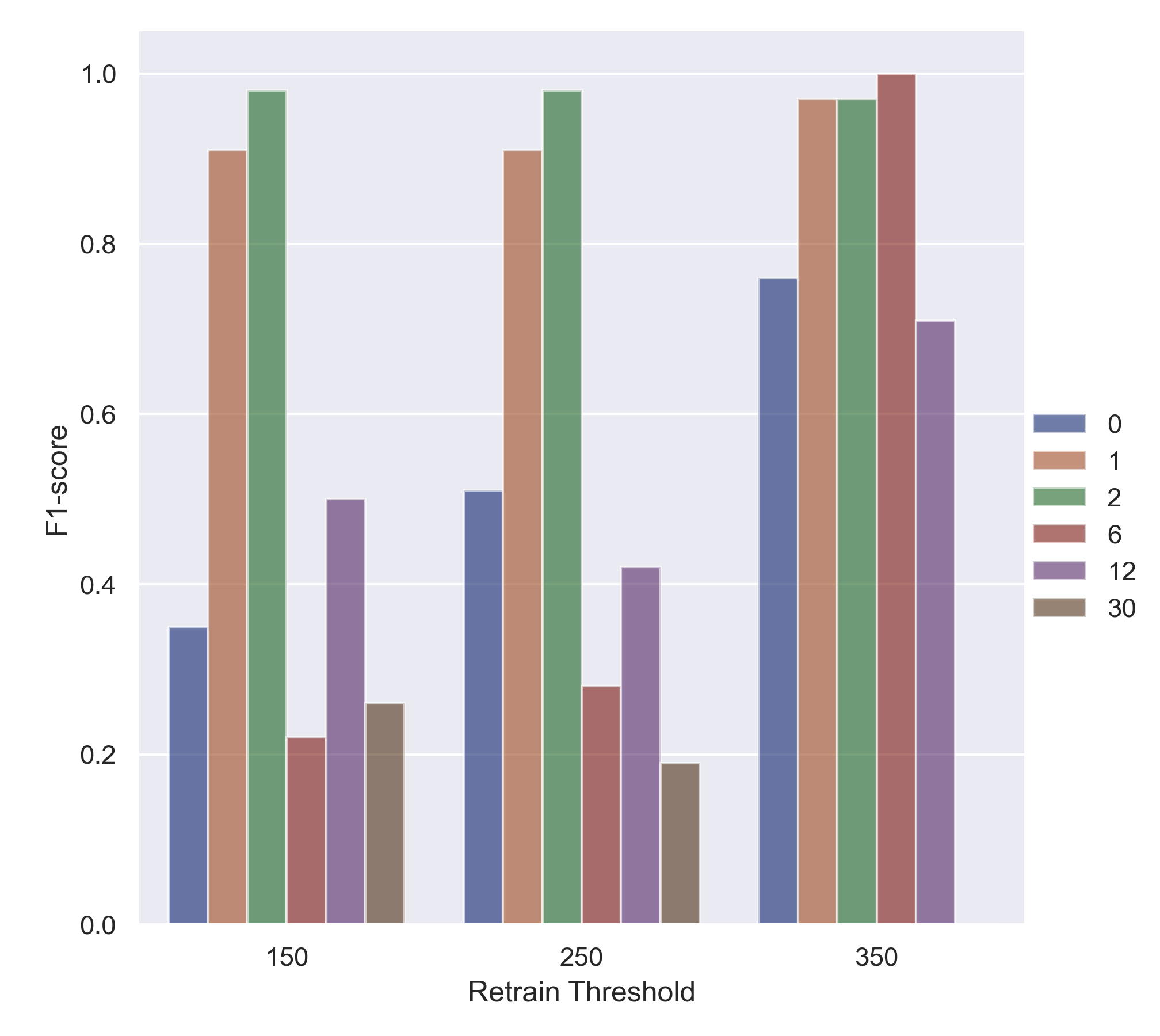} 
	\caption{Bar chart for M3 using A15}\label{figure__FUS__M3__A15__Threshold}
	\end{subfigure}
	\caption{F1-score results after retraining for the ECs BIN M4, MC M3, and MC M5: (a)-(c) Bar plots for the known cases (0,1,2,6,12) and the new case (30, corresponding to the injected anomaly 7). The plots represent the ECs results using memory size 20 and patience 15, while varying the threshold (150,250,350). 
  }\label{figure__results__RT__threshold}
\end{figure*}

Fig. \ref{figure__results__RT__threshold} displays the EC M3 performance for each class while effectuating variations on the threshold size (150,250,350) for the anomalies (7,8,15).  
The best performance corresponds to the anomaly (8), in which the EC M3 detects the fault cases (0,1,2,6,12) often correctly, and it has limited anomaly detection. In contrast, the EC M3 presents a lower performance while applying the anomalies (7,15).

\paragraph{Effects of the window size}
Table \ref{table__results__RT__memory} presents the F1-scores of the MC ECs M3, M4, and M5 trained with the fault cases (0,1,2,6,12). The retraining parameters threshold size and detection patience are fixed, with values of $th=250$ and $pt=15$, respectively.
The MC EC M3 presented average F1-scores higher than 0.84 using window size (10,50) for the anomaly (7). Alternatively, the MC EC M3 presented average F1-scores higher than 0.72 for the anomaly (8) using the window size (20,50). In contrast, the MC EC M3 presented higher results using a window size $ws=50$ with an average F1-score of 0.74 for the anomaly (15), in comparison with the values of 0.50 and 0.55, which correspond to the window sizes (150, 250), respectively.

\begin{table*}[!ht]
\centering
\caption{Anomaly detection results of MC EC M3 using the fault cases (0,1,2,6,12), the anomalies (7,8,15), window size variations (10,20,50), threshold (250), patience (15), and F1-score.}
\begin{tabular}{c|lllllllll}
\hline
\multirow{2}{*}{\textbf{Fault}} & \multicolumn{9}{c}{\textbf{MC EC M3}} \\ \cline{2-10} 
 & \multicolumn{3}{c|}{\textbf{A7}} & \multicolumn{3}{c|}{\textbf{A8}} & \multicolumn{3}{c}{\textbf{A15}} \\ \hline
 & \textbf{me=10} & \textbf{me=20} & \multicolumn{1}{l|}{\textbf{me=50}} & \textbf{me=10} & \textbf{me=20} & \multicolumn{1}{l|}{\textbf{me=50}} & \textbf{me=10} & \textbf{me=20} & \textbf{me=50} \\ \cline{2-10} 
0 & 0.97 & 0.38 & \multicolumn{1}{l|}{0.97} & 0.8 & 0.91 & \multicolumn{1}{l|}{0.85} & 0.54 & 0.51 & 0.76 \\
1 & 0.99 & 0.95 & \multicolumn{1}{l|}{0.99} & 0.9 & 0.95 & \multicolumn{1}{l|}{0.93} & 0.86 & 0.91 & 0.98 \\
2 & 0.98 & 0.98 & \multicolumn{1}{l|}{0.97} & 1 & 0.91 & \multicolumn{1}{l|}{0.95} & 0.78 & 0.98 & 0.98 \\
6 & 0.99 & 0.22 & \multicolumn{1}{l|}{0.99} & 0.2 & 0.99 & \multicolumn{1}{l|}{0.99} & 0.24 & 0.28 & 0.99 \\
12 & 0.43 & 0.44 & \multicolumn{1}{l|}{0.42} & 0.4 & 0.73 & \multicolumn{1}{l|}{0} & 0.51 & 0.42 & 0.56 \\
30 & 0.72 & 0.42 & \multicolumn{1}{l|}{0.72} & 0.3 & 0.42 & \multicolumn{1}{l|}{0.61} & 0.09 & 0.19 & 0.14 \\ \hline
\textbf{Avg F1-score} & \multicolumn{1}{c}{\textbf{0.85}} & \multicolumn{1}{c}{\textbf{0.57}} & \multicolumn{1}{c|}{\textbf{0.84}} & \multicolumn{1}{c}{\textbf{0.62}} & \multicolumn{1}{c}{\textbf{0.82}} & \multicolumn{1}{c|}{\textbf{0.72}} & \multicolumn{1}{c}{\textbf{0.50}} & \multicolumn{1}{c}{\textbf{0.55}} & \multicolumn{1}{c}{\textbf{0.74}} \\ \hline
\end{tabular}
\label{table__results__RT__memory}
\end{table*}

\begin{figure*}[!ht]
	\centering
	\begin{subfigure}[b]{0.3\textwidth}
	\includegraphics[width=\textwidth,keepaspectratio]{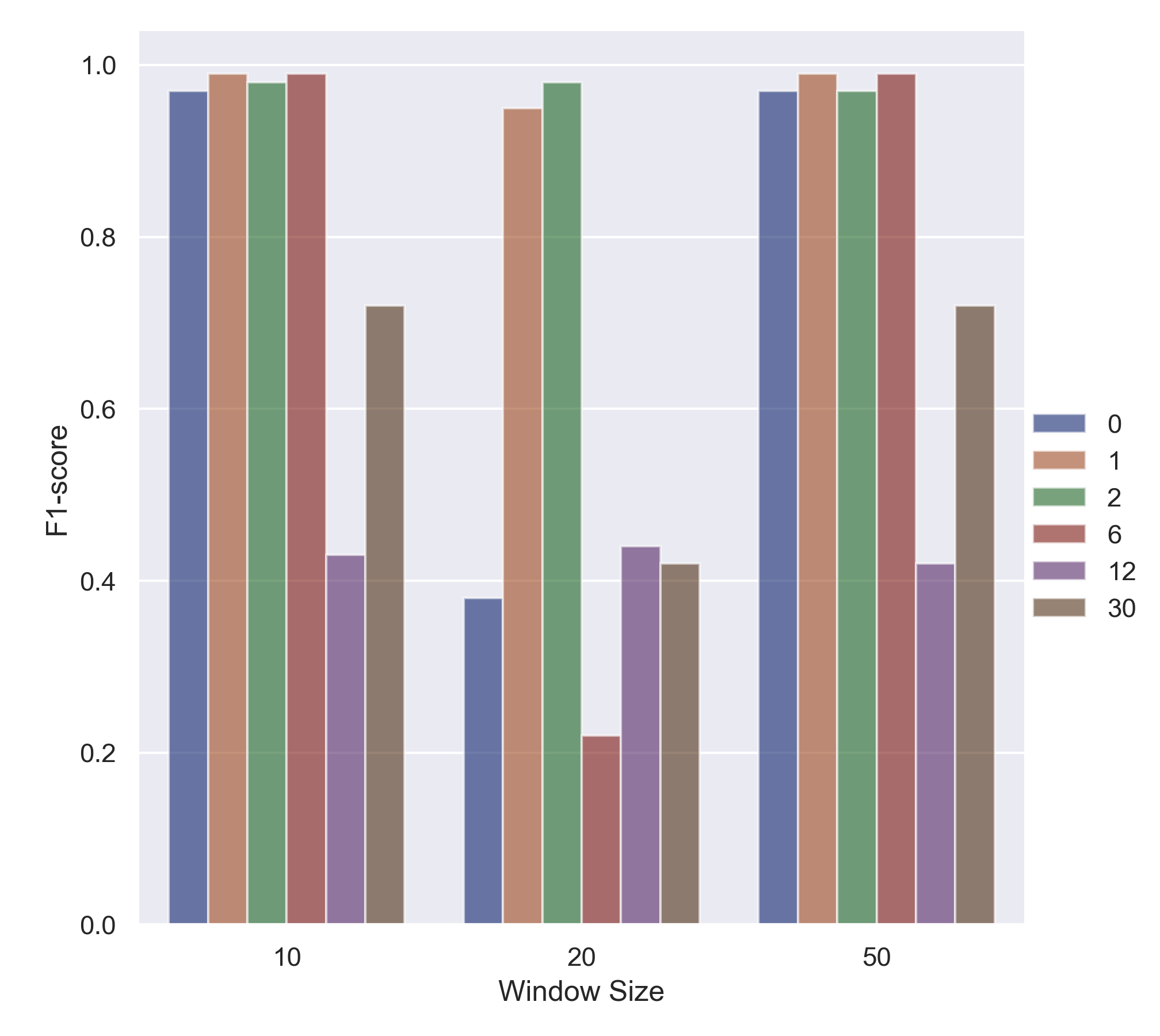} 
	\caption{Bar chart for M3 using A7}\label{figure__use_FUS__M3__A7__Memory}
	\end{subfigure}
	\begin{subfigure}[b]{0.3\textwidth}
	\includegraphics[width=\textwidth,keepaspectratio]{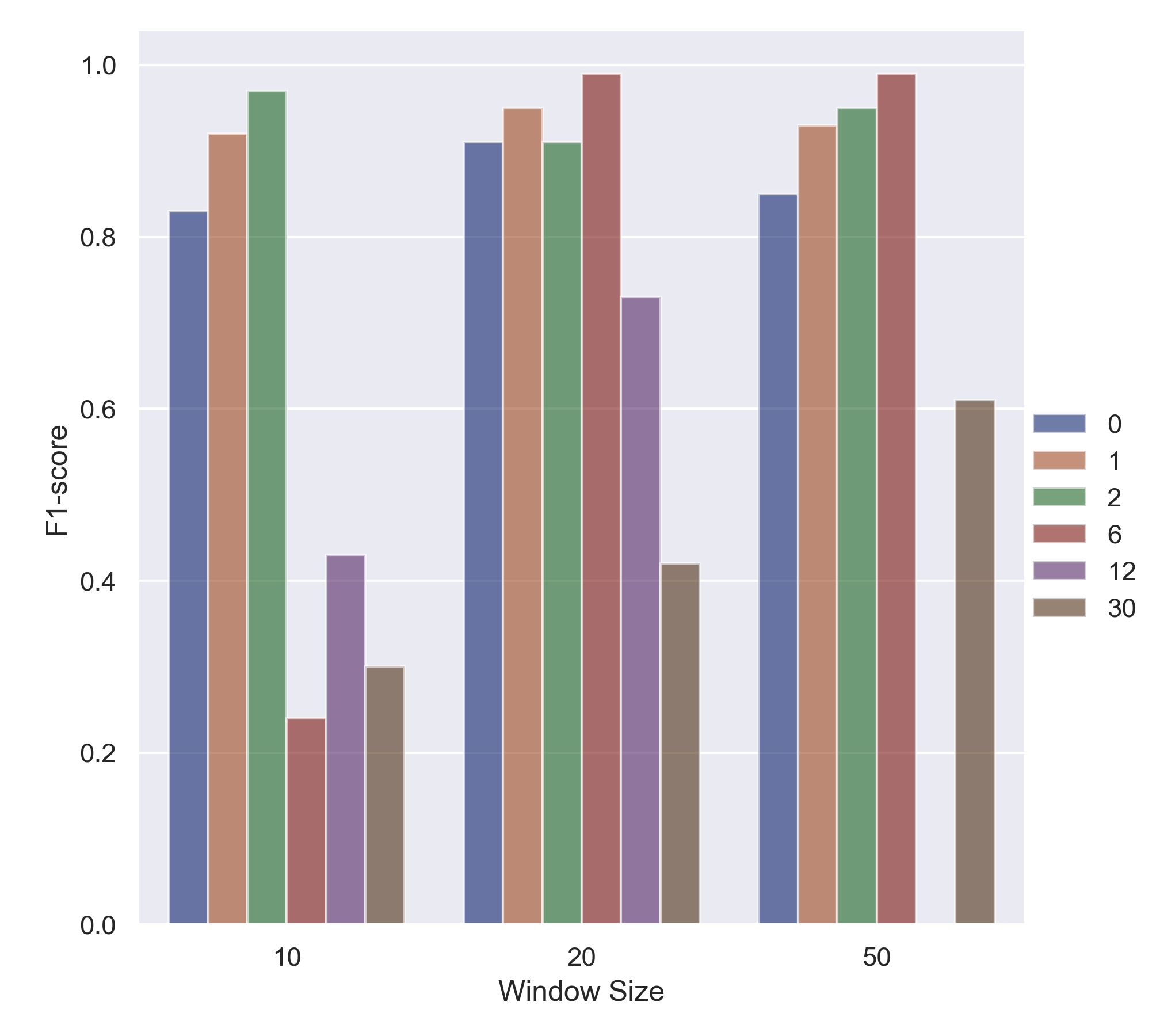} 
	\caption{Bar chart for M3 using A8}\label{figure__FUS__M3__A8__Memory}
	\end{subfigure}
	\begin{subfigure}[b]{0.3\textwidth}
	\includegraphics[width=\textwidth,keepaspectratio]{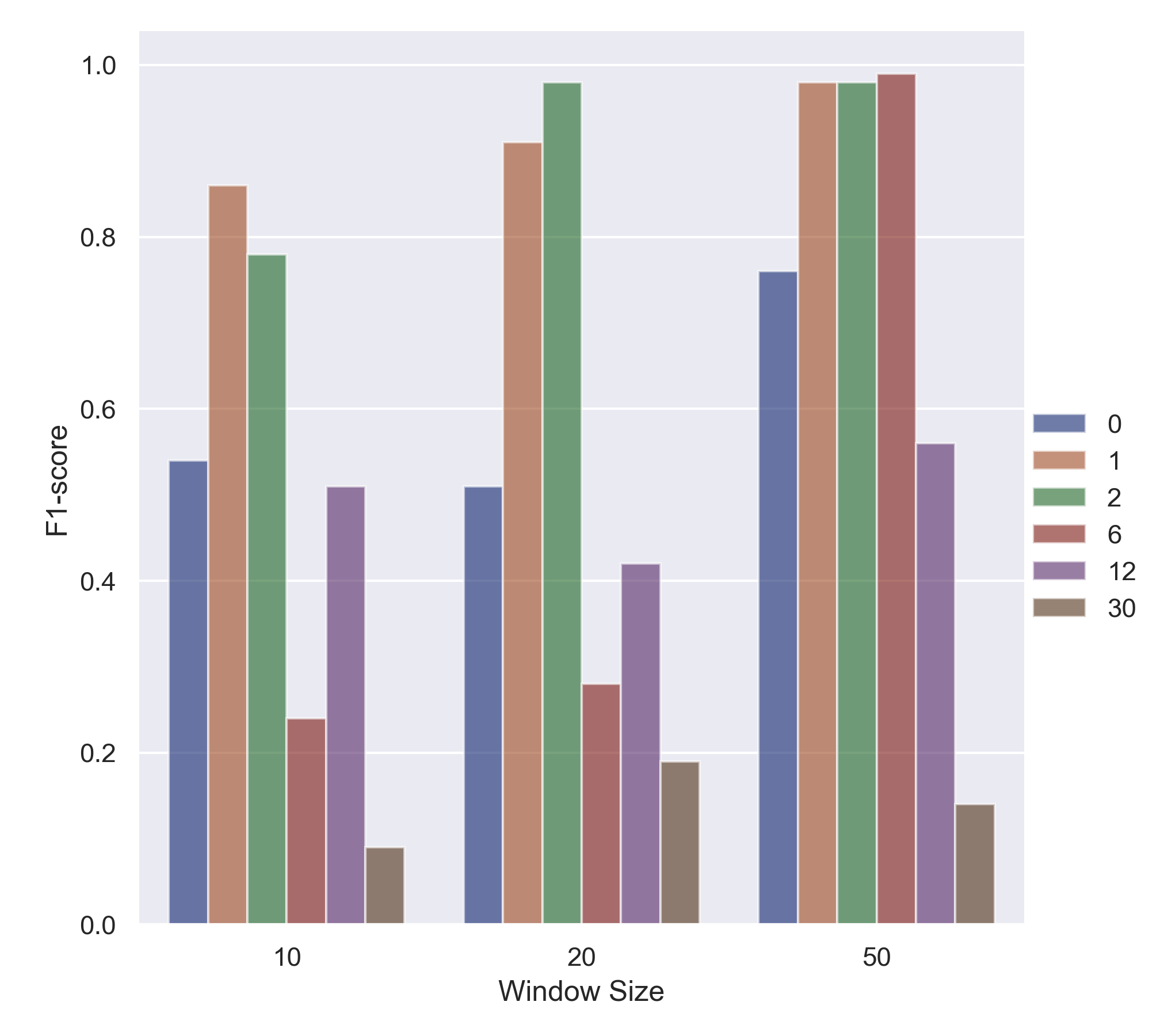} 
	\caption{Bar chart for M3 using A15}\label{figure__FUS__M3__A15__Memory}
	\end{subfigure}
	\caption{F1-score results after retraining for the ECs BIN M4, MC M3, and MC M5: (a)-(c) Bar plots for the known cases (0,1,2,6,12) and the new case (30, corresponding to the injected anomaly 7). The plots represent the ECs results using threshold 250 and patience 15, while varying the memory size (10,20,50). 
  }\label{figure__results__RT__memory}
\end{figure*}

Fig. \ref{figure__results__RT__memory} displays the EC M3 performance for each class while effectuating variations on the memory size (10,20,50) for the anomalies (7,8,15). The best performance corresponds to the anomaly (8) using a window size $ws=20$, in which the EC M3 detects the fault cases (0,1,2,6,12) mostly correct, and it has a limited anomaly detection. In contrast, the EC M3 presents a lower performance while applying the anomalies (7,15).

\paragraph{Effects of the detection patience}
Table \ref{table__results__RT__patience} presents the F1-scores of the MC ECs M3, M4, and M5 trained with the fault cases (0,1,2,6,12). The retraining parameters threshold size and window size are fixed with values of $th=250$ and $ws=20$, respectively. 
The MC EC M3 presented an average F1-scores of 0.84 using detection patience of $pt=5$ and $pt=30$, respectively, compared to the average F1-score of 0.57 for $pt=15$. In the case of anomaly (8), the MC EC M3 presented higher results using detection patience $pt=15$ with an average F1-score of 0.82, in comparison with the values of 0.78 and 0.58, which correspond to the detection patience (5,30), respectively.
The MC EC M3 presented average F1-scores higher than 0.73 for the detection patience (5,30), while the average F1-score of 0.55 is obtained with the detection patience $pt=15$.

\begin{table*}[!ht]
\centering
\caption{Anomaly detection results of MC EC M3 using the fault cases (0,1,2,6,12), the anomalies (7,8,15), patience variations (5,15,30), threshold (250), memory size (20), and F1-score}
\begin{tabular}{c|ccccccccc}
\hline
\multirow{2}{*}{\textbf{Fault}} & \multicolumn{9}{c}{\textbf{MC EC M3}} \\ \cline{2-10} 
 & \multicolumn{3}{c|}{\textbf{A7}} & \multicolumn{3}{c|}{\textbf{A8}} & \multicolumn{3}{c}{\textbf{A15}} \\ \hline
 & \textbf{pt=5} & \textbf{pt=15} & \multicolumn{1}{c|}{\textbf{pt=30}} & \textbf{pt=5} & \textbf{pt=15} & \multicolumn{1}{c|}{\textbf{pt=30}} & \textbf{pt=5} & \textbf{pt=15} & \textbf{pt=30} \\ \cline{2-10} 
0 & 0.96 & 0.38 & \multicolumn{1}{c|}{0.97} & 0.9 & 0.91 & \multicolumn{1}{c|}{0.64} & 0.75 & 0.51 & 0.77 \\
1 & 0.99 & 0.95 & \multicolumn{1}{c|}{0.98} & 1 & 0.95 & \multicolumn{1}{c|}{0.85} & 0.98 & 0.91 & 0.95 \\
2 & 0.94 & 0.98 & \multicolumn{1}{c|}{0.97} & 0.9 & 0.91 & \multicolumn{1}{c|}{0.97} & 0.96 & 0.98 & 0.97 \\
6 & 0.99 & 0.22 & \multicolumn{1}{c|}{1} & 1 & 0.99 & \multicolumn{1}{c|}{0.24} & 0.99 & 0.28 & 0.99 \\
12 & 0.41 & 0.44 & \multicolumn{1}{c|}{0.42} & 0.5 & 0.73 & \multicolumn{1}{c|}{0.55} & 0.59 & 0.42 & 0.72 \\
30 & 0.72 & 0.42 & \multicolumn{1}{c|}{0.72} & 0.4 & 0.42 & \multicolumn{1}{c|}{0.25} & 0.14 & 0.19 & 0 \\ \hline
\textbf{Avg F1-score} & \textbf{0.84} & \textbf{0.57} & \multicolumn{1}{c|}{\textbf{0.84}} & \textbf{0.78} & \textbf{0.82} & \multicolumn{1}{c|}{\textbf{0.58}} & \textbf{0.74} & \textbf{0.55} & \textbf{0.73} \\ \hline
\end{tabular}
\label{table__results__RT__patience}
\end{table*}

\begin{figure*}[!ht]
	\centering
	\begin{subfigure}[b]{0.3\textwidth}
	\includegraphics[width=\textwidth,keepaspectratio]{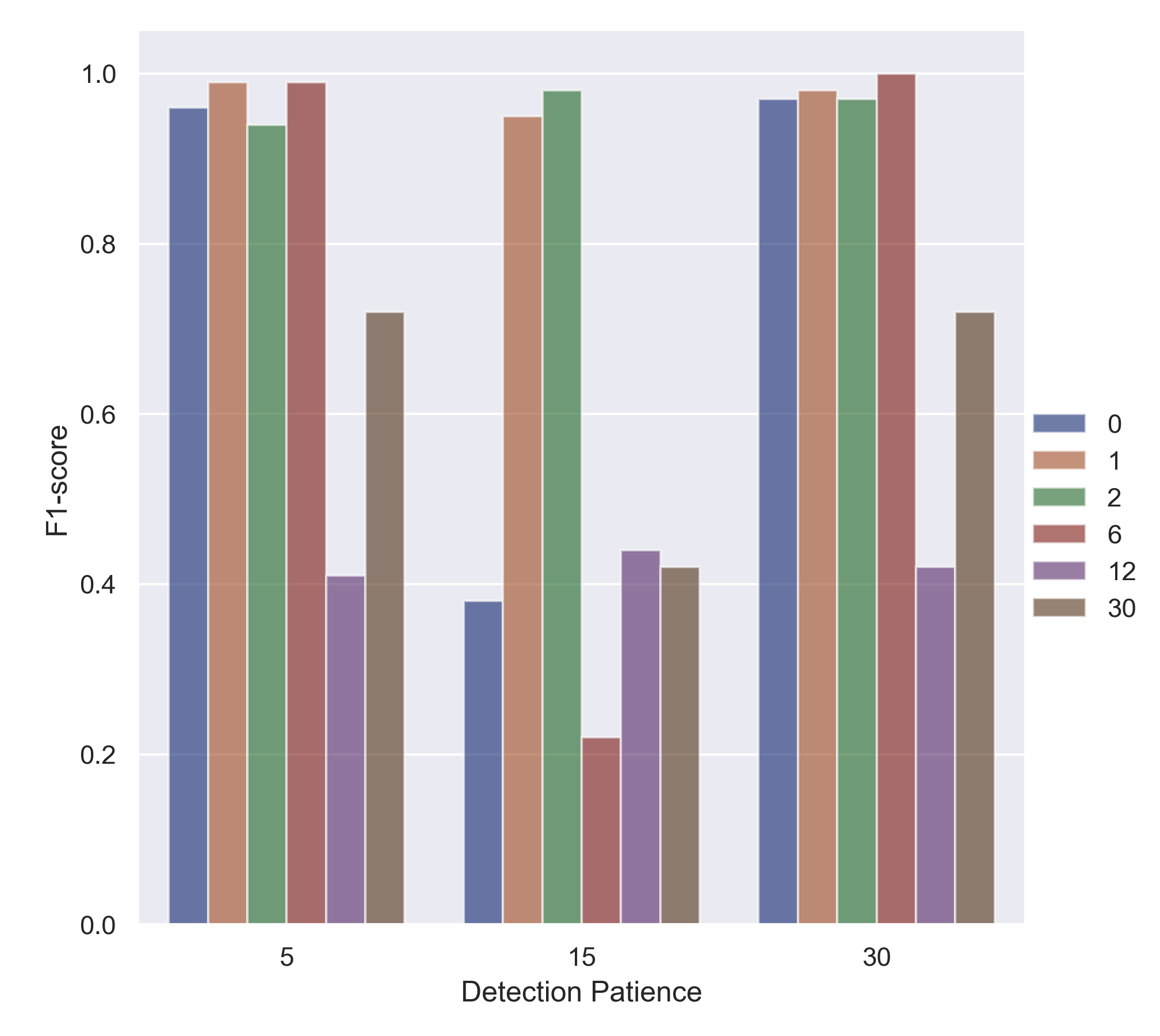} 
	\caption{Bar chart for M3 using A7}\label{figure__use_FUS__M3__A7__Patience}
	\end{subfigure}
	\begin{subfigure}[b]{0.3\textwidth}
	\includegraphics[width=\textwidth,keepaspectratio]{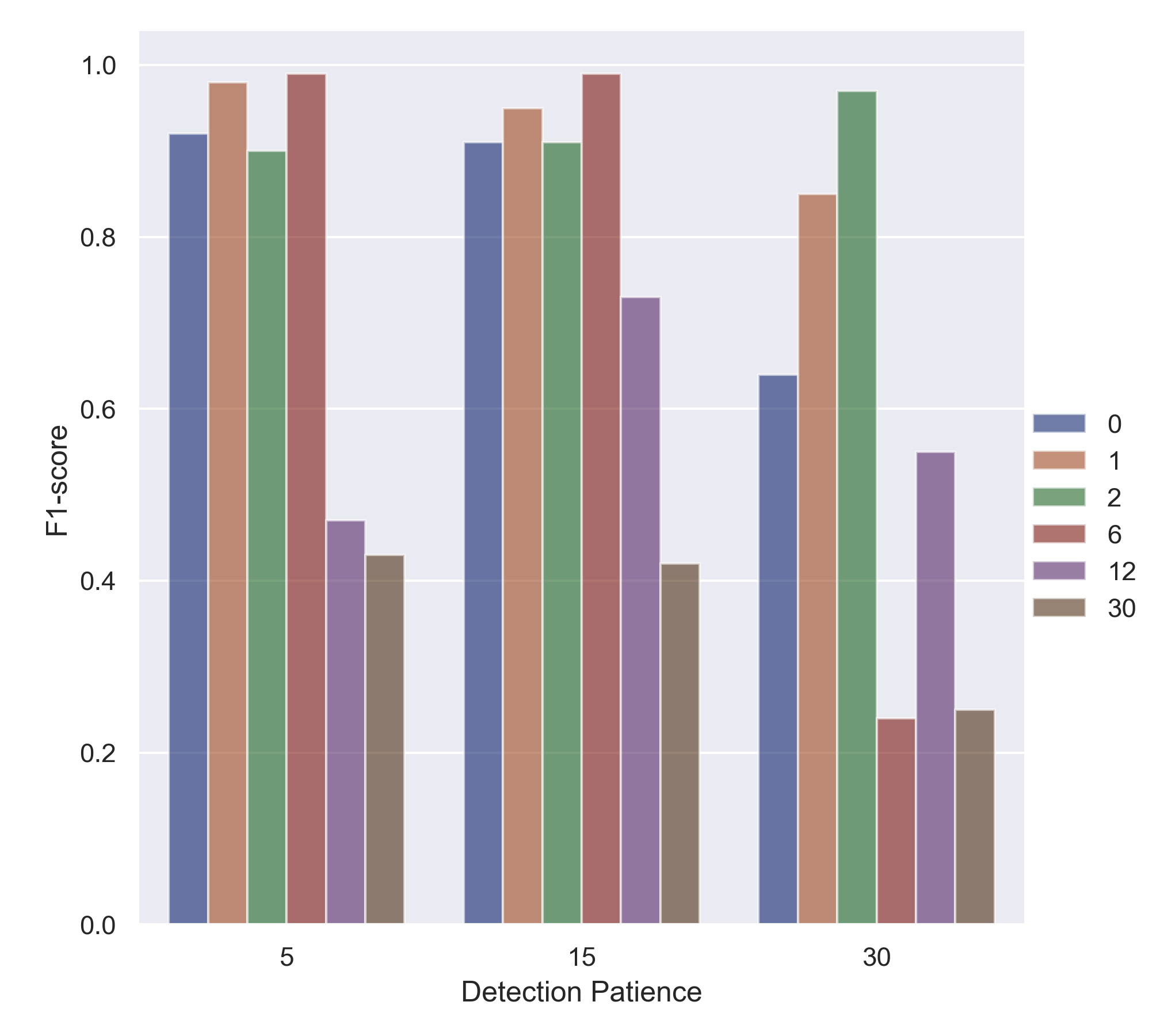} 
	\caption{Bar chart for M3 using A8}\label{figure__FUS__M3__A8__Patience}
	\end{subfigure}
	\begin{subfigure}[b]{0.3\textwidth}
	\includegraphics[width=\textwidth,keepaspectratio]{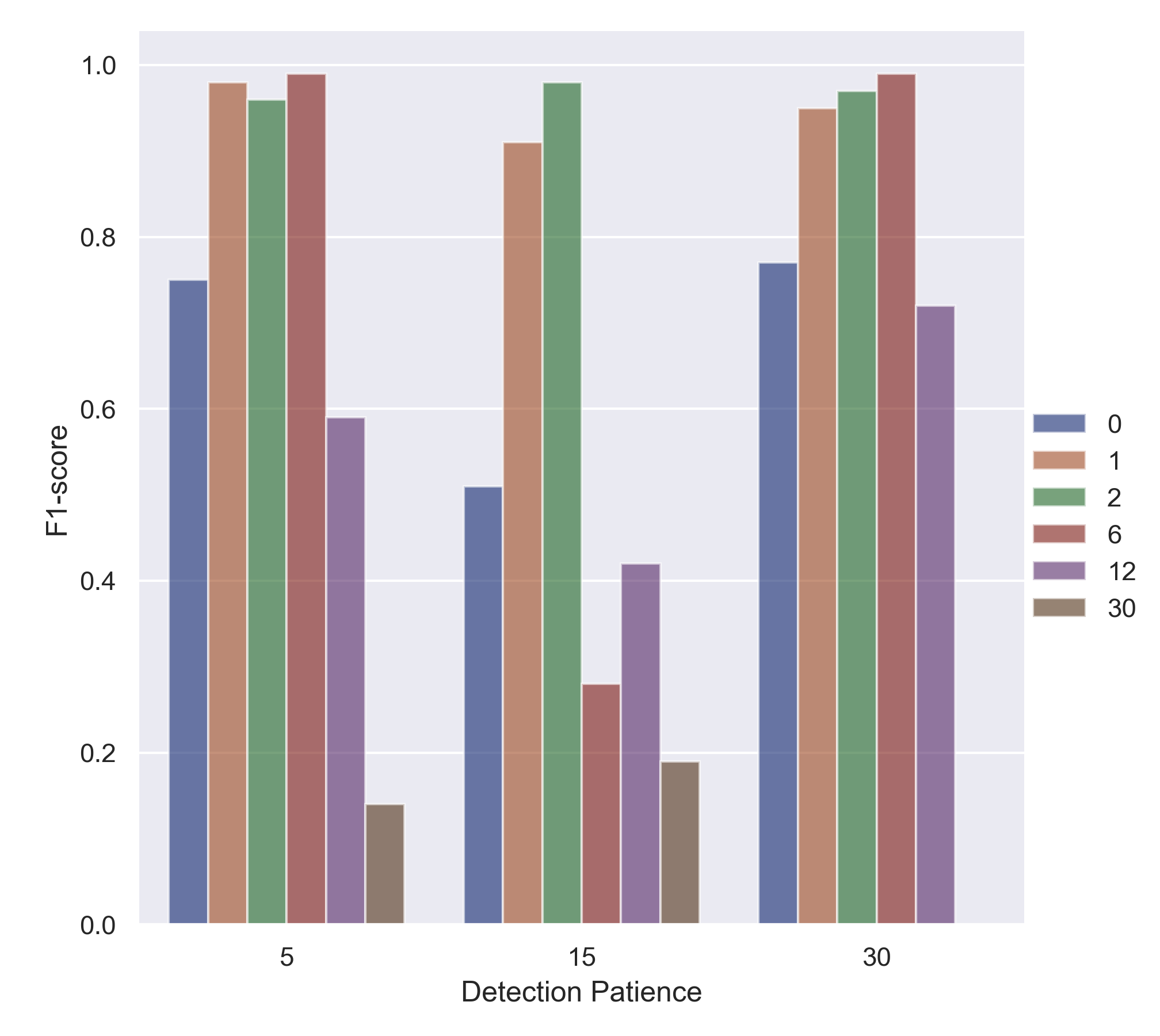} 
	\caption{Bar chart for M3 using A15}\label{figure__FUS__M3__A15__Patience}
	\end{subfigure}
	\caption{F1-score results after retraining for the ECs BIN M4, MC M3, and MC M5: (a)-(c) Bar plots for the known cases (0,1,2,6,12) and the new case (30, corresponding to the injected anomaly 7). The plots represent the ECs results using threshold $th=250$ and window size $me=20$, while varying the patience (5,15,30). 
  }\label{figure__results__RT__patience}
\end{figure*}

Fig. \ref{figure__results__RT__patience} displays the EC M3 performance for each class while effectuating variations on the detection patience (5,15,30) for the anomalies (7,8,15). The best performance corresponds to the anomaly (8) using detection patience $pt=15$, in which the EC M3 detects the fault cases (0,1,2,6,12) mostly correct and has a limited anomaly detection. In contrast, the EC M3 presents a lower performance while applying the anomalies (7,15).

Finally, we present the performance of the ECs with the \textit{tuned retraining parameters}. 
Table \ref{table__results__RT__tuned} presents the F1-scores of the MC ECs M3, M4, and M5 retrained with the fault cases (0,1,2,6,12) and the respective anomaly. In this case, the anomalies cases are all fault cases except for the original training cases. The retraining dataset contains the original fault cases and the detected data from the anomaly (unknown fault case from the data). The retraining parameters are threshold size $th=250$, window size $ws=20$, and detection patience $pt=15$.
The MC ECs M3, M4 and M5 present comparable results with an average F1-score of 0.39, 0.42, and 0.42, respectively. 
The MC EC M3 detected the anomalies (7,11) with F1-scores higher or equal to 0.55 and the anomalies (9,13,17) with F1-scores higher or equal to 0.34 and less than 0.42. The MC EC M4 detected the anomalies (8,14,17) with F1-scores higher or equal to 0.67 and the anomalies (7,10,11,15) with F1-scores higher or equal to 0.38 and less than 0.54.  
Alternatively, the EC M5 detected the anomalies (14,18) with F1-scores higher or equal to 0.68 and the anomalies (7,11,15,17,20) with F1-scores higher or equal to 0.31 and less than 0.54.    

\begin{table}[!ht]
\centering
\caption{Classification results of the RT ECs after retraining using all the fault cases, and F1-score. The retraining parameters are threshold size $th=250$, window size $ws=20$, and detection patience $pt=15$.}
\begin{tabular}{c|lll}
\hline
\multirow{2}{*}{\textbf{Fault}} & \multicolumn{3}{c}{\textbf{RT MC EC (0,1,2,6,12)}} \\ \cline{2-4} 
 & \multicolumn{1}{c}{\textbf{M3}} & \multicolumn{1}{c}{\textbf{M4}} & \multicolumn{1}{c}{\textbf{M5}} \\ \hline
1 & 0.98 & 0.99 & 0.99 \\
2 & 0.99 & 0.99 & 0.98 \\
3 & 0.00 & 0.02 & 0.01 \\
4 & 0.00 & 0.03 & 0.01 \\
5 & 0.00 & 0.18 & 0.13 \\
6 & 1.00 & 1.00 & 1.00 \\
7 & 0.72 & 0.54 & 0.31 \\
8 & 0.29 & 0.71 & 0.44 \\
9 & 0.34 & 0.00 & 0.00 \\
10 & 0.27 & 0.38 & 0.22 \\
11 & 0.55 & 0.51 & 0.54 \\
12 & 0.95 & 0.95 & 0.95 \\
13 & 0.35 & 0.20 & 0.15 \\
14 & 0.26 & 0.77 & 0.76 \\
15 & 0.13 & 0.39 & 0.31 \\
16 & 0.26 & 0.09 & 0.20 \\
17 & 0.42 & 0.67 & 0.51 \\
18 & 0.06 & 0.02 & 0.68 \\
19 & 0.20 & 0.07 & 0.05 \\
20 & 0.28 & 0.23 & 0.52 \\
21 & 0.07 & 0.13 & 0.01 \\ \hline
\textbf{Avg   F1-score} & \textbf{0.39} & \textbf{0.42} & \textbf{0.42} \\ \hline
\end{tabular}
\label{table__results__RT__tuned}
\end{table}

\subsection{Comparison with Literature}
Though the current approach can automatically update the models while detecting unknown fault cases from the data, the stored data to retrain the models might be insufficient for some fault cases. Thus, the stored data for some fault cases might not capture the essential patterns to identify the condition. 
In contrast, the contributions of literature presented in the comparison consider all the extent of the testing data.

Table \ref{table__results__RT__comparison_1} compares the anomaly detection results between the proposed approach and the literature. The multiclass ECs M3, M4, and M5 are originally trained using the fault cases (0,1,2,6,12). The testing data consists of the fault cases (3,9,15,21), which represent unknown conditions to the ECs. For this purpose, each EC is retrained with one fault case at a time.
We use the F1-score as a performance metric to compare the proposed approach with other literature contributions. It is essential to mention that the MC EC H5-2 from a previous work \cite{ArevaloIbrahim2023} uses the full extent of testing data, as well in the case of Top-K DCCA \cite{ChadhaIslam2021}.
The results of the ECs M3, M4, and M5 present lower results with average F1-scores of 20.36\%, 3.50\%, and 2.59\%, respectively. The results of H5-2 and Top-K DCCA present general scores of 63.69\% and 50.04\%, respectively.
Only M3 presents a score of 31.07\% for the fault case 21, which still lies under the better performance results of H5-2 and Top-K DCCA with scores of 63.1\% and 50.05\%, respectively.

\begin{table*}[!ht]
\centering
\caption{Classification results of the ECs after retraining using all the fault cases, and F1-score. The retraining parameters are threshold size $th=250$, window size $ws=20$, and detection patience $pt=15$.}
\begin{tabular}{c|ccc|c|c}
\hline
\multirow{2}{*}{\textbf{Fault}} & \multicolumn{3}{c|}{\textbf{RT MC EC (0,1,2,6,12)}} & \multicolumn{1}{l|}{\textbf{MC EC (0,1,2,6,12)} \cite{ArevaloIbrahim2023}} & \multirow{2}{*}{\textbf{Top-K DCCA} \cite{ChadhaIslam2021}} \\ \cline{2-5}
 & \textbf{M3} & \textbf{M4} & \textbf{M5} & \textbf{H5-2} &  \\ \hline
3 & 0.00 & 7.43 & 0.00 & 64.3 & 53.82 \\
9 & 28.87 & 6.21 & 5.81 & 63.01 & 52.31 \\
15 & 21.49 & 0.00 & 0.45 & 64.35 & 43.98 \\
21 & 31.07 & 0.35 & 4.09 & 63.1 & 50.05 \\ \hline
\textbf{Avg F1-score} & \textbf{20.36} & \textbf{3.50} & \textbf{2.59} & \textbf{63.69} & \textbf{50.04} \\ \hline
\end{tabular}
\label{table__results__RT__comparison_1}
\end{table*}

Table \ref{table__results__RT__comparison_2} compares the anomaly detection results between our approach and the literature. We use the FDR to compare our results with the literature results. The retrained MC ECs M3, M4, and M5 present lower results with average FDR scores of 53.02\%, 41.68\%, and 35.04\%, respectively. The MC ECs M3 and H3-4 present FDR scores of 87.97\% and 73.76\%, respectively. The approaches DPCA-DR, AAE, and MOD-PLS have FDR scores of 83.51\%, 78.55\%, and 83.83\%, respectively. 

\begin{table*}[!ht]
\centering
\caption{Classification results of the ECs after retraining using all the fault cases, and FDR. The retraining parameters are threshold size $th=250$, window size $ws=20$, and detection patience $pt=15$.}
\begin{tabular}{c|ccc|cc|c|c|c}
\hline
\multirow{2}{*}{\textbf{Fault}} & \multicolumn{3}{c|}{\textbf{RT MC EC (0,1,2,6,12)}} & \multicolumn{2}{c|}{\textbf{MC EC (0,1,2,6,12)}} & \multirow{2}{*}{\textbf{DPCA-DR}} & \multirow{2}{*}{\textbf{AAE}} & \multirow{2}{*}{\textbf{MOD-PLS}} \\ \cline{2-6}
 & \textbf{M3} & \textbf{M4} & \textbf{M5} & \textbf{M3} & \textbf{H3-4} &  &  &  \\ \hline
1 & 98.50 & 98.50 & 98.50 & 93.5 & 82.13 & 99.6 & 100 & 99.88 \\
2 & 97.75 & 97.63 & 97.75 & 94.75 & 83.63 & 98.5 & 99 & 98.75 \\
3 & 0.00 & 8.50 & 0.00 & 91.88 & 75 & 2.1 & 34.88 & 18.73 \\
4 & 0.00 & 3.38 & 0.00 & 89.75 & 76.5 & 99.8 & 98.62 & 99.88 \\
5 & 0.00 & 12.75 & 11.00 & 90.63 & 75.25 & 99.9 & 55 & 99.88 \\
6 & 99.88 & 99.88 & 99.88 & 97.5 & 64.75 & 99.9 & 100 & 99.88 \\
7 & 96.25 & 82.75 & 23.13 & 88.25 & 77.5 & 99.9 & 100 & 99.88 \\
8 & 47.25 & 43.63 & 0.00 & 87 & 76.38 & 98.5 & 97.88 & 98.5 \\
9 & 41.75 & 4.75 & 4.25 & 90.75 & 74.88 & 2 & 33.62 & 12.11 \\
10 & 47.00 & 24.00 & 31.25 & 88.13 & 75.13 & 95.6 & 74 & 91.01 \\
11 & 45.88 & 52.38 & 0.00 & 92.5 & 73.63 & 96.5 & 82 & 83.15 \\
12 & 92.50 & 91.75 & 92.00 & 81.5 & 80.5 & 99.8 & 99.75 & 99.75 \\
13 & 35.88 & 35.50 & 0.00 & 84.38 & 71.38 & 95.8 & 96.25 & 95.38 \\
14 & 77.63 & 86.88 & 100.00 & 91 & 71.75 & 99.8 & 100 & 99.88 \\
15 & 36.88 & 0.00 & 0.50 & 91.25 & 66.25 & 38.5 & 31.25 & 23.22 \\
16 & 83.50 & 0.00 & 7.25 & 90.25 & 76.25 & 97.6 & 64.75 & 94.26 \\
17 & 0.00 & 69.00 & 98.75 & 91.25 & 75.75 & 97.6 & 96 & 97 \\
18 & 62.63 & 12.63 & 5.75 & 37.63 & 56 & 90.5 & 95 & 91.14 \\
19 & 20.13 & 2.88 & 5.00 & 92 & 74.5 & 97.1 & 53.87 & 94.13 \\
20 & 51.75 & 48.25 & 57.63 & 89.38 & 72.63 & 90.8 & 78.62 & 91.26 \\
21 & 78.38 & 0.25 & 3.13 & 94.13 & 69.25 & 53.9 & 59 & 72.66 \\ \hline
\textbf{Avg F1-score} & \textbf{53.02} & \textbf{41.68} & \textbf{35.04} & \textbf{87.97} & \textbf{73.76} & \textbf{83.51} & \textbf{78.55} & \textbf{83.83} \\ \hline
\end{tabular}
\label{table__results__RT__comparison_2}
\end{table*}

\subsection{Discussion}
The ECs improved the anomaly detection capability after implementing the \textit{window size}. In the case of the MC EC M5, the general F1-score improved from 0.6 to 0.65 using a window of $w=50$ for the latest score. In the case of H5-2, the results are remarkable, in which the general F1-score score improved from 0.63 to 0.88 using a window of $w=50$ for the latest score.    
However, a side effect of the window is a delay effect on the ensemble prediction, which is reflected while comparing Fig. \ref{fig__window__prediction__w0} and Fig. \ref{fig__window__prediction__w50}. 

There are remarkable effects on the EC M3 performance while doing variations on the retraining parameters, namely, threshold size, window size, and detection patience. The results are mixed, and the average performance depends on the studied anomaly. However, from the results, it is possible to identify that a \textit{threshold} of $Th=150$ presented the best average results for anomaly 7. In contrast, a threshold of $Th=350$ presented the best results for anomaly 8.
Alternatively, the plots of Fig. \ref{figure__results__RT__threshold} visualize the performance of each class while doing variations on the threshold. The MC EC M3 presents an overall good performance while applying anomaly 8, in which the EC classifies the known cases mostly correctly and has a limited detection of the anomaly. In contrast, the anomaly detection feature decreases the performance of the known fault cases for some fault cases, which is visually represented in Fig \ref{figure__use_FUS__M3__A7__Threshold} while applying anomaly 7.  
\textit{Variation of the window size} reported favorable average performance results for a window of $me=50$ while considering all the anomalies (7,8,15). In contrast, the plots of Fig. \ref{figure__results__RT__memory} show that the best results correspond to the window size $me=20$ while applying anomaly 8, in which the EC classifies known cases properly, and it has a limited detection of the anomaly. Likewise the threshold experiments, a similar effect of decreasing classification performance of the known cases is detected.
Generally, a \textit{patience} of $pt=5$ presented the best average results for all the anomalies (7,8,15). In contrast, the plots of Fig. \ref{figure__FUS__M3__A8__Patience} show that the best results correspond to the patience $pt=15$ while applying anomaly 8, in which likewise the window size experiment, the EC classifies the known cases mostly correctly, and it has a limited detection of the anomaly. Likewise the threshold and window size experiments, the performance of the EC is affected by some faults while using the anomaly detection approach.

The retrained MC ECs M3, M4, and M5 presented mixed results using the same retraining parameters: threshold size $th=250$, window size $me=20$, and patience $pt=15$. The average F1-score of M3, M4, and M5 presented values of 0.67, 0.44, and 0.42, respectively. 
For this configuration, M3 presented the best results, however, it is important to remark that the anomalies (14-19) are not detected. In contrast, M4 and M5 detected the faults (14,17,18), though the average scores are lower than M3 scores.

The performance of the retrained MC ECs presented mixed results. For instance, the EC M3 detected the anomaly cases (4,5,7,11,13) with FDR scores higher than 77\% and the anomalies (10,20,21) with FDR scores higher than 53\%. However, the results of the retrained ECs presented a lower performance than other literature contributions. The average FDR scores of M3, M4, and M5 are 50.18\%, 43.60\%, and 51.44\%. It is important to remark that the retrained models only use 250 samples as training data (only 52\% of the available data), in which other fault cases might be included as a side effect of the parameter patience.   

\section{Use Case: Production Assessment using INFUSION on a Bulk Good System}\label{section__usecase__assistance}

As described in section \ref{section__information__fusion}, the approach's novelty is a methodology for the information fusion of data-based and knowledge-based models. The methodology primarily uses a novel framework for combining $n$ number of models using DSET.

This section presents the results of the information fusion approach and an ablation study considering the different system configurations. The system configurations consist of the detection system using: the data-based model, the knowledge model, or a hybrid model (data-based model together with a knowledge model) using information fusion. 
We test the approach using a dataset of an industrial setup, namely, a bulk good system laboratory plant. We describe the testbed and the dataset. We present the results and a discussion of the findings. Fig. \ref{figure__F2__infusion__overview} displays the main blocks of this section: the data-based model (ECET), the knowledge-based model (KLAFATE), and the outer module for the information fusion of both models.


\subsection{Description of the Bulk Good System Laboratory Plant and Dataset}
The bulk good system (BGS) laboratory plant is an industrial setup used for testing production and fault detection experiments. The BGS consists of four stations that represent standard modules of a bulk good handling system on a small scale: loading, storing, filling, and weighing stations. A detailed description of the BGS and applications can be found in \cite{ArevaloPiolo2022} \cite{ArevaloNguyen2017}. 
The stations are built using state-of-the-art hardware regarding industrial controllers, communication protocols, sensors, and actors.
The BGS dataset contains 14055 rows of data, each containing 133 features and three classes. The features represent information about sensors, actors, and controllers.
The classes represent the different machine conditions, namely, low quality (LQ), low production (LP), and normal production (NP or the normal condition). Each class is associated with a failure mode (fm), which in this case, translates into LQ (fm1), LP (fm2), and NP (fm3). In the case of the class NP, it does not represent a failure mode but is represented in the same framework for consistency purposes of the knowledge model.



\subsection{Experiment Design}
This subsection presents the methodology followed for the ECET and INFUSION experiments using the BGS dataset. Besides, we describe the performance metric used to compare the experiments. 

\subsubsection{ECET using the BGS Data}
We followed the same methodology of \cite{ArevaloIbrahim2023} for the creation of MC ECs using the BGS data, which includes the pool of base classifiers, the grid of hyperparameters of each classifier, and the grid of hyperparameters for each EC. We used the data-based models: decision tree (DTR), K-nearest neighbors (KNN), AdaBoost (ADB), support vector machine (SVM), and naive Bayes (NBY).
For this purpose, we first trained the pool of classifiers using only ML models, which implies the search for the proper hyperparameters for each model. The second step is creating the ECs, using the EC hyperparameters. The last step presents the inference results of the ECs while injecting the BGS data.  

\subsubsection{INFUSION using the BGS data}
The knowledge-based model KEXT was presented in \cite{ArevaloPiolo2022}, in which we describe the knowledge rules. We only use the failure modes fm1, fm2, and fm3 for the INFUSION experiments.
We present a comparison table using knowledge, data fusion, and knowledge and data fusion models. The KEXT model represents the knowledge model. The data fusion models are represented by the ECET ECs models and a fusion of two data-based models. Lastly, the knowledge and data fusion models are represented by the combination of the SVM-KNN-KEXT models and the INFUSION models composed of an MC EC and the KEXT model.
        
\subsubsection{Performance Metrics}
We use the F1-score as the main performance metric to compare the different experiments. Panda et al. \cite{Panda2021} present a detailed description of the F1-score calculation.

\subsection{Results}
This subsection presents the results using the BGS data for the ECET and the INFUSION architectures. For this purpose, we present the F1-score results of the models or ECs. Besides, we display the confusion matrix, classification predictions, and uncertainty for the different architectures.

\subsubsection{ECET using the BGS Data}
The first is to train the pool of base classifiers, which we performed using the module grid search of scikit-learn. Table \ref{table__grid__hyperparameters__base_classifiers__BGS} presents the hyperparameters of the base classifiers trained with the cases (1,2,3), which corresponds to the failure modes (fm1, fm2, fm3), respectively.


\begin{table}[!ht]
\centering
\caption{Grid of hyperparameters for base classifiers using  the\textit{BGS} dataset and the cases (1,2,3) }
\begin{tabular}{l|l|c}
\hline
\multicolumn{1}{c|}{\multirow{2}{*}{\textbf{Model}}} & \multicolumn{1}{c|}{\multirow{2}{*}{\textbf{Hyperparameters}}} & \textbf{MC} \\ \cline{3-3} 
\multicolumn{1}{c|}{} & \multicolumn{1}{c|}{} & \textbf{1,2,3} \\ \hline
ADB & learning\_rate & 0.01 \\
 & n\_estimators & 10 \\ \hline
DTR & criterion & entropy \\
 & max\_depth & 10 \\ \hline
KNN & criterion & manhattan \\
 & n\_neighbors & 7 \\
 & weights & distance \\ \hline
NBY & - & NP \\ \hline
SVM & C & 1000 \\
 & gamma & 0.01 \\
 & kernel & rbf \\ \hline
\end{tabular}
\label{table__grid__hyperparameters__base_classifiers__BGS}
\end{table}

The next step is applying the ECET methodology to find the most performing MC ECs. We obtained the ML-based MC ECs, shown in Table \ref{table__EC__hyperparameters__BGS}. The hyperparameters expert (Exp), diversity (Div), version of diversity (Ver), and pre-cut (PC) are set to False. 

\begin{table*}[!ht]
\centering
\caption{EC hyperparameters using the \textit{BGS} dataset and the cases (1,2,3). }
\begin{tabular}{c|c|c|llll|l}
\hline
\textbf{Type} & \textbf{Dataset} & \textbf{EC} & \multicolumn{1}{c}{\textbf{Exp}} & \multicolumn{1}{c}{\textbf{Div}} & \multicolumn{1}{c}{\textbf{Ver}} & \multicolumn{1}{c|}{\textbf{PC}} & \multicolumn{1}{c}{\textbf{Pool of base classifiers}} \\ \hline
\multirow{3}{*}{MC} & \multirow{3}{*}{1,2,3} & M3 & False & False & False & False & ADB-DTR-KNN \\
 &  & M4 & False & False & False & False & ADB-DTR-KNN-NBY \\
 &  & M5 & False & False & False & False & ADB-DTR-KNN-NBY-SVM \\ \hline
\end{tabular}
\label{table__EC__hyperparameters__BGS}
\end{table*}

Table \ref{table__results__inference__classification__BGS} presents the F1-scores of the MC ECs M3, M4, and M5 and the base MC classifiers DTR, KNN, and ADB. The MC ECs M3, M4, and M5 present the same average F1-score of 1.00, whereas the base classifiers DTR, KNN, and ADB have values of 1.0, 1.0, and 0.96, respectively. 

\begin{table}[!ht]
\centering
\caption{Inference results of selected MC ECs using the cases (1,2,3) of the \textit{BGS} dataset, and F1-score.}
\begin{tabular}{cccc|ccc}
\hline
\multicolumn{1}{c|}{\multirow{2}{*}{\textbf{Fault}}} & \multicolumn{3}{c|}{\textbf{MC EC}} & \multicolumn{3}{c}{\textbf{INDIV}} \\ \cline{2-7} 
\multicolumn{1}{c|}{} & \textbf{M3} & \textbf{M4} & \textbf{M5} & \textbf{DTR} & \textbf{KNN} & \textbf{ADB} \\ \hline
\multicolumn{1}{c|}{1} & 1 & 1 & 1 & 1 & 1 & 1 \\
\multicolumn{1}{c|}{2} & 1 & 1 & 1 & 1 & 1 & 0.97 \\
\multicolumn{1}{c|}{3} & 0.99 & 0.99 & 0.99 & 1 & 0.99 & 0.91 \\ \hline
\multicolumn{1}{l|}{Avg F1-score} & 1 & 1 & 1 & 1 & 1 & 0.96 \\
\hline
\end{tabular}
\label{table__results__inference__classification__BGS}
\end{table}

\begin{figure*}[!htbp]
	\centering
      \begin{subfigure}[b]{0.3\textwidth}
    \includegraphics[width=\textwidth,keepaspectratio]{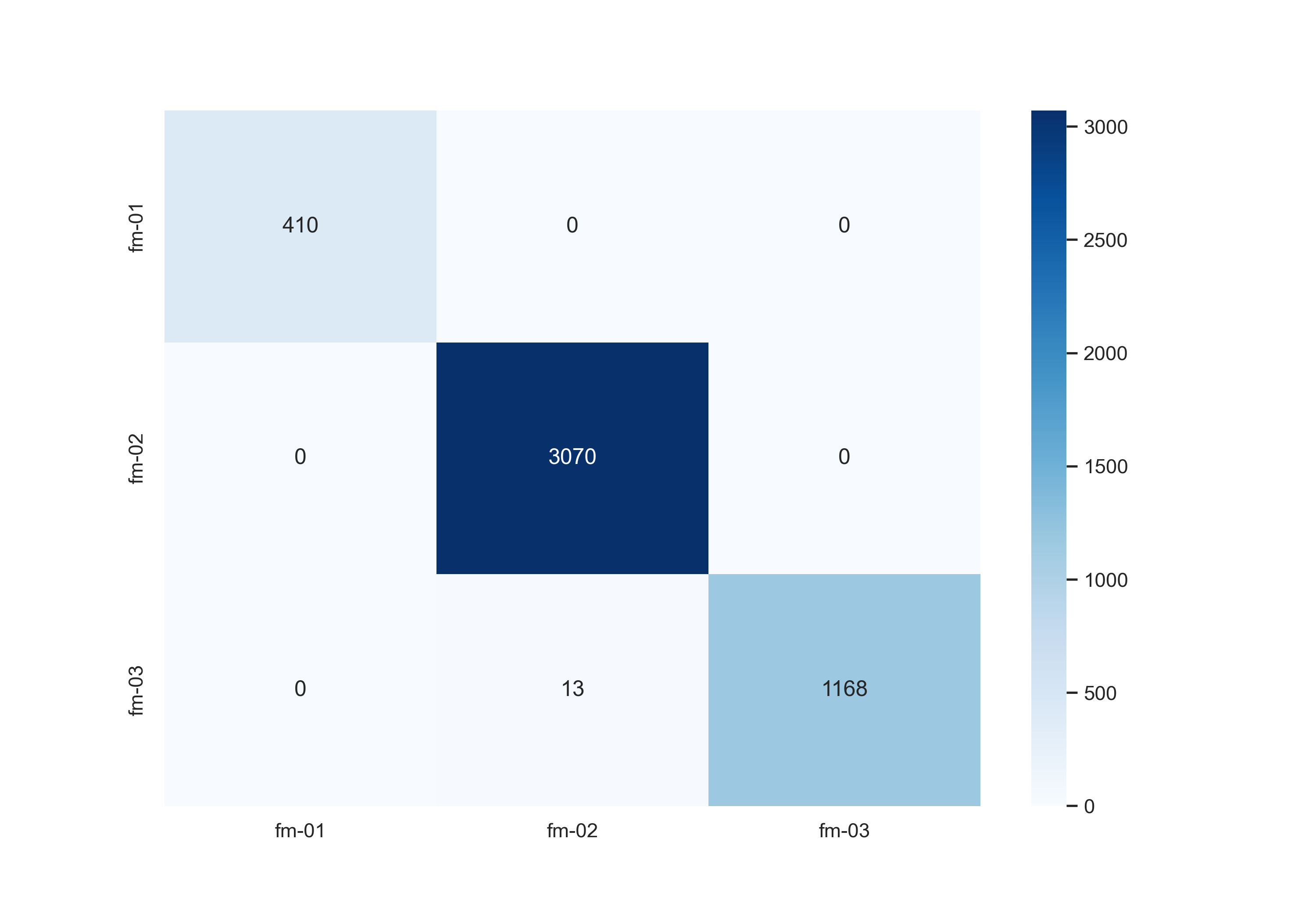}
    \caption{CM of M3}\label{Fig__ecet_M3__cm}
    \end{subfigure}
        \begin{subfigure}[b]{0.3\textwidth}    
	\includegraphics[width=\textwidth,keepaspectratio]{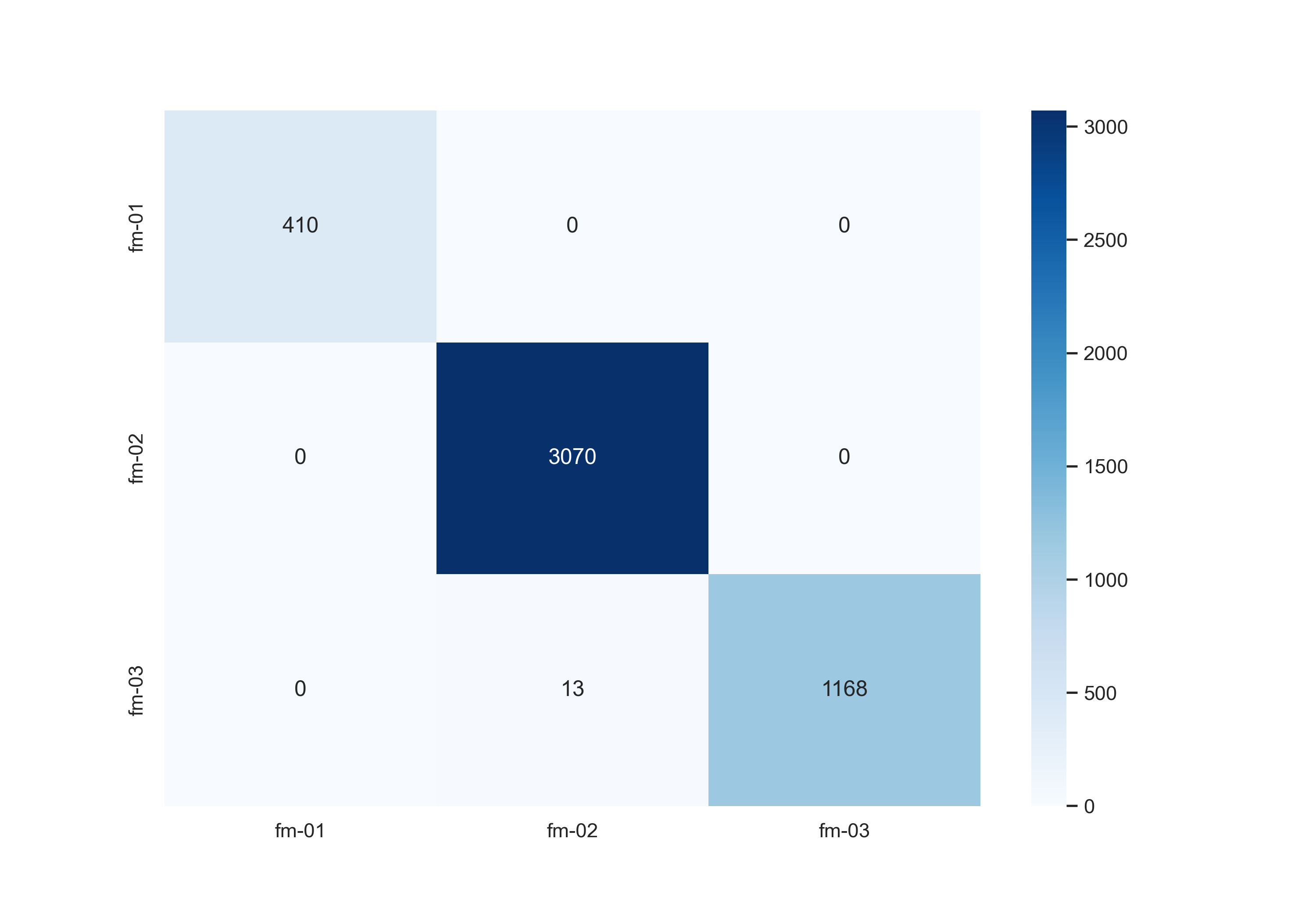}
	\caption{CM of M4}\label{Fig__ecet_M4__cm}
	\end{subfigure}
     \begin{subfigure}[b]{0.3\textwidth}
	\includegraphics[width=\textwidth,keepaspectratio]{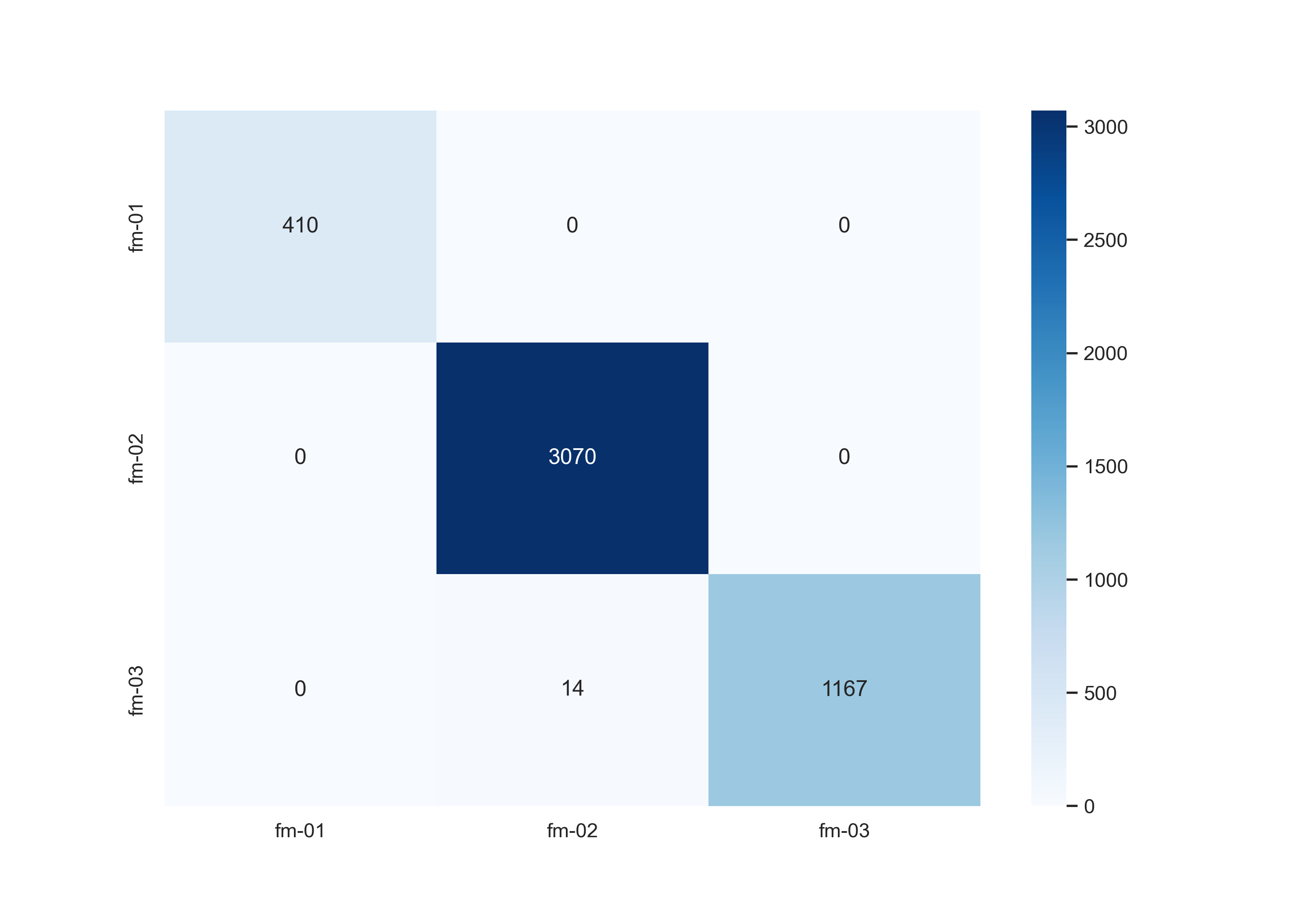}
	\caption{CM of M5}\label{Fig__ecet_M5__cm}
	\end{subfigure}
		~
    \begin{subfigure}[b]{0.3\textwidth}
    \includegraphics[width=\textwidth,keepaspectratio]{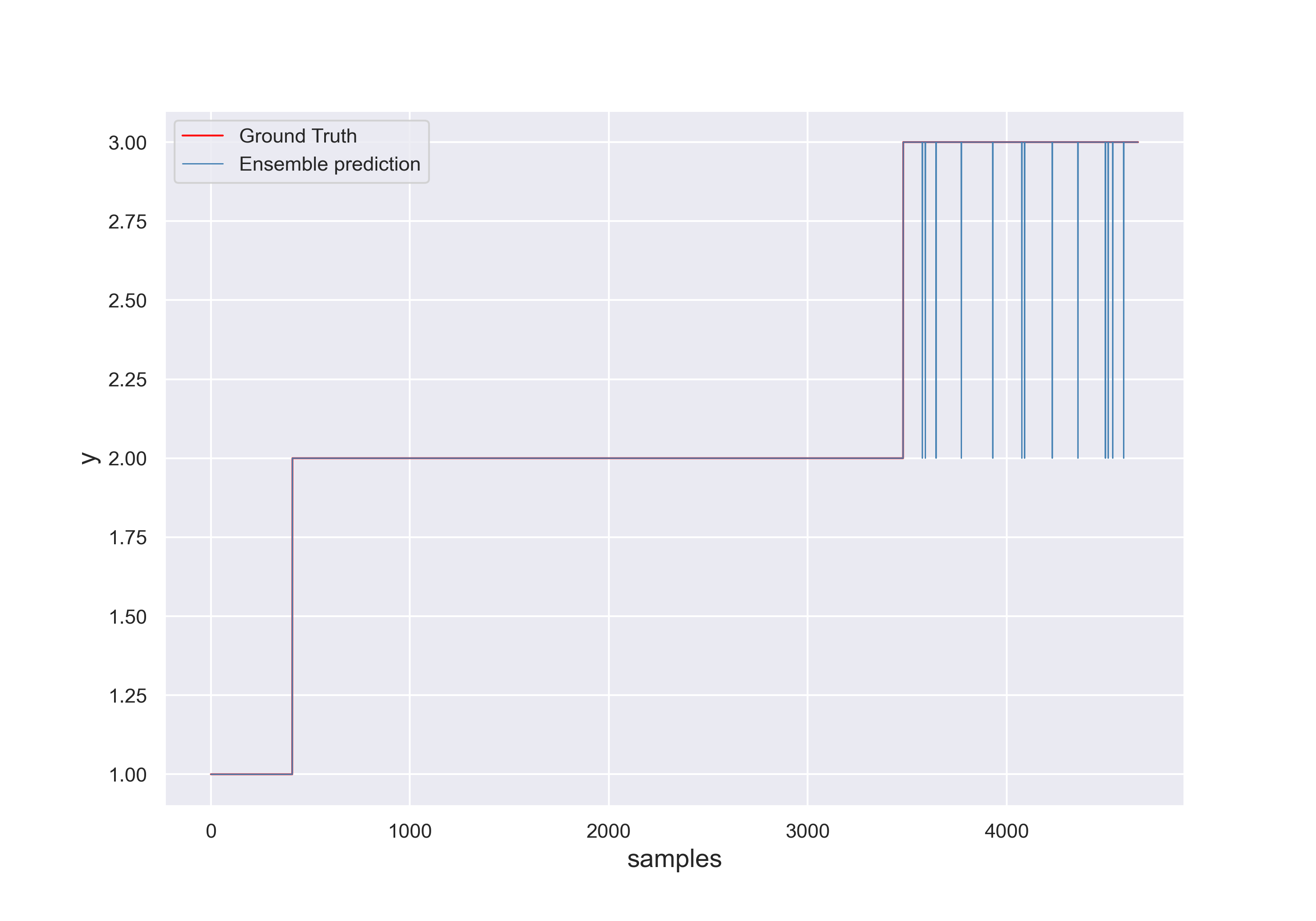}
    \caption{Predictions of M3}\label{Fig__ecet_M3__Classific__Prediction}
    \end{subfigure}
        \begin{subfigure}[b]{0.3\textwidth}    
	\includegraphics[width=\textwidth,keepaspectratio]{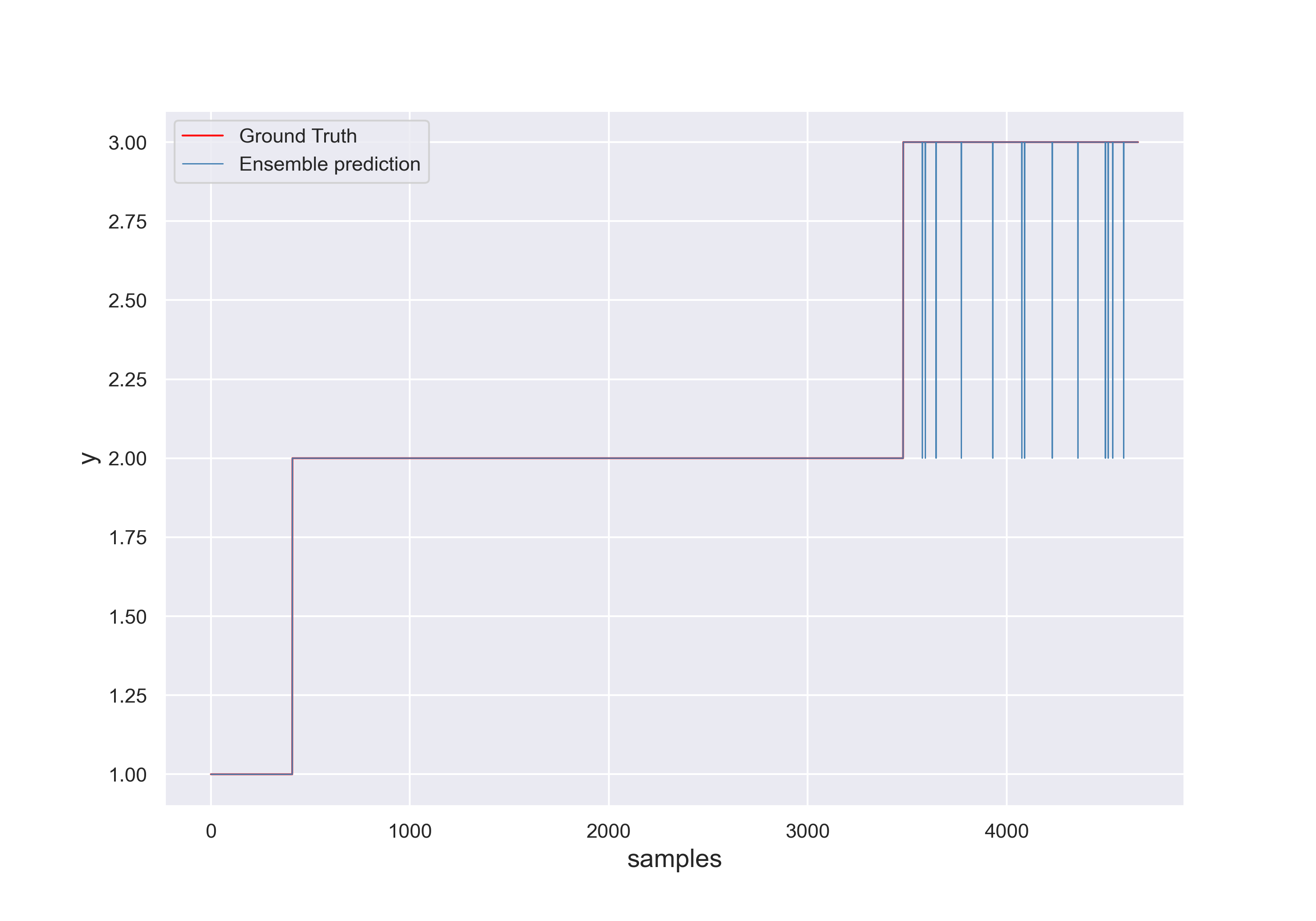}
	\caption{Predictions of M4}\label{Fig__ecet_M4__Classific__Prediction}
	\end{subfigure}
     \begin{subfigure}[b]{0.3\textwidth}
	\includegraphics[width=\textwidth,keepaspectratio]{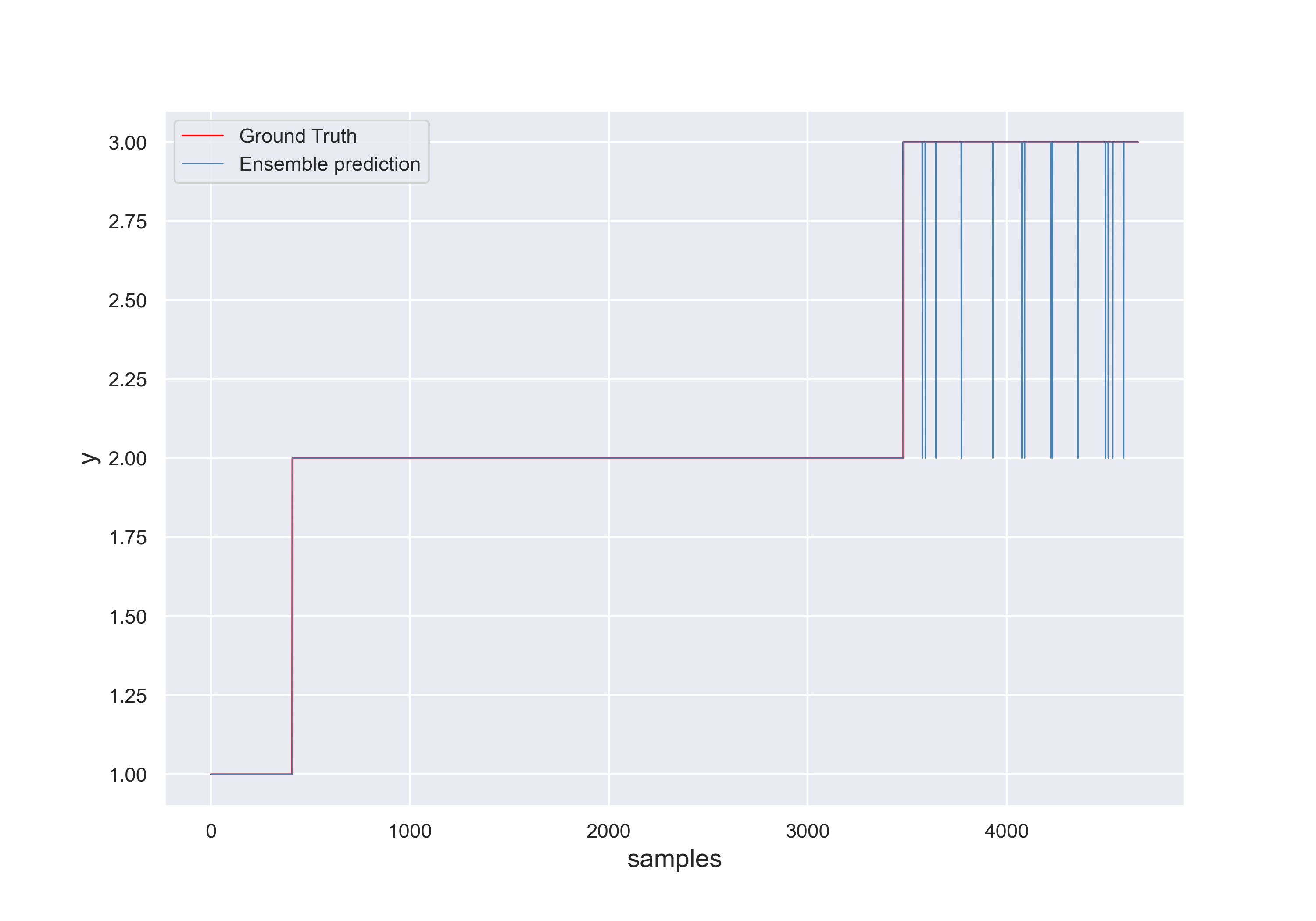}
	\caption{Predictions of M5}\label{Fig__ecet_M5__Classific__Prediction}
	\end{subfigure}
		~
	\begin{subfigure}[b]{0.3\textwidth}
	\includegraphics[width=\textwidth,keepaspectratio]{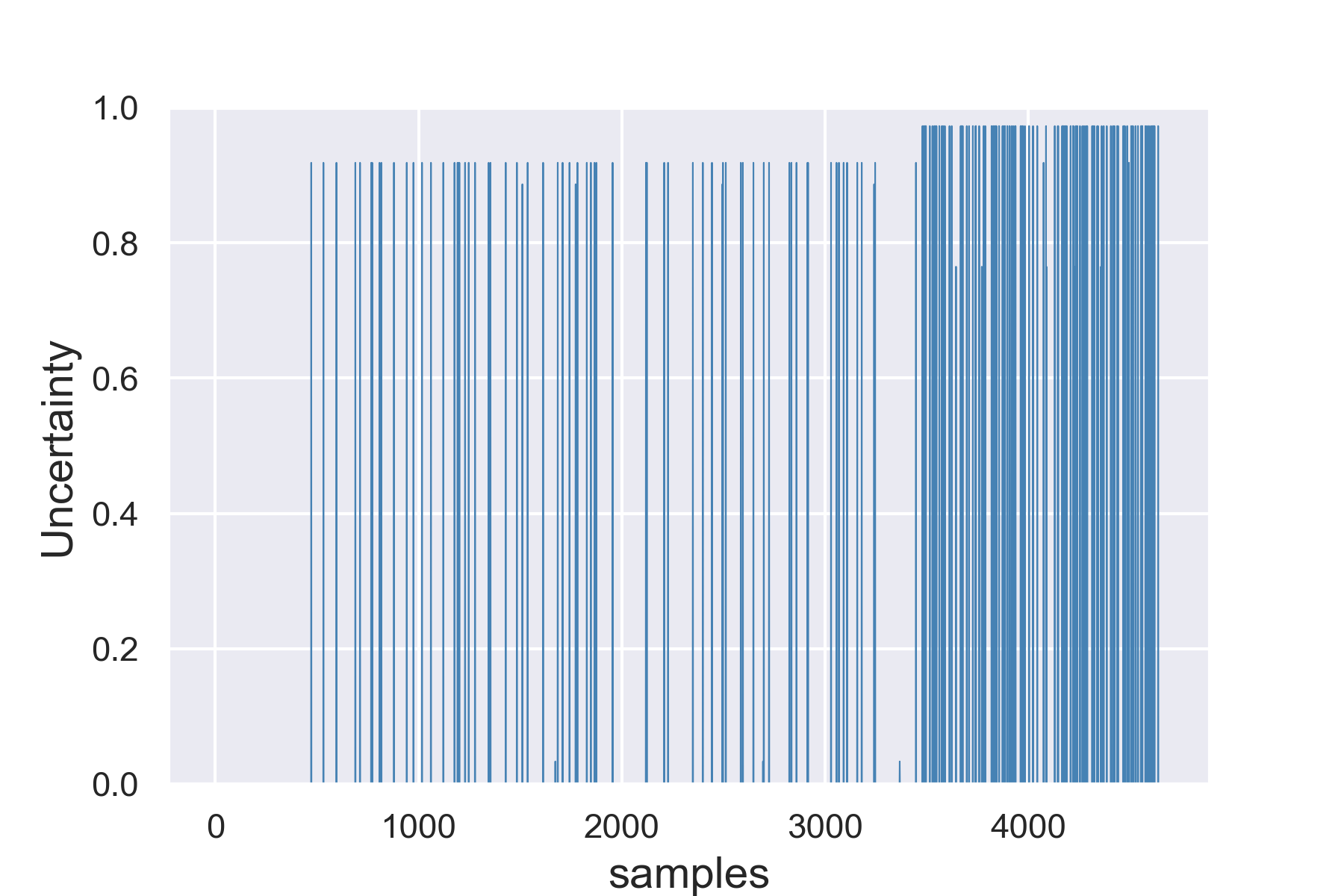}
	\caption{UQ of M3}\label{Fig__ecet_M3__UQDSET__Uncertainty}
	\end{subfigure}
	\begin{subfigure}[b]{0.3\textwidth}
	\includegraphics[width=\textwidth,keepaspectratio]{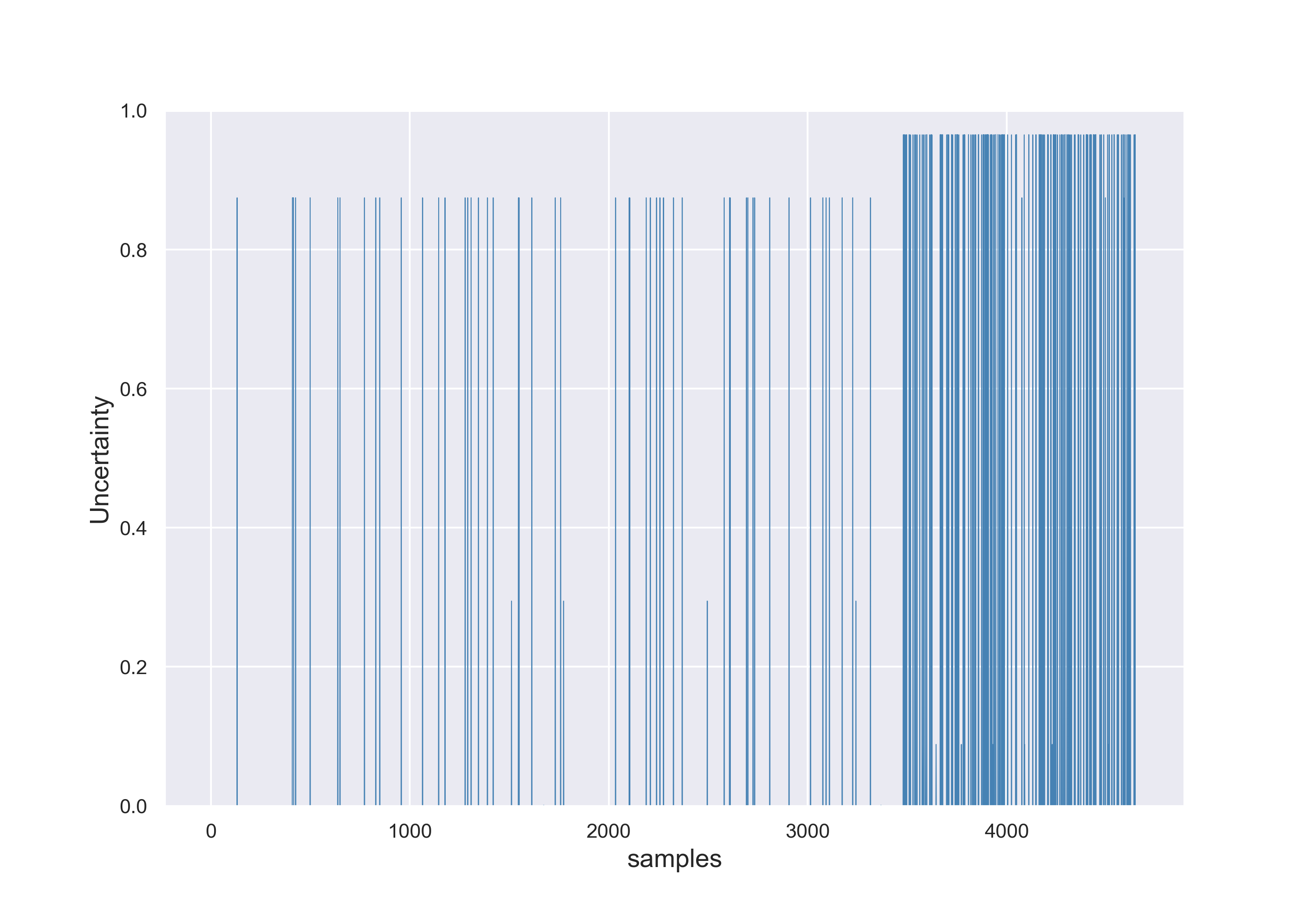}
	\caption{DSET UQ of M4}\label{Fig__ecet_M4__UQDSET__Uncertainty}
	\end{subfigure}
    \begin{subfigure}[b]{0.3\textwidth}
	\includegraphics[width=\textwidth,keepaspectratio]{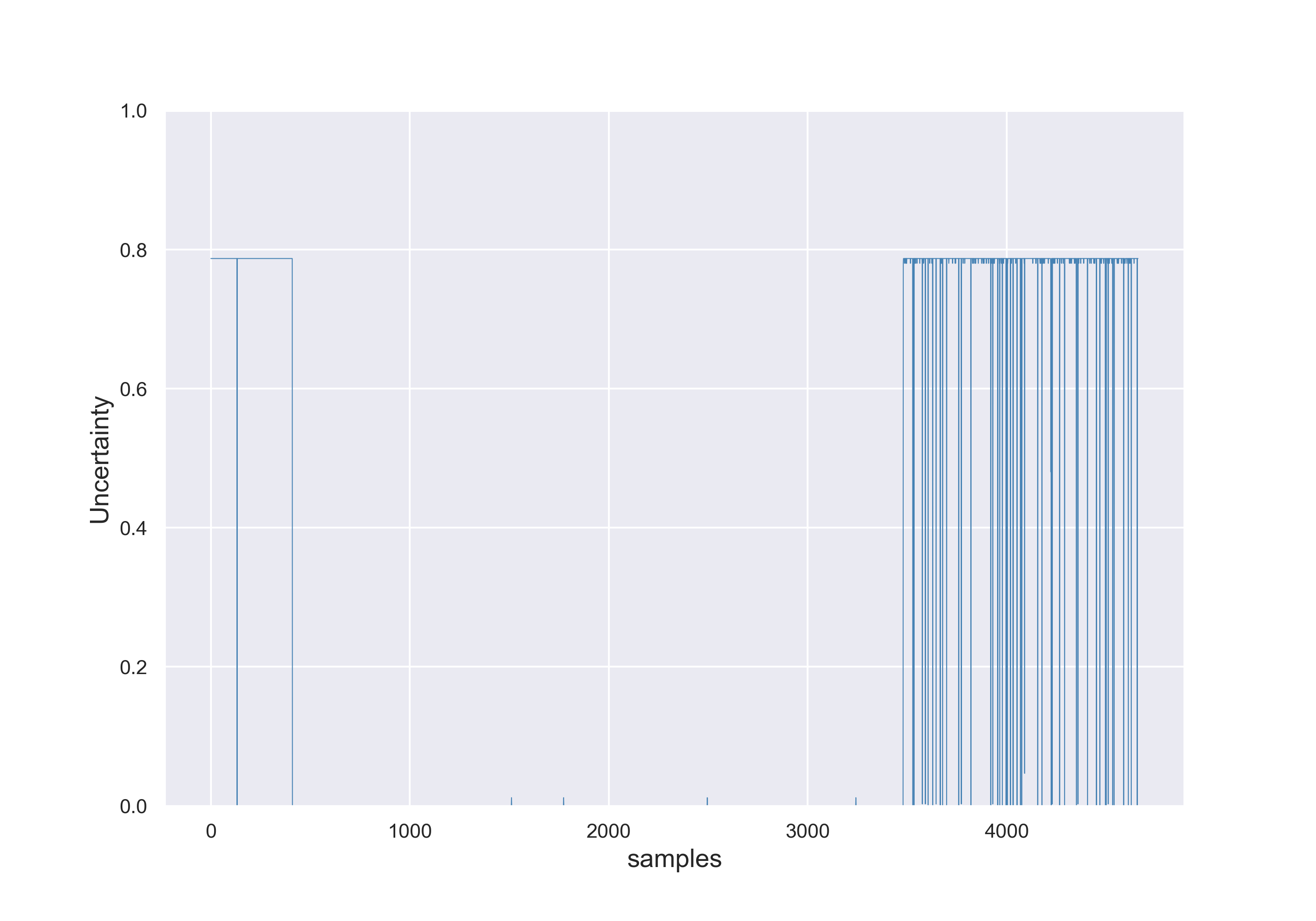}
	\caption{DSET UQ of M5}\label{Fig__ecet_M5__UQDSET__Uncertainty}
	\end{subfigure}
    \caption{Results using different models KEXT, ECET (MC EC M3), and SYS using cases (1,2,3): Confusion matrices (a)-(c), classification results (d)-(f), and DSET UQ (g)-(i).}\label{plots__ECET__BGS}
\end{figure*}

Fig. \ref{plots__ECET__BGS} presents the plots of MC ECs M3, M4, and M5 trained using the cases (1,2,3), which correspond to the failure modes (fm1, fm2, fm3), respectively. Fig. \ref{Fig__ecet_M3__cm}, \ref{Fig__ecet_M4__cm}, \ref{Fig__ecet_M5__cm} show the confusion matrices for the MC ECs M3, M4, and M5, respectively. The confusion matrices present the same performance for the MC ECs M3, M4, and M5.
Fig. \ref{Fig__ecet_M3__Classific__Prediction}, Fig. \ref{Fig__ecet_M4__Classific__Prediction}, Fig. \ref{Fig__ecet_M5__Classific__Prediction} display the predictions in blue color compared with the ground truth in red color for the MC ECs M3, M4, and M5, respectively. Likewise, in the previous case, the prediction plots are identical for the MC ECs M3, M4, and M5.
Fig. \ref{Fig__ecet_M3__UQDSET__Uncertainty}, Fig. \ref{Fig__ecet_M4__UQDSET__Uncertainty}, Fig. \ref{Fig__ecet_M5__UQDSET__Uncertainty} present the DSET UQ for MC ECs M3, M4, and M5, respectively. In contrast to the previous plots, the uncertainty is reduced as the ensemble size increases. In the case of the MC EC M5, the model presents the clearest plot, except for the fm3, which has a noisy behavior.

\subsubsection{INFUSION using the BGS data}

Table \ref{table__results__inference__classification__Infusion} presents the F1-scores of the knowledge-based model, the fusion of data-based models, and the fusion of data-based and knowledge-based models. The knowledge-based model is represented by the model using the KEXT methodology. The fusion of data-based models is represented by the models using the ECET methodology (M3, M4, M5) and an additional case performing a DSET fusion of the data-based models KNN and SVM (without the ECET methodology). The fusion of data-based models and the knowledge-based model is represented by the models using the INFUSION methodology (IFS3, IFS4, IFS5) and an additional case performing a fusion of the models KNN, SVM, and KEXT.
The KEXT model presents an average F1-score of 0.75, whereas the individual cases (1,2,3) presented values of 0.95, 0.79, and 0.52, respectively.
The ECET and INFUSION models (IFS3, IFS4, IFS5) present the best average F1-score with a value of 1.00.
The fusion of SVM and KNN presents an average F1-score of 0.96, whereas the fusion of KEXT, SVM, and KNN presents an improved average F1-score with a value of 0.98. 

\begin{table*}[!ht]
\centering
\caption{Inference results of knowledge-based model KEXT, data fusion models (ECET and SVM-KNN), and knowledge and data fusion (INFUSION and SVM-KNN-KEXT) using the cases (1, 2, 3) of the \textit{BGS} dataset, and F1-score.. The ECET models are the MC ECs M3, M4 and M5. The INFUSION models are IFS3, IFS4, and IFS5.}
\begin{tabular}{c|c|cccc|cccc}
\hline
\multirow{3}{*}{\textbf{Fault}} & \textbf{Knowledge} & \multicolumn{4}{c|}{\textbf{Data Fusion}} & \multicolumn{4}{c}{\textbf{Knowledge and Data   Fusion}} \\ \cline{2-10} 
 & \multirow{2}{*}{\textbf{KEXT}} & \multicolumn{1}{c|}{\multirow{2}{*}{\textbf{SVM-KNN}}} & \multicolumn{3}{c|}{\textbf{ECET}} & \multicolumn{1}{c|}{\multirow{2}{*}{\textbf{SVM-KNN-KEXT}}} & \multicolumn{3}{c}{\textbf{INFUSION}} \\ \cline{4-6} \cline{8-10} 
 &  & \multicolumn{1}{c|}{} & \textbf{M3} & \textbf{M4} & \textbf{M5} & \multicolumn{1}{c|}{} & \textbf{IFS3} & \textbf{IFS4} & \textbf{IFS5} \\ \hline
1 & 0.95 & \multicolumn{1}{c|}{1} & 1 & 1 & 1 & \multicolumn{1}{c|}{1} & 1 & 1 & 1 \\
2 & 0.79 & \multicolumn{1}{c|}{0.97} & 1 & 1 & 1 & \multicolumn{1}{c|}{0.98} & 1 & 1 & 1 \\
3 & 0.52 & \multicolumn{1}{c|}{0.92} & 0.99 & 0.99 & 0.99 & \multicolumn{1}{c|}{0.95} & 0.99 & 0.99 & 0.99 \\ \hline
\multicolumn{1}{l|}{Avg F1-score} & 0.75 & \multicolumn{1}{c|}{0.96} & 1.00 & 1.00 & 1.00 & \multicolumn{1}{c|}{0.98} & 1.00 & 1.00 & 1.00 \\ \hline
\end{tabular}
\label{table__results__inference__classification__Infusion}
\end{table*}

\begin{figure*}[!htbp]
	\centering
      \begin{subfigure}[b]{0.3\textwidth}
    \includegraphics[width=\textwidth,keepaspectratio]{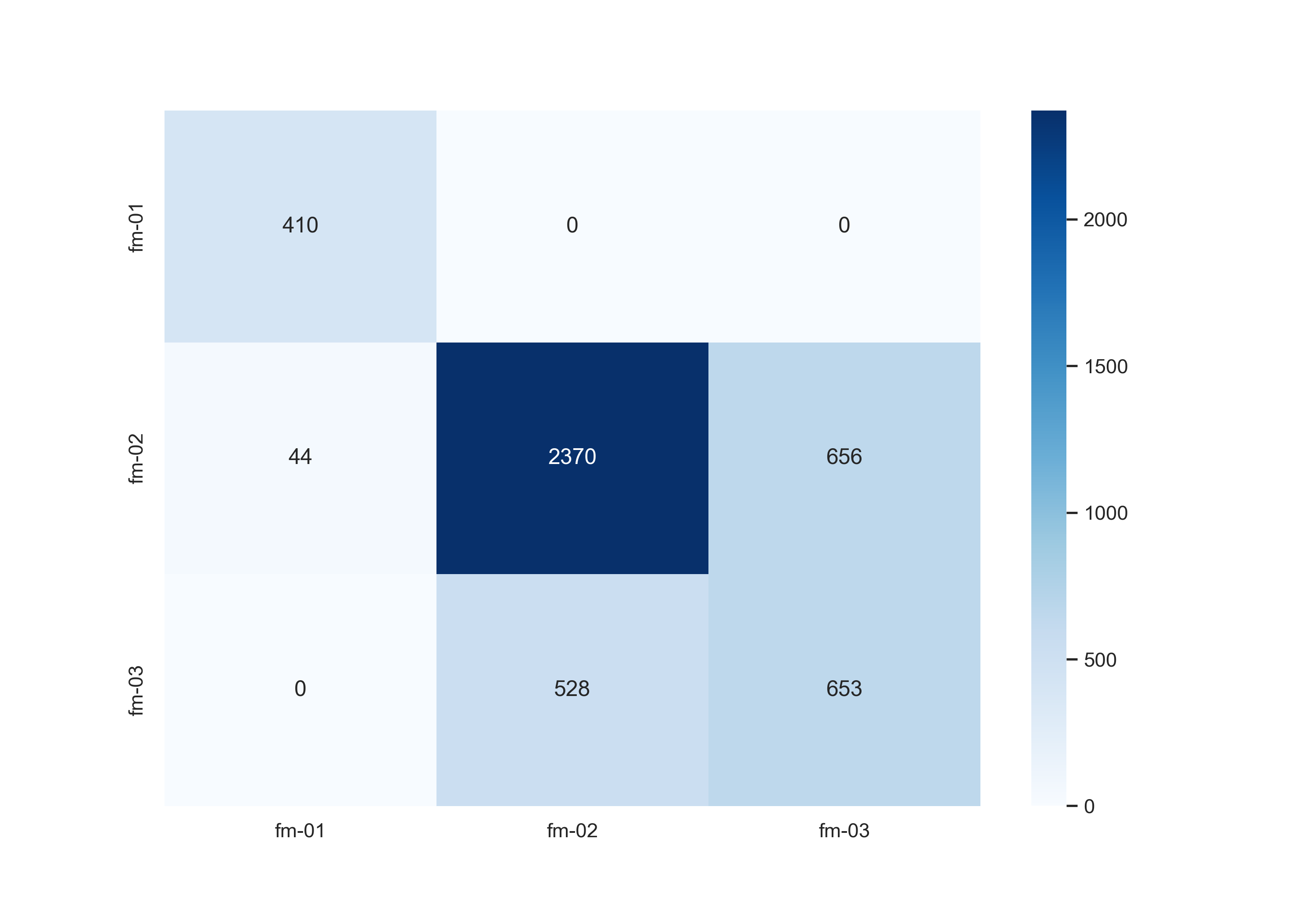}
    \caption{CM of KEXT}\label{Fig__kext__cm}
    \end{subfigure}
        \begin{subfigure}[b]{0.3\textwidth}    
	\includegraphics[width=\textwidth,keepaspectratio]{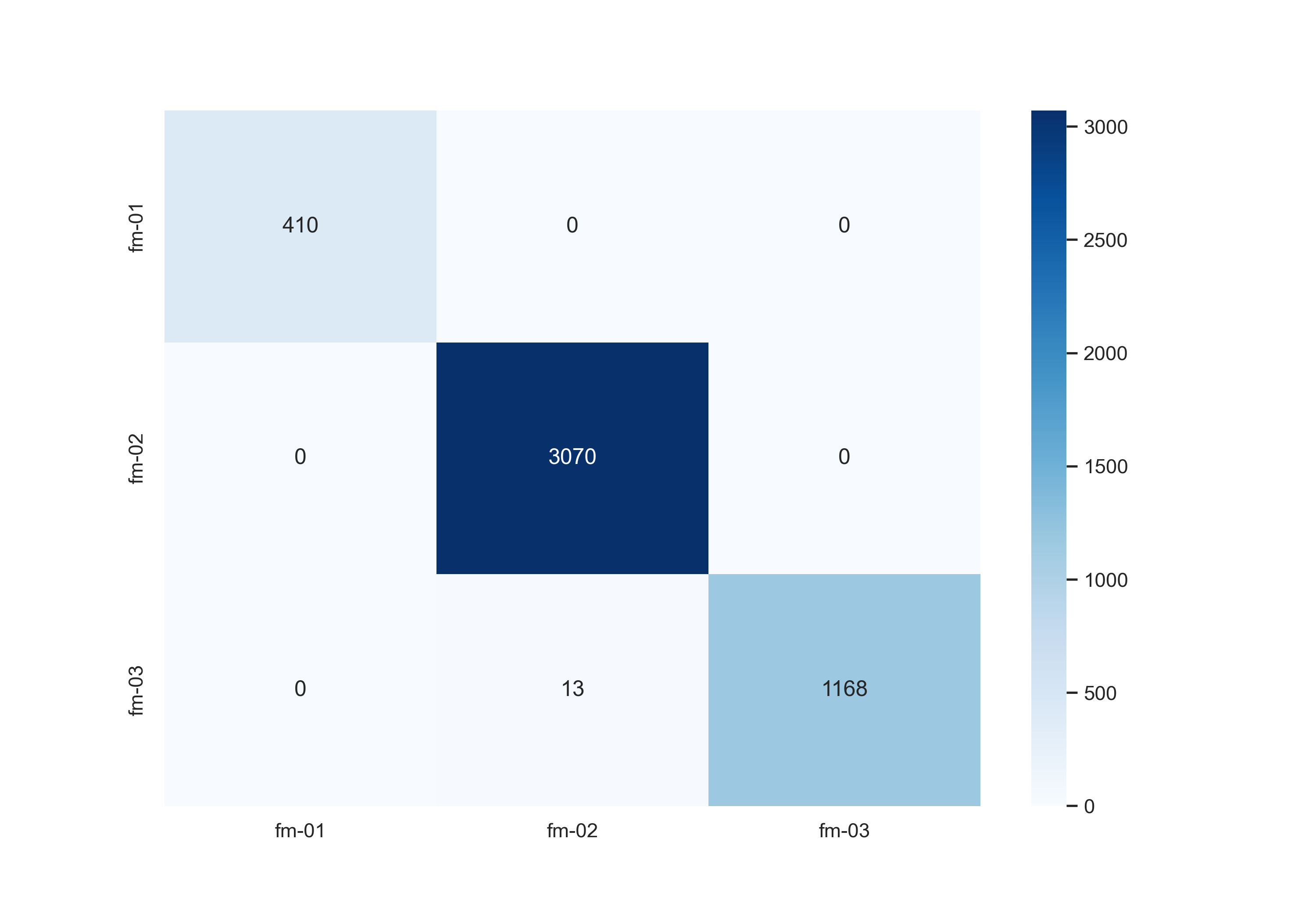}
	\caption{CM of ECET}\label{Fig__ecetM3__cm}
	\end{subfigure}
     \begin{subfigure}[b]{0.3\textwidth}
	\includegraphics[width=\textwidth,keepaspectratio]{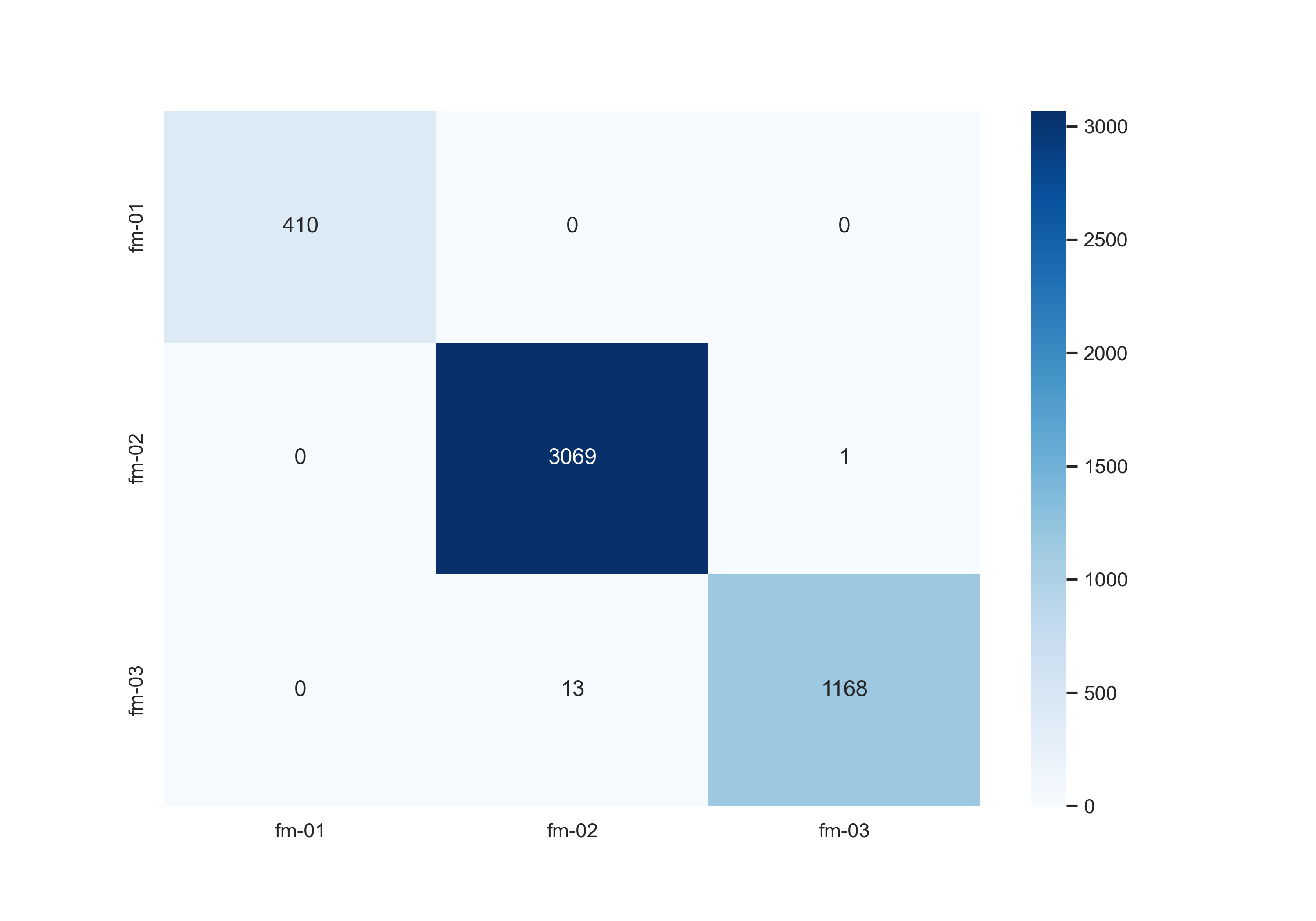}
	\caption{CM of INFUSION}\label{Fig__IFS3__cm}
	\end{subfigure}
		~
    \begin{subfigure}[b]{0.3\textwidth}
    \includegraphics[width=\textwidth,keepaspectratio]{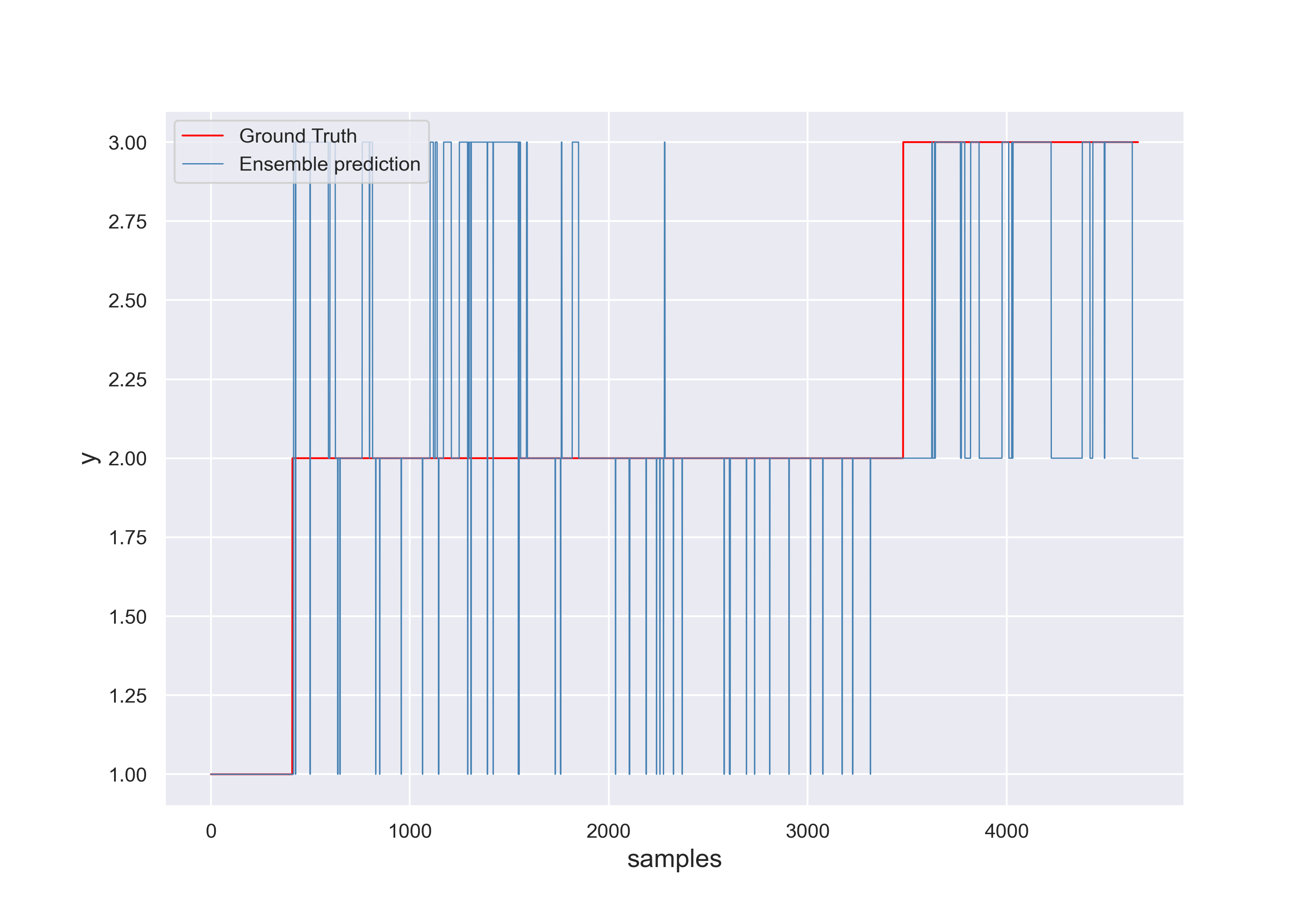}
    \caption{Predictions of KEXT}\label{Fig__kext__Classific__Prediction}
    \end{subfigure}
        \begin{subfigure}[b]{0.3\textwidth}    
	\includegraphics[width=\textwidth,keepaspectratio]{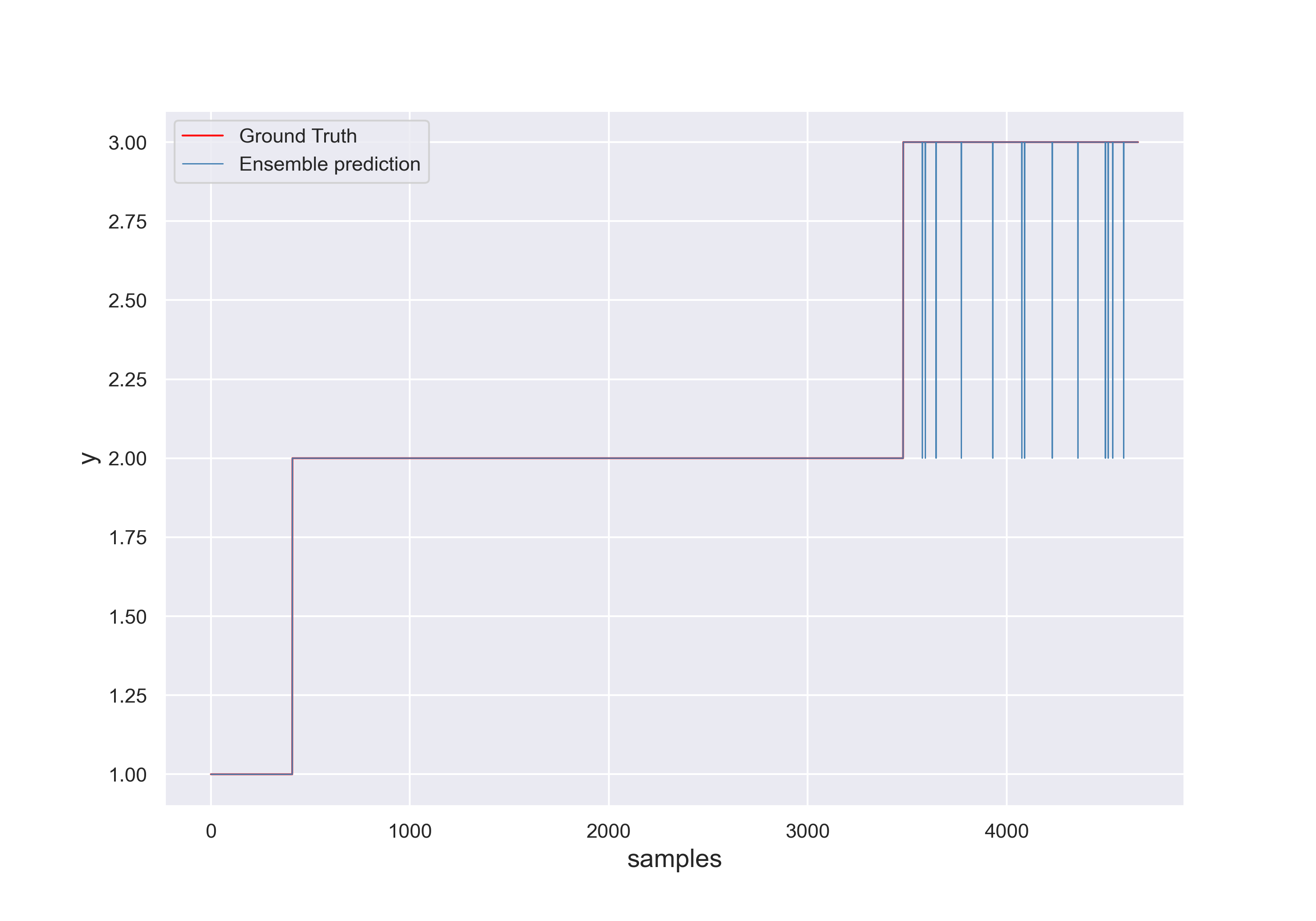}
	\caption{Predictions of ECET}\label{Fig__ecetM3__Classific__Prediction}
	\end{subfigure}
     \begin{subfigure}[b]{0.3\textwidth}
	\includegraphics[width=\textwidth,keepaspectratio]{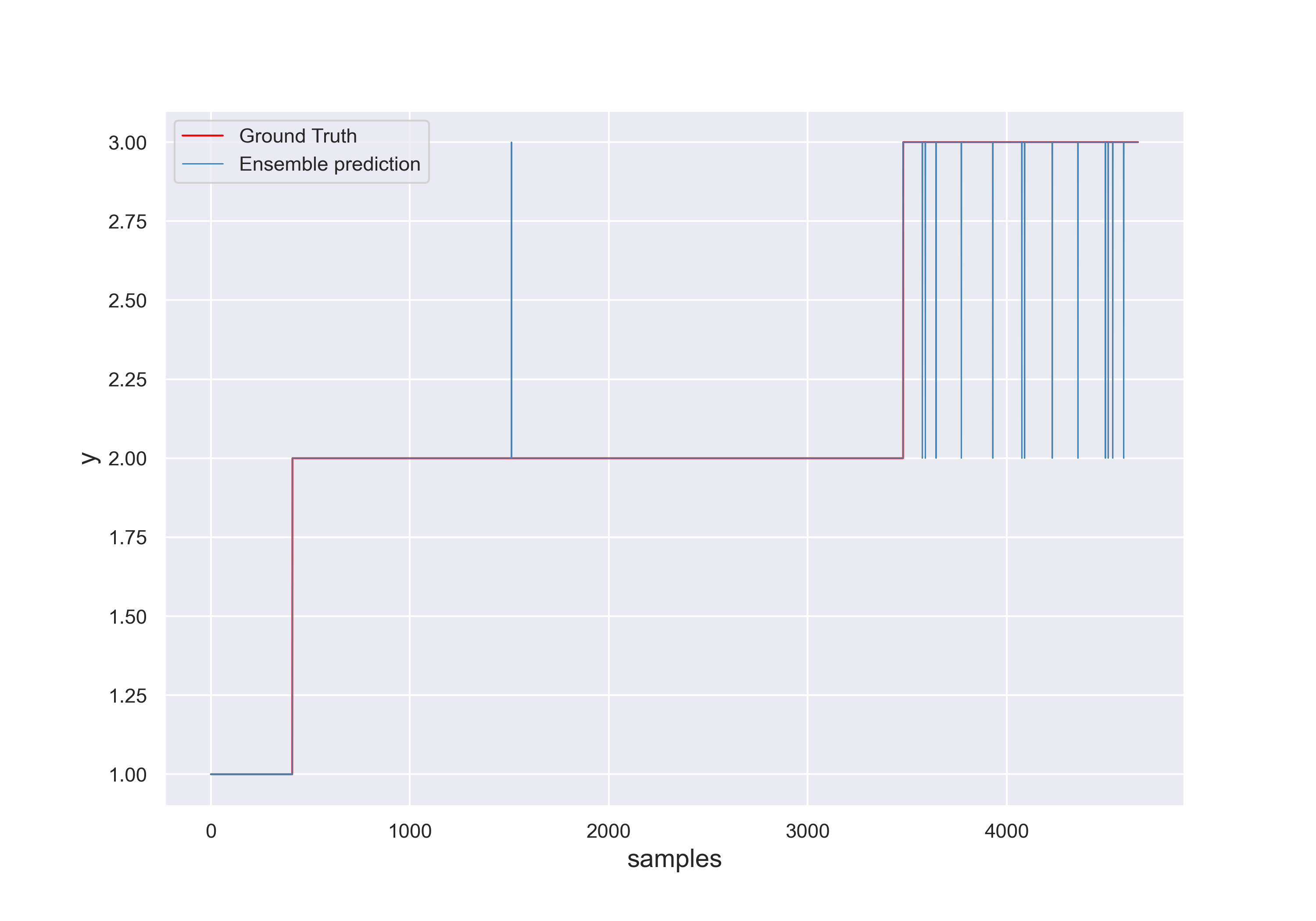}
	\caption{Predictions of INFUSION}\label{Fig__IFS3__Classific__Prediction}
	\end{subfigure}
		~
	\begin{subfigure}[b]{0.3\textwidth}
	\includegraphics[width=\textwidth,keepaspectratio]{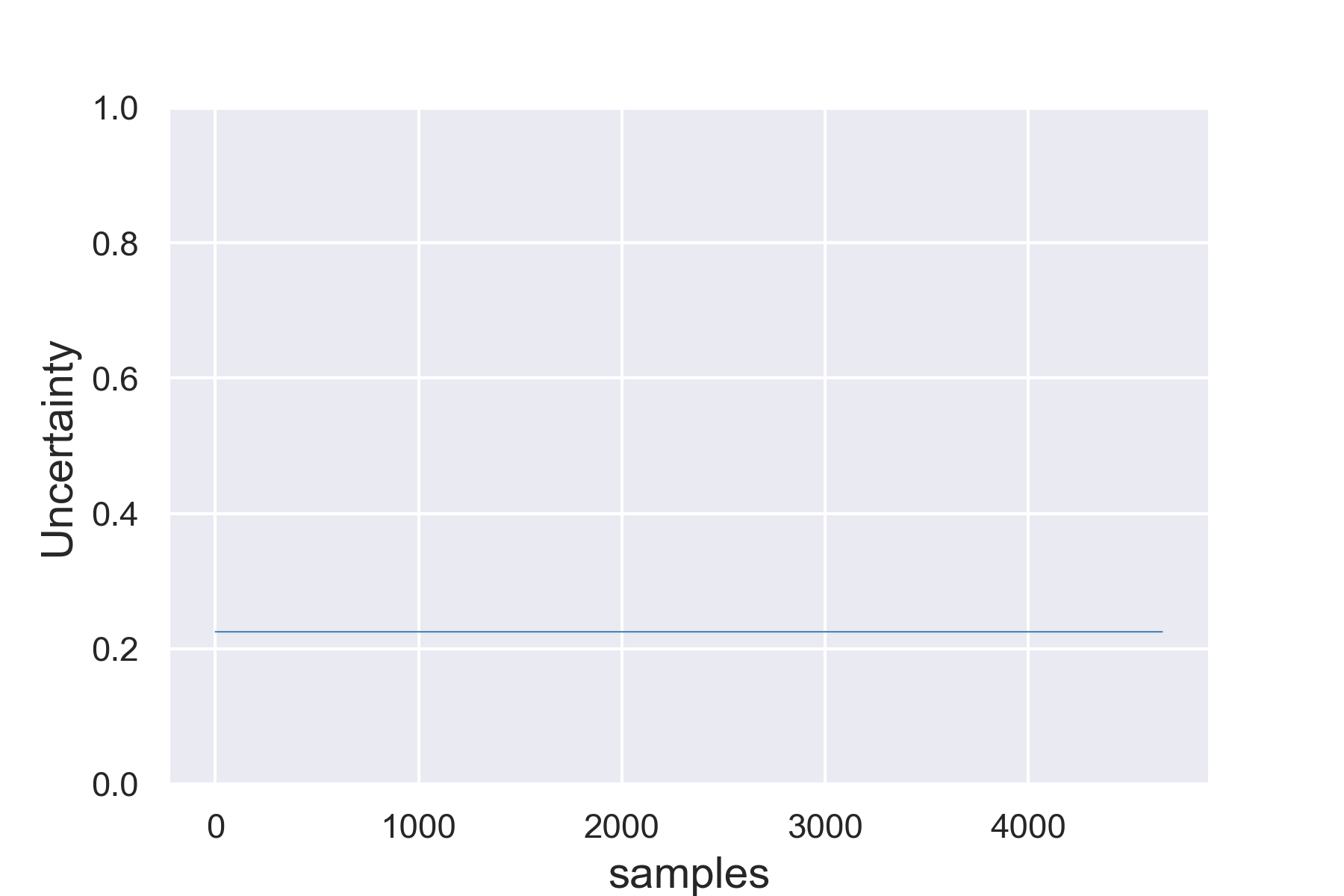}
	\caption{UQ of KEXT}\label{Fig__kext__UQDSET__Uncertainty}
	\end{subfigure}
	\begin{subfigure}[b]{0.3\textwidth}
	\includegraphics[width=\textwidth,keepaspectratio]{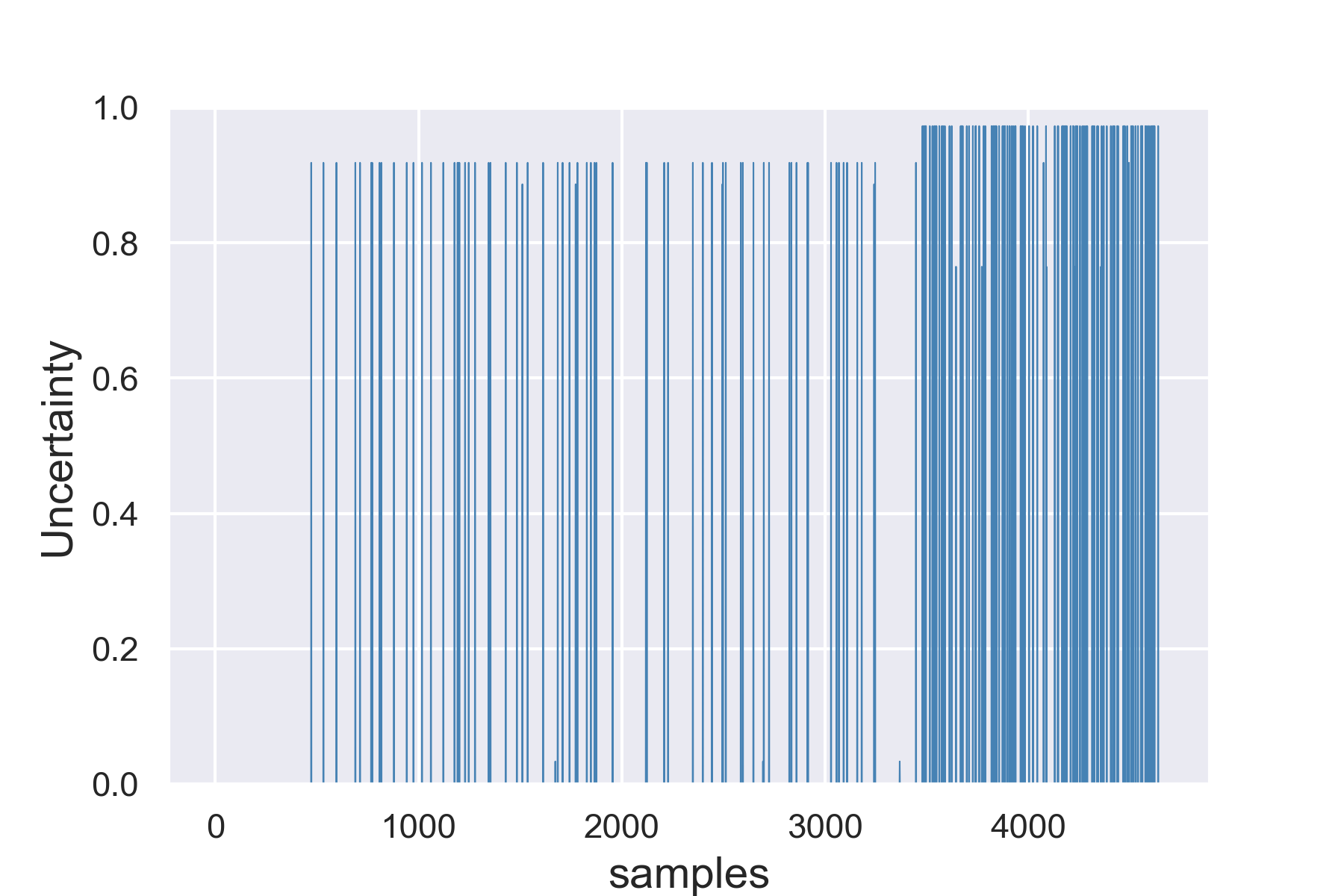}
	\caption{DSET UQ of ECET}\label{Fig__ecetM3__UQDSET__Uncertainty}
	\end{subfigure}
    \begin{subfigure}[b]{0.3\textwidth}
	\includegraphics[width=\textwidth,keepaspectratio]{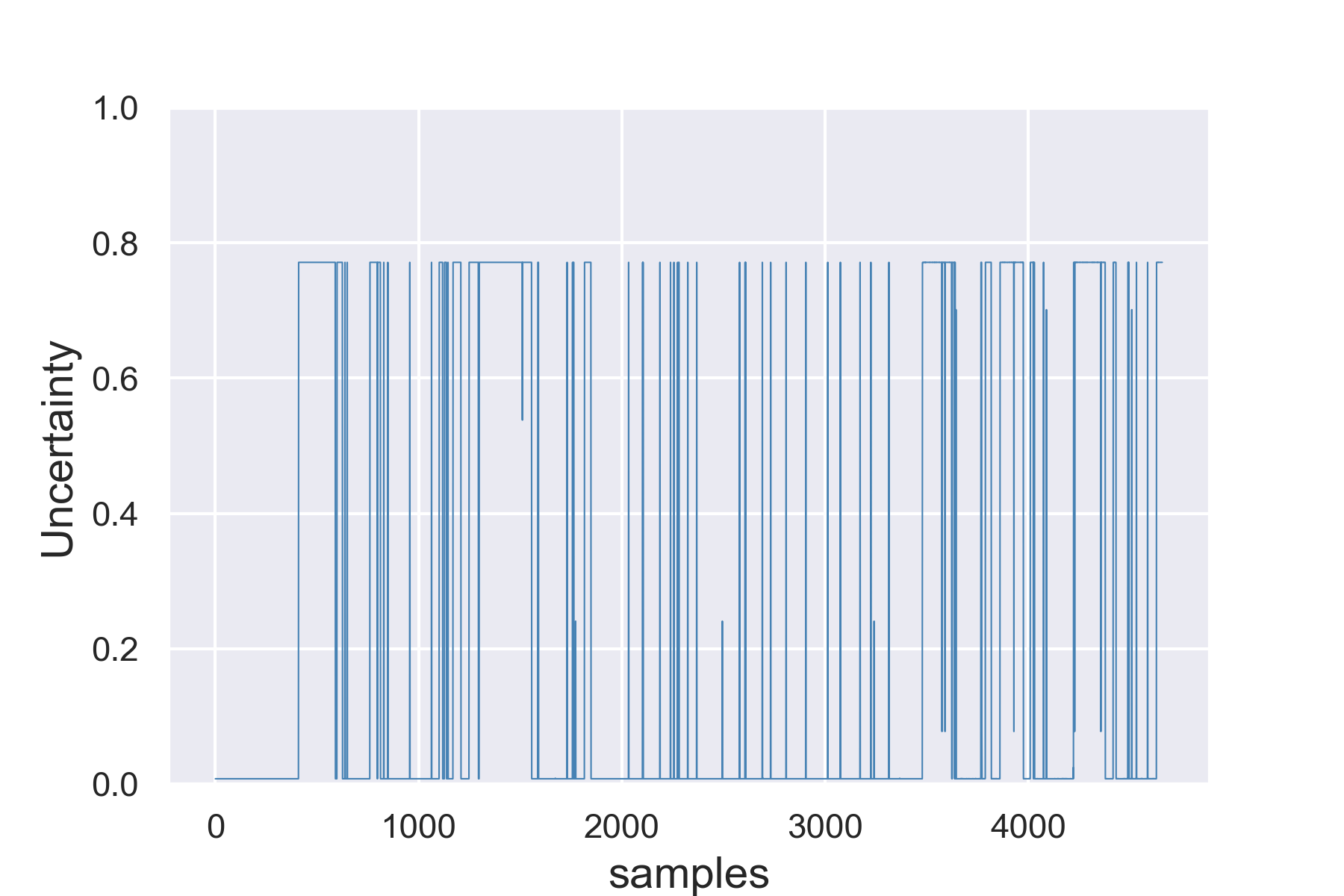}
	\caption{DSET UQ of INFUSION}\label{Fig__IFS3__UQDSET__Uncertainty}
	\end{subfigure}
    \caption{Results using different models KEXT, ECET (MC EC M3), and INFUSION (MC EC M3 and KEXT) using cases (1,2,3): Confusion matrices (a)-(c), classification results (d)-(f), and DSET UQ (g)-(i).}\label{plots__KEXT__ECET__INFUSION__BGS}
\end{figure*}

Fig. \ref{plots__KEXT__ECET__INFUSION__BGS} presents the plots of the main models: the KEXT knowledge-based model, ECET data-based model (M3), and the INFUSION model (fusion of KEXT and ECET). Fig. \ref{Fig__kext__cm}, \ref{Fig__ecetM3__cm}, \ref{Fig__IFS3__cm} show the confusion matrices for the models KEXT, ECET (M3), and INFUSION (IFS3), respectively. The confusion matrices with the best performance correspond to the models ECET and INFUSION. In contrast, KEXT presents a poor performance by detecting fm3.
Fig. \ref{Fig__kext__Classific__Prediction}, Fig. \ref{Fig__ecetM3__Classific__Prediction}, Fig. \ref{Fig__IFS3__Classific__Prediction} display the predictions in blue color compared with the ground truth in red color for the models KEXT, ECET, and INFUSION, respectively. The clearest plots correspond to the ECET and INFUSION models, whereas the KEXT model presents a noisy plot.
Fig. \ref{Fig__kext__UQDSET__Uncertainty}, Fig. \ref{Fig__ecetM3__UQDSET__Uncertainty}, Fig. \ref{Fig__IFS3__UQDSET__Uncertainty} present the DSET UQ for the models KEXT, ECET, and INFUSION, respectively. In the case of KEXT, the plot presents a continuous line since the expert team can only change the uncertainty's value. In contrast, ECET presents an extremely noisy plot for the fm3. In the case of INFUSION, the plot presents a steadier uncertainty.

It is important to remark on the INFUSION robustness, in which we perform the fusion of a high-performing ECET with a low-performing KEXT. The low performance of KEXT for some fault cases did not affect INFUSION's performance.
INFUSION performance presents a steady high performance while examining table \ref{table__results__inference__classification__Infusion} and the confusion matrix from Fig. \ref{Fig__IFS3__cm}. Alternatively, a detailed examination of the uncertainty provides an additional perspective on INFUSION's performance, in which the uncertainty presents areas with high values. Thus, uncertainty monitoring can be used to evaluate ECET and KEXT to determine the causes of low performance.

\subsection{Discussion}

The knowledge-based model KEXT presented mixed results, in which some faults are well identified or predicted. However, the strength of this approach relies on how well the rule represents a machine condition. Representing knowledge rules is a challenging task and often time demanding. An additional positive characteristic of the knowledge-based model relies on its explainability: an expert user can directly observe the logic and transform the rules. 

Alternatively, the data-based models using ECET outperformed the knowledge-based model, which is clearly reflected in the F1-scores of Table \ref{table__results__inference__classification__Infusion}. However, the relationships between the features and outputs are often hidden (except for data-based models such as DTR, where the rules can be observed). It is important to remark on the number of features the models use, in which the knowledge-based models are built using less than ten features. In contrast, the ECET models are built using 133 features. 

The fusion of data-based and knowledge-based models slightly improved the overall system's performance. The fusion model SVM-KNN-KEXT presented an improvement of fault 3 to the fusion model SVM-KNN, with scores of 0.95 and 0.92, respectively. In the case of INFUSION, the ECET results were already outstanding, resulting in a predominant effect on the fusion. The poor performance of some fault cases of KEXT did not affect the system performance.    

The INFUSION methodology performed a fusion of the KEXT knowledge-based model and the ECET data-based models. No performance changes were reported since the ECET data-based models (M3, M4, and M5) presented already outstanding performance, and the INFUSION models (IFS3, IFS4, and IFS5) presented the same performance.

\section{Conclusion}\label{section__conclusions}
We presented a novel approach for assistance systems using information fusion in production assessment. We focused on two main topics of the assistance system: improving anomaly detection and information fusion. The anomaly detection system was improved by adding the capability of automatic retraining of the models while feeding unknown fault cases into the data.
For this purpose, we presented an EC retraining strategy based on uncertainty monitoring of the EC predictions.
The retraining results of the use case validated the approach, in which the benchmark TE dataset was used to test different anomalies. Different experiments were performed to analyze the impact of the main parameters of the retraining approach, namely, threshold size, window size, and detection patience. Though the results were not entirely comparable with the literature, the approach's claim was validated, in which the EC updated the models while feeding unknown fault cases into the data.
Furthermore, we proposed an information fusion approach to combine an EC and a knowledge-based model at the decision level. The approach was tested using the data of an industrial setup. We performed an ablation study to compare the performance of the systems, namely, EC, knowledge-based models, and the fusion of both systems. The system performance reported better results while using an information fusion of the EC and the knowledge-based model, confirming, thus, the approach's claim.   

Future research includes a semi-supervised approach in which the EC results are confronted with an unsupervised model. The purpose of the approach is to validate the data samples of the detected anomaly by examining the location of the samples in the input space. Furthermore, we will test other rules of combination to improve the anomaly detection results, thus increasing the size of anomalous data.

\bibliographystyle{IEEEtran}
\bibliography{IEEEabrv,FUSIONref}

\end{document}